**Handling Noisy Labels via One-Step Abductive Multi-Target Learning and Its Application to Helicobacter Pylori Segmentation**

Yongquan Yang[a], Yiming Yang[a], Jie Chen[a], Jiayi Zheng[b], Zhongxi, Zheng[a]

[a] Institute of Pathology, West China Hospital, Sichuan University, 37 Guo Xue Road, 610041 Chengdu, China
[b] Pre-medical school, University of San Francisco, CA94117-1081 San Francisco, USA

## Abstract

Learning from noisy labels is an important concern in plenty of real-world scenarios. Various approaches for this concern first make corrections corresponding to potentially noisy-labeled instances, and then update predictive model with information of the made corrections. However, in specific areas, such as medical histopathology whole slide image analysis (MHWSIA), it is often difficult or impossible for experts to manually achieve the noisy-free ground-truth labels which leads to labels with complex noise. This situation raises two more difficult problems: 1) the methodology of approaches making corrections corresponding to potentially noisy-labeled instances has limitations due to the complex noise existing in labels; and 2) the appropriate evaluation strategy for validation/testing is unclear because of the great difficulty in collecting the noisy-free ground-truth labels. For the problem 1), we present one-step abductive multi-target learning (OSAMTL) that imposes a one-step logical reasoning upon machine learning via a multi-target learning procedure to constrain the predictions of the learning model to be subject to our prior knowledge about the true target. For the problem 2), we propose a logical assessment formula (LAF) that evaluates the logical rationality of the outputs of an approach by estimating the consistencies between the predictions of the learning model and the logical facts narrated from the results of the one-step logical reasoning of OSAMTL. Based on the Helicobacter pylori (H. pylori) segmentation task in MHWSIA, we show that OSAMTL enables the machine learning model achieving logically more rational predictions, which is beyond various state-of-the-art approaches in handling complex noisy labels.

Emails: Yongquan Yang <remy_yang@foxmail.com>, Yiming Yang <minggto@foxmail.com>, Jie Chen <jzcjedu@foxmail.com >, Jiayi Zheng <jzheng88@mail.ccsf.edu>, Zhongxi Zheng <digital.pathology@outlook.com>. Corresponding author: Zhongxi Zheng.

## 1. Introduction

Learning from noisy labels [1–3], which is a typical type of weakly supervised learning [4], concerns the situation where the labels suffer from some sort of errors. Due to the lack of accurate ground-truth labels in plenty of real-world scenarios, learning from noisy labels has become a concern of great practical value in the community of artificial intelligence. In practice, the basic idea of various approaches for this concern is first to make some corrections corresponding to potentially noisy-labeled instances, and then to update predictive model with information of the made corrections [4]. Based on this basic idea, in recent years, various approaches, including backward and forward correction [5], bootstrapping [6, 7], dimensionality-driven learning (D2L) [8], symmetric cross entropy (SCE) [9], peer loss (Peer) [10] and etc, have been proposed to handle noisy labels and have achieved state-of-the-art results in a variety of applications. However, in specific areas, such as medical histopathology whole slide image analysis (MHWSIA), it is often difficult or even impossible for experts to manually achieve the accurate ground-truth labels which leads to labels with complex noise. This situation raises two more difficult problems: 1) the methodology of approaches making corrections corresponding to the potentially noisy-labeled instances has limitations due to the complex noise existing in labels; and 2) the appropriate evaluation strategy for validation/testing is unclear because of the great difficulty in collecting the noisy-free ground-truth labels. Therefore, new advances still need to be made for this situation.

In this paper, we aim to alleviate the two raised problems. For the problem 1), we present one-step abductive multi-target learning (OSAMTL) that imposes a one-step logical reasoning upon machine learning via a multi-target learning procedure to constrain the predictions of the learning model to be subject to our prior knowledge. For the problem 2), we propose a logical assessment formula (LAF) that evaluates the logical rationality of the outputs of an approach by estimating the consistencies between the predictions of the learning model and the logical facts narrated from the results of the one-step logical reasoning of OSAMTL. The one-step logical reasoning here is in versus with iterative logical reasoning, which will be discussed later. In the one-step logical reasoning, we first extract groundings (logical facts) from the noisy labels. Second, we estimate the inconsistencies between the groundings extracted from the noisy labels and our prior knowledge about the true target of a specific task from the knowledge base. Third, we revise the groundings of the noisy labels by logical abduction based on reducing the inconsistencies estimated in the second step. Finally, we leverage the revised groundings to abduce multiple targets that can represent different modes for the true target of the specific task. Further, we impose the multiple targets abduced by the one-step logical reasoning upon machine learning via a multi-target learning procedure. The multi-target learning procedure constructs a joint loss that estimates the error between the predictions of the learning model and a set of multiple inaccurate targets, and optimizes the joint loss to update the leaning model. Since the multiple targets are abduced from the groundings revised by logical abduction based on decreasing the inconsistencies of the groundings of the noisy labels with our prior knowledge about the true target, the abduced multiple targets can contain information consistent with our prior knowledge about the true target. Hence, by imposing the abduced multiple targets upon machine learning, the multi-target learning procedure can constrain the predictions of the learning model to be subject to our prior knowledge about the true target. Moreover, to form LAF, we first narrate logical facts from the multiple targets abduced by the one-step logical reasoning. Second, we estimate the consistencies between the predictions of the learning model and the logical facts narrated from the multiple targets abduced by the one-step logical reasoning. Finally, using the estimated consistencies, we build a series of logical assessment metrics to evaluate the logical rationality of the outputs of an approach. Due to the fact that the narrated logical facts are derived from the abduced multiple targets that contain information consistent with our prior knowledge about the true target, LAF can be formed to be reasonable to estimate the logical rationality of the outputs of an approach. More details about the formulations of OSAMTL and LAF can be found in Section 3.

Recently, Zhou [11] proposed abductive learning (ABL) that accommodates and enables machine leaning and logical reasoning to work together. With given instances, a classifier (a learnt predictive model) and a knowledge base as input materials, ABL imposes logical reasoning upon machine learning in an iterative strategy that constrains the predictions of the learning model via a single-target learning procedure. Different from ABL, with given instances, available labels and a knowledge base as input materials, our presented OSAMTL only imposes logical reasoning upon machine learning by a one-step strategy that constrains the predictions of the learning model via a multi-target learning procedure. From the side of one-step logical reasoning of OSAMTL versus iterative logical reasoning of ABL, OSAMTL can be viewed as a particular case of ABL. By imposing a one-step logical reasoning upon machine learning via a multi-target learning procedure. OSAMTL is also different from various state-of-the art approaches for learning from noisy labels, including backward and forward correction [5], bootstrapping [6], D2L [8], SCE [9], Peer [10] and etc. The methodology of these state-of-the-art approaches employs a strategy that aims to identify potentially noisy-labeled instances based on various pre-assumptions about noisy-labeled instances and make some corresponding corrections to approach the true target serially. Differently, the methodology of OSAMTL employs a strategy that aims to abduce multiple targets that contain information consistent with our prior knowledge about the true target, and use the abduced multiple targets to approximate the true target parallelly. In summary, the methodology of these state-of-the-art approaches is a direct strategy while the methodology of OSAMTL is an indirect strategy. In addition, different from usual evaluation metrics defined on noisy-free ground-truth labels, LAF is defined on noisy labels which enables us to evaluate the logical rationality of the outputs of an approach without requiring noise-free ground-truth labels. More detailed analyses about OSAMTL and LAF and their differences from existing approaches can be found in Section 4.

OSAMTL is simple to implement, however, we found it is fairly effective to handle complex noisy labels under circumstances that we have some prior knowledge about the true target of a specific task. Referring to OSAMTL to implement an OSAMTL-based image semantic segmentation solution for the Helicobacter pylori (H. pylori) segmentation task [12] in MHWSIA and referring to LAF to implement a LAF-based evaluation strategy appropriate in the context of H. pylori segmentation, we show that OSAMTL is able to enable the deep CNN-based segmentation model achieving logically more rational predictions, which is beyond the capability of various state-of-the-art approaches in handling complex noisy labels. More details of OSAMTL and LAF applied to H. pylori segmentation can be found in Section 5. With this H. pylori segmentation application, we report on the first positive case which supports that OSAMTL can be effective in MHWSIA to handle complex noisy labels.

A preliminary version of this work was presented in [12], in which we described our approach for learning from noisy labels in perspective of weakly supervised multi-task learning and investigated its fundamental effectiveness. Realizing that our approach is intrinsically a type of abductive learning [11], some inaccurate descriptions exist in previous presentation, and some problems remain unaddressed, this paper adds to the initial version in the following aspects: 1) In perspective of abductive learning, we reformulate our approach into OSAMTL; 2) We propose LAF that, without requiring noisy-free ground-truth labels, evaluates the logical rationality of the outputs of an approach; 3) We add more detailed analyses about OSAMTL and ALF, and their differences from existing approaches are also presented; 4) To address the H. pylori segmentation task in MHWSIA, we implement an OSAMTL-based image semantic segmentation solution and a LAF-based evaluation strategy by referring to the presented formulations of OSAMTL and LAF; 5) On the basis of the H. pylori segmentation application, considerable new experiments and analyses are added to investigate the capability of OSAMTL in handling complex noisy labels; 6) More appropriate literature reviews are made as an addition to the initial version. In this paper, we define that multi-target learning concerns the situation where an instance of the training dataset has multiple targets for a single task. This is different from multi-task learning [13, 14] which concerns the situation where multiple different tasks are defined on the same training dataset. Realizing this difference, we change the expression ‘multi-task learning’ in our previous work [12] into the expression ‘multi-target learning’ in this paper, since we only have one task to consider here.

The rest of this paper is structured as follows. In section 2, we review related works. In section 3, we present our approach, i.e. OSAMTL and LAF. Section 4 gives detailed analyses about OSAMTL and LAF and their differences from existing approaches. In section 5, we implement an OSAMTL-based image semantic segmentation solution and a LAF-based evaluation strategy to address the H. pylori segmentation task in MHWSIA. In section 6, on the basis of the H. pylori segmentation application, we conduct extensive experiments to investigate the capability of OSAMTL in handling complex noisy labels and analyses corresponding results. Finally, in section 7, we discuss our work and future plans.

# 2. Related Work

### *2.1 Abductive learning*

Abductive learning (ABL) is a new learning paradigm that was recently proposed by Zhou [11]. ABL is targeted at unifying machine learning and logic reasoning (two separate abilities of AI) in a mutually beneficial way. On the basis of a hand-written equation parsing application, Dai, et. al., [15] presented an ALF-based solution as an instance implementation of ABL and demonstrated that the predictive model evolved by ABL are beyond state-of-the-art deep learning models. Huang, et. al., [16] proposed a semi-supervised ABL approach for theft judicial sentencing, putting forward the first work that leverages the ABL framework to tackle a real-world problem. In this paper, we present one-step abductive multi-target learning (OSAMTL) that only imposes one-step logical reasoning upon on machine learning via a multi-target learning procedure. And, based on the H. pylori segmentation task in medical histopathology whole slide image analysis (MHWSIA), we implement an OSAMTL-based solution to handle complex noise labels of this task. With a new evaluation strategy implemented by referring to the proposed logical assessment formular (LAF), we show that OSAMTL can be fairly effective to leverage our prior knowledge about the true target of the H. pylori segmentation task to enable the learning model achieving logically more rationale predictions, which is beyond the capability of various state-of-the-art approaches in handling complex noisy labels.

### *2.2 Learning from noisy labels*

Many theoretical studies [17–20] for learning from noisy labels assume the noise of noisy labels is subject to random noise. However, in practice, this assumption has limited situations. Thus, a more practical idea of various other approaches [5, 6, 8, 9, 20–23] for learning from noisy labels is to first make corrections corresponding to potentially noisy-labeled instances, and then update predictive model with information of the made corrections.

Handling massive noisy-labeled data for image classification, Xiao, et. al., [21] first prepared a small dataset with both clean and noisy labels to predict class label and noisy type, then both predictions were leveraged to correct the distribution of the given noisy labels. The made corrections and the prepared dataset were united to update the predictive model for better predictions of class label and noisy type iteratively. To learn semantic boundaries with a coarse mask drawn with a few clicks inside an object of interest, Acuna, et. al., [22] proposed an approach entitled STEAL. Via an active alignment scheme based on level set methods, STEAL iteratively refines the given coarse masks of objects using current predictions to update the predictive model for next better predictions. Regarding noisy labels as a problem of semi-supervised learning, Li, et. al., [23] proposed an approach entitled DivideMix which was constructed based on two networks. Iteratively, DivideMix first divides the noisy dataset into a labeled set (mostly clean) and an unlabeled set (mostly noisy) with a two-component GMM fit on the predictions of one network, then uses the divided datasets to train another network in semi-supervised learning via an MixMatch method that performs corrections on both labeled and unlabeled samples. In summary, to carry out the procedure of learning from noisy labels, the approach for handling massive label noise [21] requires a clean dataset in advance, STEAL [22] needs the targeted object can be clearly defined, and DivideMix [23] needs the available samples can be divided into clean and noisy two types. These premised requirements prevent the flexibility of these approaches applied to the situation where no clean dataset is available, the targeted object cannot be clearly defined, and any of the given labels cannot be confidently regarded as clean. Since the situation with the three unsatisfied requirements is usual in medical histopathology whole slide image analysis (MHWSIA), for example the H. pylori segmentation task [12], we address this situation by proposing OSAMTL and LAF in this paper.

Patrini, et. al., [5] proposed a loss correction approach which first uses the softmax output of the deep CNN trained without correction to initially approximate the transition matrix from noisy-free labels to noisy labels, then retrains the deep CNN while correcting the original loss based on the estimated transition matrix. To correct the original loss, there are backward and forward two ways. Backward correction is performed by multiplying the estimated transition probability with its corresponding loss value, while forward correction is performed by multiplying the estimated transition probability with the softmax output of the deep CNN before applying the loss function. To further reduce the estimation error for the transition matrix, Yao, et. al., [24] proposed to factorize the original transition matrix into the product of two easy-to-estimate transition matrices by introducing an intermediate class to avoid directly estimating the noisy class posterior. Reed, et. al., [6] proposed a label correction approach that first estimates the confidence of noisy labels via cross-validation, then updates the noisy labels with the estimated confidence. Ma, et. al., [8] proposed D2L that first estimates the confidence of noisy labels based on pre-assumption that dimensional expansion exacerbate overfitting, then updates the noisy labels with the estimated confidence to prevent the dimensionality of the representation subspace from expanding at later training. Wang, et. al., [9] proposed SCE that uses the prediction of deep CNN to estimate the confidence of noisy labels, then constructs a symmetric cross entropy loss with the estimated confidence to correct the original cross entropy loss. Inspired by the peer prediction literature, Liu and Guo [10] proposed to elicit and evaluate each predictive model's predictions by referring to the noisy labels which were taken as imperfect reference answers, resulting in a family of peer loss functions. Showing that any loss can be made robust to noisy labels via a simple normalization, Ma, et. al., [25] proposed a framework named Active Passive Loss (APL) which combines two robust loss functions to mutually boost each other. Generally, these approaches identify potentially noisy-labeled instances based on various pre-assumptions about noisy-labeled instances and make corresponding corrections to approach the true target directly. Differently, our proposed OSAMTL introduces a one-step logical reasoning that, via reducing the logical inconsistencies between the noise labels and our prior knowledge about the true target of a specific task, abduces multiple targets to represent different modes of the true target to approximate it indirectly.

### *2.3 Recent similar ideas*

The key idea of OSAMTL is to exploit multiple inaccurate targets abduced from weak annotations to achieve more reasonable predictions. In our previous work [12], we have found this idea is fairly effective. Preliminary experimental results led us to believe that this idea can probably be applied to many other tasks in the field of medical analysis, since the true target in this field is usually difficult to define (the consistency between various experts is always low). With similar ideas, two recent works [26, 27] also proposed to construct approaches for different weakly supervised image semantic segmentation tasks. Experiments of these two works [26, 27] have as well shown the effectiveness of similar ideas. However, the behind principles still lack of well reveal and formalization. Though our previous work [12] has tried to achieve this, further improvements still have to be made to better reveal and formulize the behind principles. In this paper, following the framework of ABL [11], we better reveal and formulate the principles behind the idea of exploiting multiple inaccurate targets abduced from weak annotations to achieve logically more reasonable predictions, avoiding simply chasing the state-of-the-art results without reasoning.

## 3. Our Approach

The outlines of OSAMTL and LAF is shown as Fig. 1. Given instances, noisy labels and a knowledge base as input materials (highlighted in black in Fig. 1), OSAMTL constitutes of three key components: 1) one-step logical reasoning (highlighted in green) that, with the given noisy labels and knowledge base, abduces multiple targets containing information consistent with our prior knowledge about the true target; 2) learning model (highlighted in blue in Fig. 1) that maps the input instances into corresponding predictions of the true target; and 3) multi-target learning procedure (highlighted in red in Fig. 1) that constrains the predictions of the learning model to be subject to the knowledge contained in the abduced multiple targets. To evaluate the logical rationality of the outputs of an approach, LAF builds a series of logical assessment metrics based on the consistencies estimated between the predictions of the learning model and the logical facts narrated from the multiple targets abduced by the one-step logical reasoning of OSAMTL.

### *3.1 One-step abductive multi-target learning*

OSAMTL imposes a one-step logical reasoning upon machine learning via a multi-target learning procedure. As shown in the top of Fig. 1, OSAMTL constitutes of four components, including input materials, one-step logical reasoning, learning model and multi-target learning procedure. In this subsection, we describe how these components are integrated to form OSAMTL.

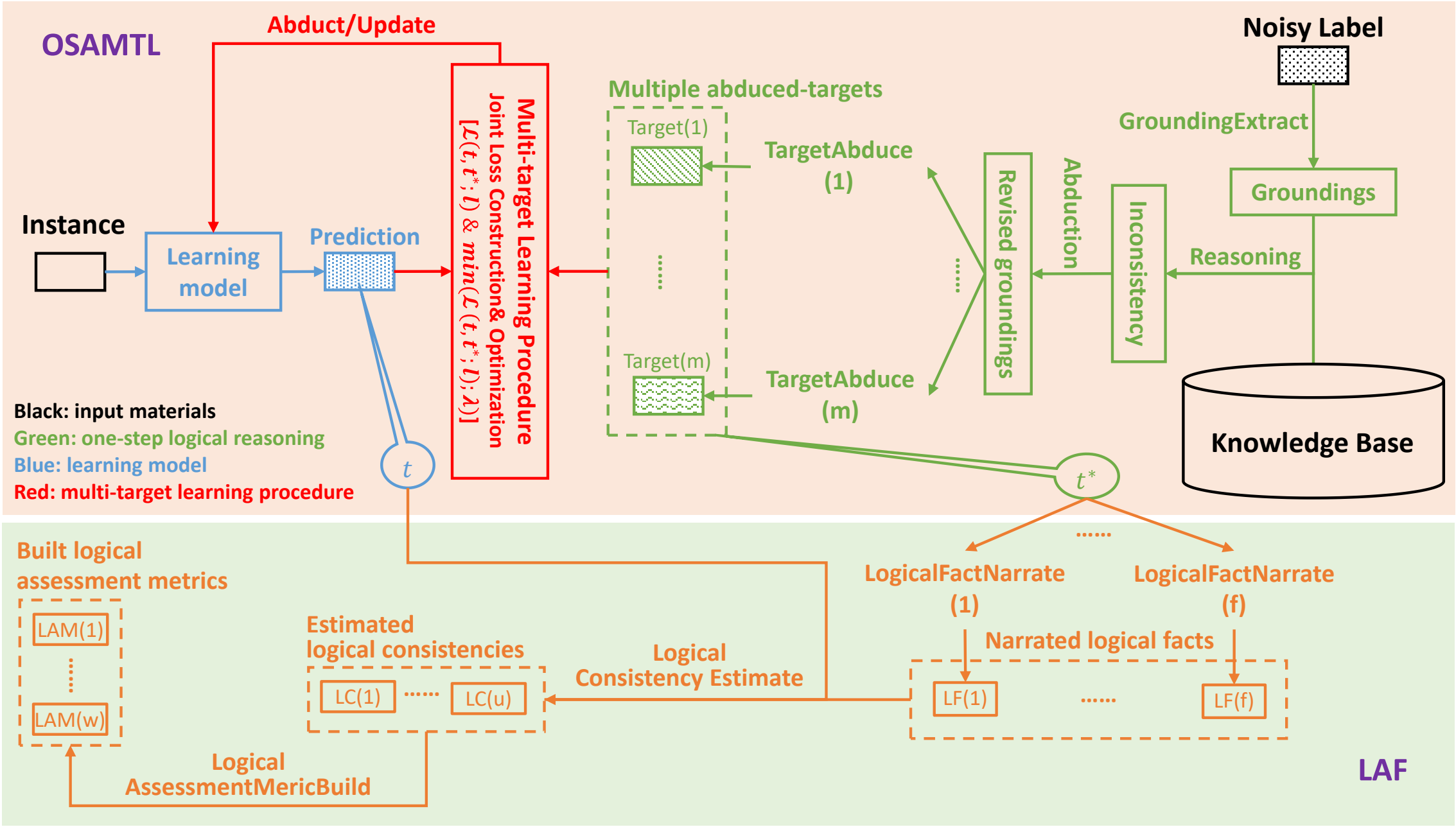

**Fig. 1**. Outlines of OSAMTL and LAF.

### 3.1.1 Input materials

The input materials of OSAMTL include instances (I) and corresponding noisy labels (L) for a specific task and a knowledge base (KB) containing a list of prior knowledge about the true target of the specific task. For simplicity, we express the given materials of OSAMTL as follows.

$$KB = \{K_1, \cdots, K_b\};\ I = \{I_1, \cdots, I_n\};\ L = \{L_1, \cdots, L_n\}.$$

### 3.1.2 One-step logical reasoning

The one-step logical reasoning, which abduces multiple targets containing information consistent with our prior knowledge about the true target with given noisy labels and knowledge base, consists of four substeps. In substep one, we extract a list of groundings from the available labels that can describe the logical facts contained in the available noisy labels. We express this as

$$G = GroundingExtract(L; param^{GE}) = \{G_1, \cdots, G_r\}. \tag{1}$$

In substep two, through logical reasoning, we estimate the inconsistency between the extracted groundings $G$ and the prior knowledge in the knowledge base $KB$. We express this as

$$Incon = Reasoning(G, KB; param^{R}) = \{Incon_1, \cdots, Incon_i\}. \tag{2}$$

In substep three, we revise the groundings of available noisy labels by logical abduction based on reducing the estimated inconsistency $Incon$. We express this (revised groundings RG) as

$$RG = LogicalAbduction(Incon; param^{LA}) = \{G_1, \cdots, G_p, \neg\ G_{p+1}, \cdots, \neg\ G_{r+z}\}. \tag{3}$$

Finally, in substep four, we leverage the revised groundings $RG$ to abduce multiple targets containing information consistent with our prior knowledge about the true target. We express this as

$$t^{*} = TargetAbduce(RG; param^{TA}) = \{Target_1, \cdots, Target_m\}. \quad (4)$$

In the four expressions (1)-(4), each $param^{?}$ denotes the hyper-parameters corresponding to the implementation of respective expression.

### 3.1.3 Learning model

The learning model maps an instance into its corresponding target prediction, which we express as

$$t = LM(I, \omega), \quad (5)$$

where $LM$ is short for learning model, and $\omega$ denotes the hyper-parameters corresponding to the construction of a specific learning model.

### 3.1.4 Multi-target learning procedure

Via a multi-target learning procedure, the multiple targets abduced in the one-step logical reasoning are jointly imposed upon machine learning to constrain its prediction of target. This multi-target learning procedure constitutes of a joint loss construction and optimization. The joint loss is constructed by estimating the error between the learning model's prediction $t$ and the set of abduced multiple targets $t^{*}$ is expressed by

$$\mathcal{L}(t, t^{*}; \ell) = \sum_{j=1}^{m} \alpha_j \ell(t, t_j) \;\; s.t. \;\; \sum_{i=1}^{m} \alpha_j = 1 \; and \; t_j \in t^{*}, \quad (6)$$

where $\ell$ denotes the hyper-parameters corresponding to the construction of the basic loss function, and $\alpha^{i}$ is the weight for estimating the error between the prediction $t$ and an abduced target $t^{i}$ of $t^{*}$. Then, we have the objective

$$\min_{t}(\mathcal{L}(t, t^{*}; \ell); \lambda), \quad (7)$$

to enable $LM$ to achieve better prediction $t$. Here, $\lambda$ denotes the hyper-parameters corresponding to the implementation of an optimization approach.

### *3.2 Logical assessment formula*

We propose a logical assessment formula (LAF) that evaluates the logical rationality of the outputs of an approach for learning from noisy labels. First, we narrate logical facts from the multiple targets abduced by the one-step logical reasoning, which we express as

$$LF = LogicalFactNarrate(t^{*}; param^{LFN}) = \left\{LFact_1, \cdots, LFact_f\right\}. \quad (8)$$

Then, we estimate the logical consistencies between the prediction of the learning model and the narrated logical facts, which we express as

$$LC = LogicalConsistencyEstimate(t, LF; param^{LCE}) = \{LC_1, \cdots, LC_u\}. \quad (9)$$

Finally, we form a series of logical assessment metrics based on the estimated logical consistencies between the prediction of the learning model and the narrated logical facts, which we express as

$$LAM = LogicalAssessmentMetricBuild(LC; param^{LAM}) = \{LAM_1, \cdots, LAM_w\}. (10)$$

Identically, each $param^{?}$ of expressions (8)-(10) denotes the hyper-parameters corresponding to the implementation of respective expression.

# 4. Analysis

## *4.1 Analysis of the methodology of OSAMTL*

The main difference between OSAMTL and ABL is that ABL imposes logical reasoning upon machine learning in an iterative strategy via a single-target learning procedure while OSAMTL imposes logical reasoning upon machine learning in a one-step strategy via a multi-target learning procedure. It is simple to implement OSAMTL, as the appropriate hyper-parameters respectively for implementations of expressions (1)-(4) can be confirmed in the one-step strategy by making the best usage of the experience we have. One good merit of ABL is that, it can gradually abduct the learning model multiple times using the single target abduced by iterative logical reasoning while OSAMTL cannot. To overcome this defect of OSAMTL, we abduce multiple targets by the one-step logical reasoning, and leverage the abduced multiple targets to constrain the learning model via a multi-target learning procedure that parallelly make as full as possible usage of the logical information contained in the abduced multiple targets. Employing the multi-target learning procedure, OSAMTL is also different from various state-of-the art approaches for learning from noisy labels, including backward and forward correction [5], bootstrapping [6], D2L [8], SCE [9] , peer loss (Peer) [10] and etc. The practical methodology of these state-of-the-art approaches employs a direct strategy, which aims to make corrections corresponding to potentially noisy-labeled instances to directly approach the true target. Generally, these state-of-the-art approaches are based on various pre-assumptions about noisy-labeled instances (see Section 2.2). Differently, the methodology of OSAMTL employs an indirect strategy, which aims to abduce multiple targets that can meet various logical facts of the true target and use a summarization of these abduced multiple targets to indirectly realize the true target with a logical approximation. OSAMTL provides new thoughts for more effectively exploiting noisy labels to achieve logically more rational predictions. The visualized differences of OSAMTL respectively from ABL and various other approaches for learning from noisy labels are provided in Supplementary 1.

## *4.2 Essence of the multi-target learning procedure of OSAMTL*

We show that the essence of the multi-target learning procedure of OSAMTL is that it can enable the learning model to learn from a weighted summarization of the abduced multiple targets $t^*$. To achieve that, we first introduce a result, as follows:

**Theorem 1**. *Suppose that the unknown accurate ground-truth target can be approximately realized as a set of abduced multiple inaccurate targets, i.e.,* $t^* = \left\{t_j \mid TargetAbduce(RG; param^{TA}); j \in \{1, \cdots, m\}\right\}$. *Then, for a classification or a regression task, the joint loss constructed by Eq. (6) of OSAMTL, i.e.,* $\mathcal{L}(t, t^*; \ell) = \sum_{j=1}^{m} \alpha_j \ell(t, t_j)$, *can be theoretically expressed as* $\mathcal{L}(t, t^*; \ell) = \ell\left(t, \sum_{j=1}^{m} \alpha_j t_j\right) + c$, *where* $c$ *is a constant term.*

We then prove Theorem 1 by following two proofs.

**Proof 1**. When we use average cross entropy (ACE) to estimate the error between two elements for a two-class classification task, the basic loss function $\ell(\cdot,\cdot)$ can be denoted by

$$ACE(t,t_0) = -\frac{1}{n}\sum_{k=1}^{n}\left[t_0^{k,f}\, log(t^k) + \left(1 - t_0^{k,f}\right) log(1 - t^k)\right]$$
$$s.t.\ t_0^{k,f} \cup \left(1 - t_0^{k,f}\right) = t_0 \tag{11}$$

where $n$ is the number of instances in the training data, $t_0^f$ is the foreground class of the target $t_0$, and $1 - t_0^f$ is the background class of the target $t_0$. Referring to Eq. (11), we rewrite Eq. (6) by

$$\mathcal{L}(t,t^*;ACE) = \sum_{j=1}^{m}\alpha_j\left\{-\frac{1}{n}\sum_{k=1}^{n}\left[t_j^{k,f}\, log(t^k) + \left(1 - t_j^{k,f}\right) log(1 - t^k)\right]\right\}$$
$$= -\frac{1}{n}\sum_{k=1}^{n}\left[\sum_{j=1}^{m}\alpha_j\, t_j^{k,f}\, log(t^k) + \sum_{j=1}^{m}\alpha_j\left(1 - t_j^{k,f}\right) log(1 - t^k)\right]. \tag{12}$$

Plugging $t_0 = \sum_{j=1}^{m}\alpha_j t_j$ and substituting into Eq. (11), we have

$$ACE\left(t,\sum_{j=1}^{m}\alpha_j t_j\right) = -\frac{1}{n}\sum_{k=1}^{n}\left[\sum_{j=1}^{m}\alpha_j\, t_j^{k,f}\, log(t^k) + \sum_{j=1}^{m}\alpha_j\left(1 - t_j^{k,f}\right) log(1 - t^k)\right]. \tag{13}$$

Comparing Eq. (12) with Eq. (13), theoretically we can have

$$\mathcal{L}(t,t^*;ACE) = ACE\left(t,\sum_{j=1}^{m}\alpha_j t_j\right). \tag{14}$$

**Proof 2**. When we use the mean squared error (MSE) to estimate the error between two elements for a regression task, the basic loss function $\ell(\cdot,\cdot)$ can be denoted by

$$MSE(t,t_0) = \frac{1}{n}\sum_{k=1}^{n}\left(t^k - t_0^k\right)^2. \tag{15}$$

Referring to Eq. (15), we rewrite Eq. (6) by

$$\mathcal{L}(t,t^*;MSE) = \sum_{j=1}^{m}\alpha_j\frac{1}{n}\sum_{k=1}^{n}\left(t^k - t_j^k\right)^2$$
$$= \frac{1}{n}\sum_{k=1}^{n}\left[\left(t^k - \sum_{j=1}^{m}\alpha_j t_j^k\right)^2 + \sum_{j=1}^{m}\alpha_j {t_j^k}^2 - \left(\sum_{j=1}^{m}\alpha_j t_j^k\right)^2\right]$$
$$= \frac{1}{n}\sum_{k=1}^{n}\left[\left(t^k - \sum_{j=1}^{m}\alpha_j t_j^k\right)^2\right] + D(t^*) \tag{16}$$

where $D(t^*) = \frac{1}{n}\sum_{k=1}^{n}\left(\sum_{j=1}^{m}\alpha_j {t_j^k}^2 - \left(\sum_{j=1}^{m}\alpha_j t_j^k\right)^2\right)$ is the variance for the multiple targets of $t^*$ and is a constant. Plugging $t_0 = \sum_{j=1}^{m}\alpha_j t_j$ and substituting into Eq. (15), we have

$$MSE\left(t,\sum_{j=1}^{m}\alpha_j t_j\right) = \frac{1}{n}\sum_{k=1}^{n}\left(t^k - \sum_{j=1}^{m}\alpha_j t_j^k\right)^2 \tag{17}$$

Comparing Eq. (17) with Eq. (16), theoretically we can have

$$\mathcal{L}(t,t^*;MSE) = MSE\left(t,\sum_{j=1}^{m}\alpha_j t_j\right) + D(t^*). \tag{18}$$

According to Theorem 1, we can see that through Eq. (7), the multi-target learning procedure of OSAMTL is a reasonable strategy to enable the learning model to achieve logically rational results. In fact, learning from the weighted summarization of multiple targets can lead to trade-off among the abduced multiple targets, which contain various facts logically consistent with our prior knowledge about the true target, thus to logically approximate the true target. From Eq. (14) and Eq. (18), we can also note that theoretically, learning from a set of abduced multiple targets through the multi-target learning procedure of OSAMTL is equivalent to learning from a weighted summarization of the set of abduced multiple targets. Therefore, for a regression task, the multi-target learning procedure of OSAMTL can be conveniently transformed into a single-target learning procedure by summarizing the abduced multiple targets to form a single target. However, for a classification task, we consider retaining the multi-target learning procedure of OSAMTL can smooth the classification process (increase the number of fussy classes, the weights for the errors of which in the loss function are greater than 0 and less than 1, to be alike to regression which can be viewed as classification with unlimited number of classes) while still maintaining the property of classification.

### *4.3 Analysis of LAF*

LAF is formed by estimating the consistencies between the predictions of the learning model and the logical facts narrated from the multiple targets abduced in the one-step logical reasoning, which means LAF is defined on noisy labels and enables us to evaluate the logical rationality of an approach without noise-free labels. This makes LAF different from usual evaluation metrics that are defined on noise-free labels. LAF is reasonable to estimate the logical rationality of the predictions of an approach for leaning with noisy labels, due to the fact that the narrated logical facts are derived from the abduced multiple targets that contain logical facts revised by logical abduction based on reducing the inconsistencies to our prior knowledge about the true target. Commonly, if the predictions of an approach for learning from noisy labels have consistencies to the logical facts, which conform to our prior knowledge about the true target, to be logically rationale, then this approach can be in a point of view regarded as effective to handle noisy labels. Hence, if the predictions of an approach for learning with noisy labels can show more logical rationality via LAF, then this approach can be regarded as being logically more effective to handle noisy labels. As a result, LAF can be a new addition, that does not require noise-free labels, to usual evaluation strategies built on noise-free labels for evaluating approaches for learning from noisy labels.

## 5. OSAMTL applied to H. pylori segmentation

We apply the presented OSAMTL and proposed LAF to H. pylori segmentation in medical histopathology whole slide image analysis (MHWSIA). This section is structured as follows. In section 5.1 we give some background information about H. pylori segmentation in MHWSIA. Referring to OSAMTL, in section 5.2, we implement an OSAMTL-based image semantic segmentation solution for H. pylori segmentation. And referring to LAF, in section 5.3, we implement an LAF-based evaluation strategy reasonable to estimate the logical rationality of the outputs of a solution in the context of H. pylori segmentation. The outline of the OSAMTL-based image semantic segmentation solution and the LAF-based evaluation strategy for H. pylori segmentation is shown as Fig. 2.

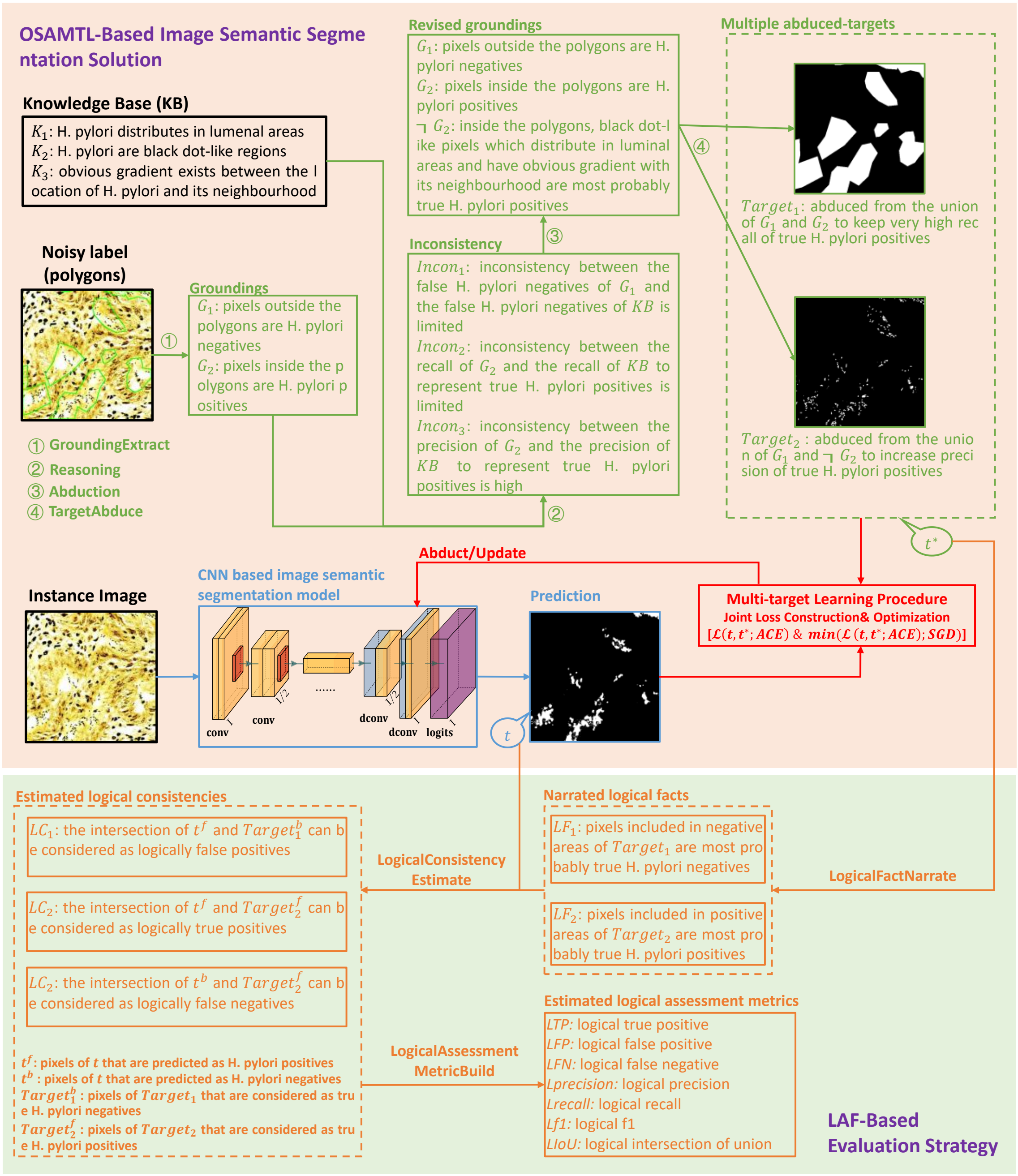

Fig. 2. Outline of OSAMTL and LAF applied to H. pylori segmentation in MHWSIA.

## *5.1 Background*

Researches have shown that Helicobacter pylori (H. pylori) is centrally related to the pathobiology of gastric cancer [8]. In 2012, one million cases of gastric cancer were estimated around the world [9]. However, presenting an appropriate computer-aided solution for H. pylori segmentation in MHWSIA is difficult. The factors that are responsible for this most likely are: 1) the great difficulty of accurately to label the morphologies of H. pylori, and 2) the complexity of processing whole slide images, of which a single slide typically has billions of pixels. An illustration for the challenge of H. pylori segmentation is shown in Fig. 3. It can be observed from the right image patch of Fig. 3 that it is highly difficult to accurately label the morphologies of H. pylori, due to its messy and confusing distribution (red boxes and blue boxes). In addition, we can also notice that the right cropped image patch occupies only a very small fraction of the left whole slide image. It can be confirmed by this that the complexity for the process of WSIs is massive. These two factors confirm that it is almost impossible for pathologist experts to manually achieve accurate ground-truth labels for H. pylori. More information about H. pylori can be found in our previous work [12].

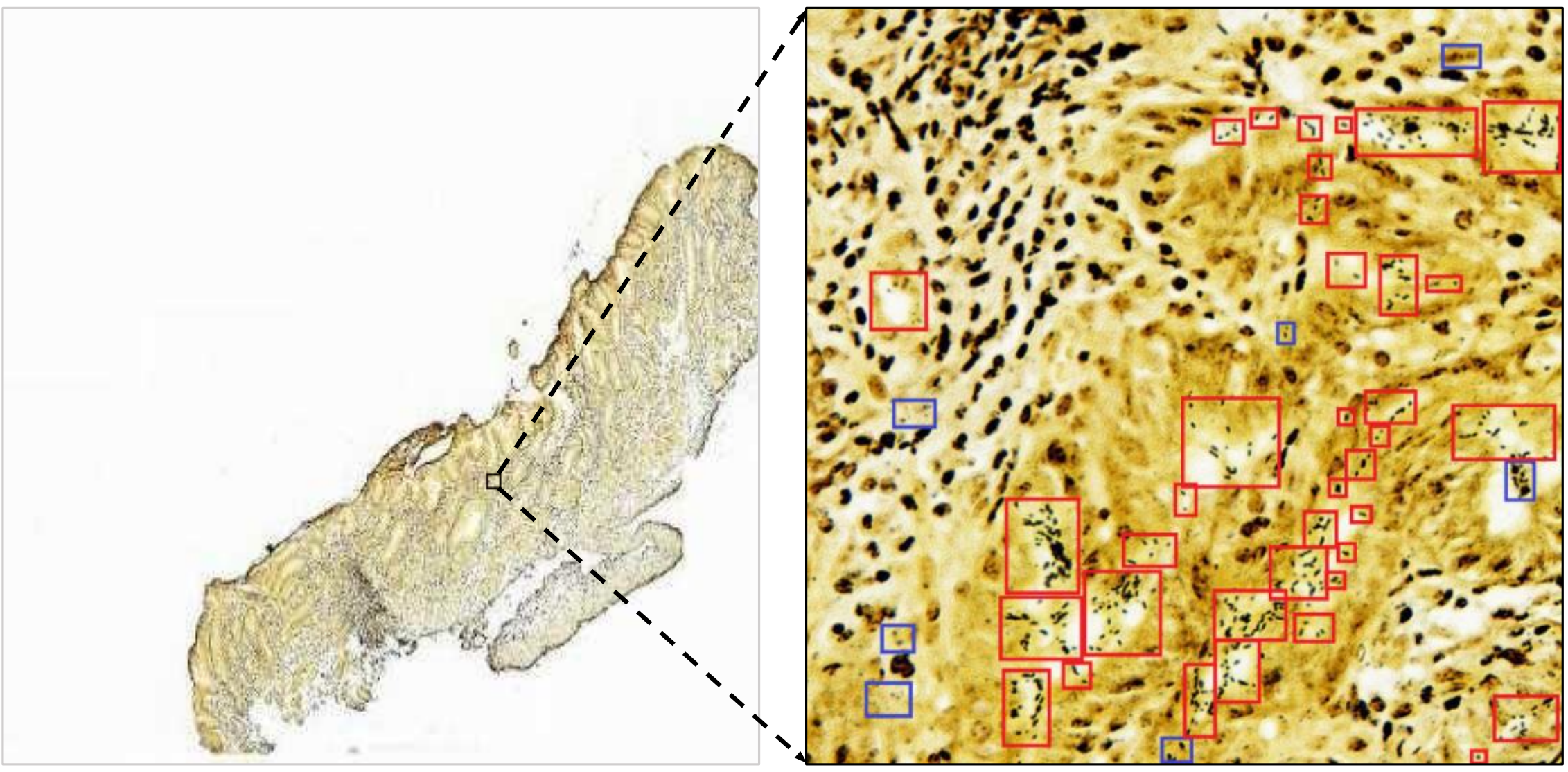

Fig. 3. Illustration for the challenge of the computer-aided H. pylori segmentation task in medical histopathology whole slide image analysis. This figure is quoted from our previous work [12]. Left: A digital 1×magnification histopathology whole slide image of a WS-stained gastric biopsy slide. Right: A cropped 40×magnification image patch of the left whole slide image at the boxed area. In the right image patch, red bounding boxes are H. pylori positive (black dot-like regions) areas, while blue bounding boxes are confusing areas. Pathology experts annotated these boxes shown in the right image patch.

### *5.2 OSAMTL-based image semantic segmentation solution*

Following the four components of OSAMTL presented in section 3.1, for the task of H. pylori segmentation, we implement an OSAMTL-based image semantic segmentation solution. The outline of this instance implementation is shown in the top of Fig. 2. Note, the proofs for the validation of the reasonings for this instance implementation are provided in Supplementary 2.

#### 5.2.1 Knowledge base, instances and labels

Knowledge base (KB), instances (I) and labels (L) are three required input materials of OSAMTL. In this H. pylori segmentation case, the knowledge base contains a list of prior knowledge about the true morphologies of H. pylori, which are shown in Table 1 and denoted as

$$KB = \{K_1, K_2, K_3\}.$$

Table 1. Knowledge provided for H. pylori segmentation

| Knowledge Base |
| --- |
| $K_1$: H. pylori distributes in luminal areas |
| $K_2$: H. pylori are black dot-like regions |
| $K_3$: obvious gradient exists between the location of H. pylori and its neighbourhood |

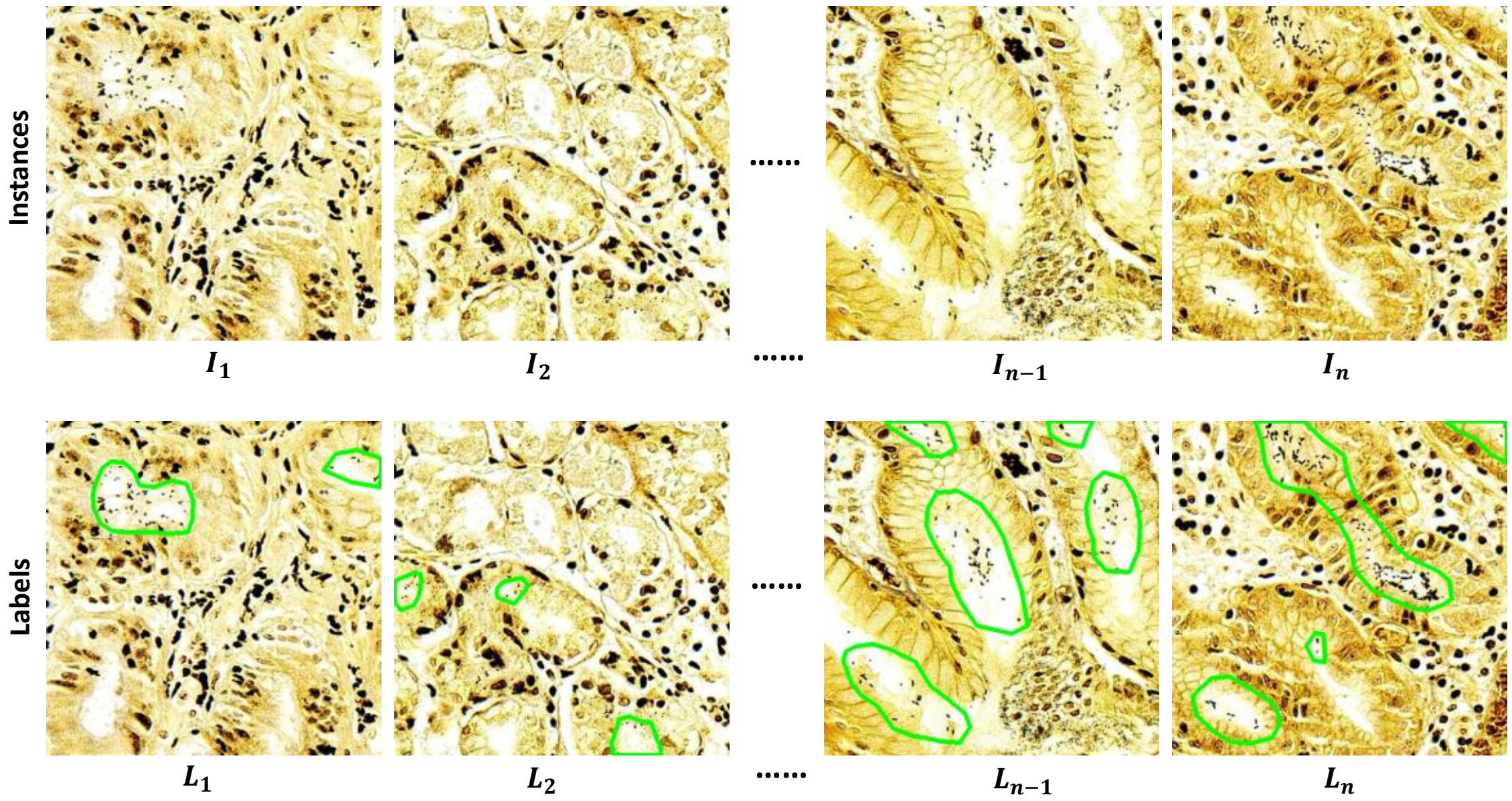

Fig. 4. Instances and corresponding labels for H. pylori segmentation task.

Due to the great difficulty of accurately to label the morphologies of H. Pylori, we employ polygon annotations to roughly mark the H. pylori positive areas. Polygon annotations significantly reduce the amount of time for labelling the morphologies of H. Pylori while contain heavy noise. Due to the massive size of whole slide images, we crop them into a list of image patches and label H. Pylori positive areas on the cropped image patches. The cropped image patches and corresponding polygon annotations are used as the other two input materials (instances and labels) for OSAMTL. Some examples of instances and labels are shown as Fig. 4. Referring to Fig. 4, in this H. pylori segmentation case, we denote the instances and labels for OSAMTL as

$$I = \{I_1, I_2, \cdots, I_{n-1}, I_n\};\ L = \{L_1, L_2, \cdots, L_{n-1}, L_n\}.$$

**5.2.2 Abduce multiple targets via one-step logical reasoning**

As the logical reasoning component of OSAMTL employs a one-step strategy, the appropriate hyper-parameters respectively for implementations of expressions (1)-(4) presented in section 3.1.2 can be confirmed by making the best usage of the experience we have. We can implement this one-step logical reasoning component with propositional logic, instead of requiring additional information (such as a noisy-free labeled dataset) to parameterize expressions (1)-(4) with predicate logic.

**(1) GroundingExtract**

The *GroundingExtract* step takes the labels $L$ as input and produces a list of groundings that describe the logical facts of the labels $L$.

Referring to Eq. (1), we use the semantics contained in $L$ as $param^{GE}$ to implement the *GroundingExtract* step, which can produce groundings as follows

$$G = GroundingExtract(L; \{semantics\ contained\ in\ L\})$$
$$= \{G_1, G_2\}.$$

Details of the extracted groundings are provided in Table 2.

Table 2. Details of the extracted groundings

| Extracted Groundings |
|---|
| $G_1$: pixels outside the polygons are tumour negatives |
| $G_2$: pixels inside the polygons are tumour negatives |

**(2) Reasoning**

The *Reasoning* step takes the groudings $G$ produced by the *GroundingExtract* step and $KB$ as inputs and produces a list of inconsistencies that describe the gap between the extracted groundings $G$ and $KB$ from various perspectives.

Based on the extracted groundings $G$ and $KB$, we derive Reasoning 1.

**Reasoning 1**. *Given $G$ and $KB$, then true H. pylori negatives represented by $G_1$ are almost all correct, the recall of $G_2$ to represent true H. pylori positives is very high, and the precision of $G_2$ to represent true H. pylori positives is very low*.

Referring to Eq. (2), we use Reasoning 1 as $param^R$ to implement the *Reasoning* step, which produces a list of inconsistencies as follows

$$Incon = Reasoning(G, KB; \{Reasoning\ 1\})$$
$$= \{Incon_1, Incon_2, Incon_3\}.$$

Details of the estimated inconsistencies are provided in Table 3.

Table 3. Details of the estimated inconsistencies

| Estimated Inconsistencies |
|---|
| $Incon_1$: inconsistency between the false H. pylori negatives of $G_1$ and the false H. pylori of $KB$ is limited |
| $Incon_2$: inconsistency between the recall of $G_2$ and the recall of $KB$ to represent true H. pylori positives is limited |
| $Incon_3$: inconsistency between the precision of $G_2$ and the precision of $KB$ to represent true H. pylori positives is high |

**(3) Abduction**

The *Abduction* step takes $Incon$ produced by the *Reasoning* step as input and produce a list of revised groundings that describe logical facts more consistent with the logical facts of $KB$.

As both $Incon_1$ and $Incon_2$ are limited, we should keep $G_1$ and $G_2$ unchanged. However, at the meantime, keeping $G_2$ unchanged will make $Incon_3$ high. Thus, in order to reduce $Incon_3$, we should revise $G_2$ to increase the precision of representing true H. Pylori positives. We revise $G_2$ by the conjunction of $G_2$ and $KB$. Based on the conjunction of $G_2$ and $KB$, we derive Reasoning 2.

**Reasoning 2**. *Given $G_2$ and $KB$, then black dot-like pixels inside $G_2$ which distribute in luminal areas and have obvious gradient with its neighbourhood are true H. pylori positives with high probability ($\neg G_2$).*

Now, we show that the result of Reasoning 2 ($\neg G_2$) can, under some additional hypotheses, increase the precision of representing true H. Pylori positives, thus to reduce $Incon_3$. Based on $\neg G_2$ and two additional hypotheses (underlined parts below), we derive Reasoning 3.

**Reasoning 3**. Given $\neg G_2$, *the precision of $G_2$ to represent true H. Pylori positives is TP/(TP+FP), and hypothesises that <u>the false H. pylori positives of $\neg G_2$ is smaller than true H. pylori positives of $\neg G_2$</u> and <u>the false H. pylori positives of $G_2$ is larger than the true H. pylori positives of $G_2$</u>, then the precision of $\neg G_2$ to represent true H. Pylori positives is higher than the precision of $G_2$ to represent true H. Pylori positives*.

In fact, the underlined parts of Reasoning 3 are two additional hypothesises that can be fulfilled by appropriate image processing procedures described in the TargetAbduce part to produce $\neg G_2$ and the fact that false H. pylori positives take the majority of $G_2$.

Referring to Eq. (3), we use Reasoning 2 and 3 as $param^{LA}$ to implement the Abduction step, which produces a list of revised groundings, as follows

$$RG = LogicalAbduction(Incon; \{Reasoning\ 2, Reasoning\ 3\})$$
$$= \{G_1, G_2, \neg G_2\}.$$

De tails of the revised groundings are provided in Table 4.

Table 4. Details of the revised groundings

| Revised Groundings |
|---|
| $G_1$: pixels outside the polygons are H. pylori negatives |
| $G_2$: pixels inside the polygons are H. pylori positives |
| $\neg G_2$: black dot-like pixels inside $G_2$ which distribute in luminal areas and have obvious gradient with its neighbourhood are true H. pylori positives with high probability |

**(4) TargetAbduce**

The *TargetAbduce* step takes $RG$ produced by the *Abduction* step as input and produces a list of multiple targets to more appropriately represent the true H. pylori positives. Specifically, we abduce $G_1$ and . $G_2$ into target one (see the top of Fig. 5). Since, in the *Reasoning* step, we have shown that true H. pylori negatives represented by $G_1$ are almost all correct and the recall of $G_2$ to represent true H. pylori positives is very high, both of which are consistent with our prior knowledge in $KB$, target one is able to confidently eliminate the majority of true H. pylori negatives to keep a very high recall of representing true H. pylori positives. Moreover, as $G_1$ and $G_2$ describes H.

pylori negatives and H. pylori positives very clearly, they can be conveniently abduced into target one using different values to signify H. pylori negatives and H. pylori positives, for instance as follows

$$NP = \begin{Bmatrix} 0: \text{H. pylori negative,} \\ 1: \text{H. pylori positive} \end{Bmatrix}$$

Thus, referring to Eq. 4 and using $NP$ as $param^{TA}$, we can implement the *TargetAbduce* step for target one.

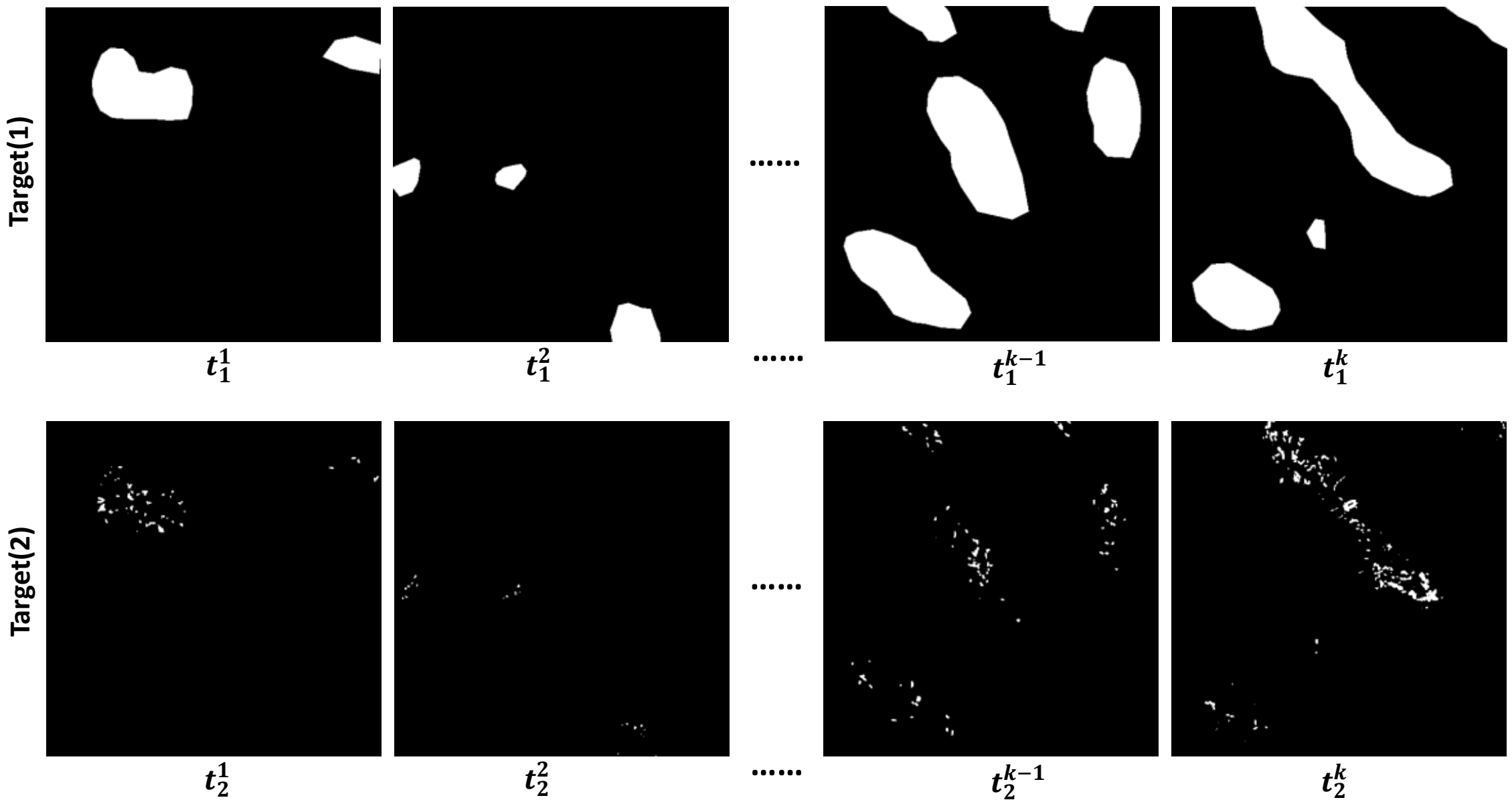

Fig. 5. Examples for the abduced multiple targets.

We also abduce $G_1$ and $\neg G_2$ into target two (see the bottom of Fig. 5). However, to abduce $\neg G_2$ into part of target one, appropriate image processing procedures must be carefully selected as have shown in the *Abduction* step that the number of false H. pylori positives in $\neg G_2$ should be less than the true H. pylori positives in $\neg G_2$ to increase the precision of representing true H. pylori positives. To achieve this, we should identify confident H. pylori positives in $G_2$ and consider the rest as H. pylori negatives. According to the grounding $\neg G_2$, we perform two image processing procedures to abduce it into part of target two. Firstly, grey threshold segmentation is leveraged to extract rough regions containing H. pylori from the polygon annotated areas. The reason that grey threshold segmentation is chosen lies in that the locations of H. pylori positives are close to black colour. Secondly, the edge detector Canny is used to locate H. pylori, since the gradient between the location of H. pylori and its neighbourhood is quite obvious. These two steps eventually produce more fine-grained morphologies of H. pylori on images, identifying confident H. pylori positives in $G_2$ and consider the rest as H. pylori negatives. For simplicity, we denote the two performed image processing procedures as

$$IP = \begin{Bmatrix} IP_1: \text{gray threshold segmentation,} \\ IP_2: \text{edge detector Canny} \end{Bmatrix}.$$

Thus, referring to Eq. (4) and using $IP$ and $NP$ as $param^{TA}$, we can implement the *TargetAbduce* step for target two.

In practice, we indeed found that target two commonly contains true H. pylori positives more than false H. pylori positives, i.e., the hypothesis that FP-EFP is smaller than TP-ETP for Reasoning 3 can be fulfilled. As a result, target two is able to significantly increase the precision of representing true H. pylori positives.

On the basis of these two implementations, we express the *TargetAbduce* step, which produces two targets representing different modes of the true target, as follows

$$t^* = TargetAbduce(RG; \{NP, IP\})$$
$$= \begin{Bmatrix} TargetAbduce(G_1, G_2; NP), \\ TargetAbduce(G_1, \neg\, G_2; \{IP, NP\}) \end{Bmatrix}$$
$$= \begin{Bmatrix} Target_1\text{: top row of Fig. 4}, \\ Target_2\text{: bottom row of Fig. 4} \end{Bmatrix}.$$

### 5.2.3 Deep CNN-based image semantic segmentation model

Since H. pylori segmentation needs to be performed on whole slide images, to speed up the segmentation process, we construct a tiny deep convolutional neural network (CNN) backbone, instead of using very deep backbones like VGG16 [28], GoogleNet [29] or ResNet [30], for image semantic segmentation. Based on this tiny backbone, we implement a symmetric image semantic segmentation architecture referring to the most commonly used FCN [31], which is the representative for the fully convolutional networks based solutions and has inspired various other solutions [32–36] achieving state-of-the-art performances in image semantic segmentation. As a result, the deep CNN-based image semantic segmentation model built for H. pylori segmentation is shown as Fig. 6.

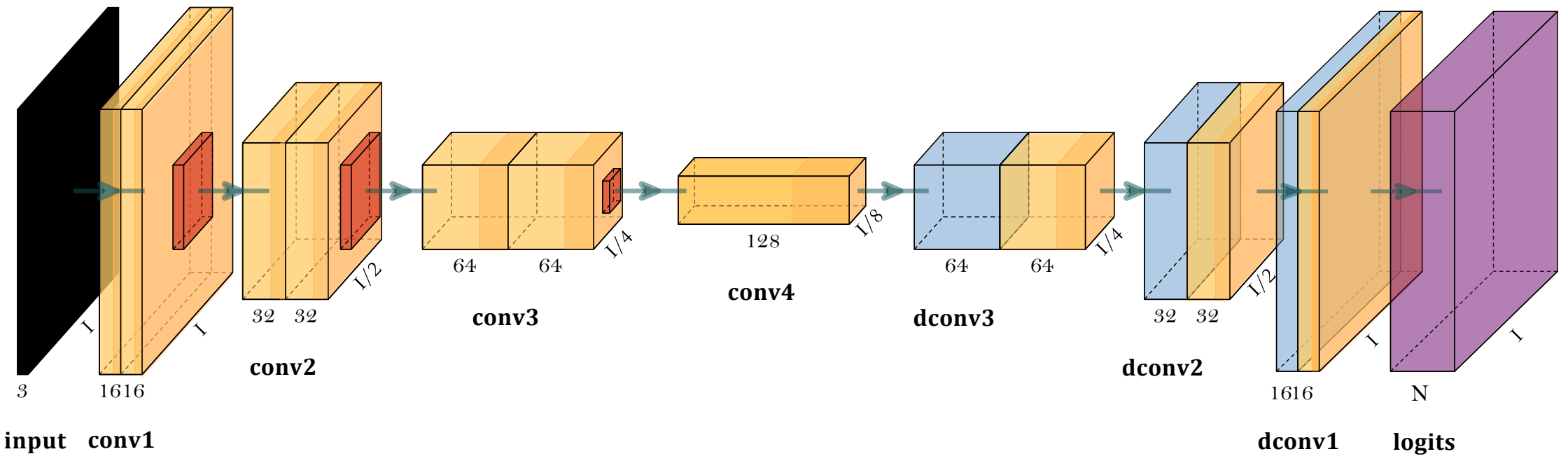

Fig. 6. Deep CNN-based image semantic segmentation model built for H. pylori segmentation.

We denote $\{cnn_l\}_{l=0}^{X}$ as the transformation for each of the $X$ layers from the DCNN architecture shown in Fig. 6 and assume that $\{cnn_l\}_{l=0}^{X}$ is parameterized by $\{w_l\}_{l=0}^{X}$. Referring to Eq. (5) and using instances shown in the top of Fig. 4, for this H. pylori segmentation case, we denote the machine learning model for the OSAMTL-based image semantic segmentation solution as

$$t = ML(I; DCNN) \qquad s.t.\ DCNN = \{\{cnn_l\}_{l=0}^{X}, \{w_l\}_{l=0}^{X}\}.$$

### 5.2.4 Constrain segmentation model via multi-target learning

Based on the abduced two targets (Section 5.2.2) and the built deep CNN-based image semantic segmentation model (Section 5.2.3), we implement the multi-target learning procedure of OSAMTL (joint loss construction and optimization, Eq. (6) and (7)) to constrain the segmentation model for better prediction.

Usually, for an image semantic segmentation task, the fundamental problem is inclined to be the classification task for pixel-level labelling. We employ ACE (Eq. (11)) as the base loss function to estimate the error between the predictions of segmentation model ($t$) and corresponding multiple targets ($t^*$). Referring to Eq. (6), for this H. pylori segmentation, we denote the joint loss by

$$\mathcal{L}(t,t^*;ACE) = \sum_{j=1}^{2} \alpha_j \left\{-\frac{1}{n}\sum_{k=1}^{n}\left[Target_j^{k,f}\, log(t^k) + \left(1 - Target_j^{k,f}\right) log(1-t^k)\right]\right\},$$

where the number of pixels in $t$ is denoted by $n$. Acquiescently, both $\alpha_1$ and $\alpha_2$ are set to 0.5. As we built the segmentation model based on deep CNN, we can minimize $\mathcal{L}(t,t^*;ACE)$ through back propagation of SGD variants. As a result, referring to Eq. (7), for this H. pylori segmentation, we have the following objective

$$\min_t(\mathcal{L}(t,t^*;ACE);SGD),$$

to abduct the segmentation model for better prediction $t$ compared with corresponding $t^*$.

## *5.3 LAF-based evaluation strategy*

To evaluate a solution for H. pylori segmentation, the key problem is the lack of accurate ground-truths due to the great difficulty of collecting noisy-free labels. Fortunately, via the one-step logical reasoning of OSAMTL (Section 5.2.2), we have abduced multiple targets that contain various information consistent with our prior knowledge in the knowledge base ($KB$). As a result, avoiding to collect noisy-free labels for evaluation we can narrate some logical facts from these abduced multiple targets and utilize the narrated logical facts to estimate the logical rationality of the outputs of a solution for H. pylori segmentation. Thus, following the formulas presented in section 3.2, we implement an evaluation strategy reasonable in the context of H. pylori segmentation. The outline of this instance implementation is shown in the bottom of Fig. 2. The proofs for the validation of the reasonings for this instance implementation are provided in Supplementary 2.

### 5.3.1 Narrated logical facts

Based on the abduced $Target_1$, we derive Reasoning 4. And based on the abduced $Target_2$, we derive Reasoning 5.

**Reasoning 4**. *Given $Target_1$, then pixels included in negative areas of $Target_1$ are most probably true H. pylori negatives.*

**Reasoning 5**. *Given $Target_2$, then pixels included in positive areas of $Target_2$ are most probably true H. pylori positives.*

Thus, referring to Eq. (8) and using Reasoning 4 as $param^{LFN}$, we can implement the $LogicalFactNarrate$ for logical fact narrated from $Target_1$. And referring to Eq. (8) and using Reasoning 5 as $param^{LFN}$, we can implement the $LogicalFactNarrate$ for logical fact narrated from $Target_2$. Finally, on the basis of these two implementations, we narrate two logical facts from $t^*$, as follows

$$LF = LogicalFactNarrate(t^*; \{Reasoning\ 4, Reasoning\ 5\})$$
$$= \begin{Bmatrix} LogicalFactNarrate(Target_1; \{Reasoning\ 4\}), \\ LogicalFactNarrate(Target_2; \{Reasoning\ 5\}) \end{Bmatrix}$$
$$= \{LF_1, LF_2\}.$$

Details of the narrated logical facts are provided in Table 5.

Table 5. Details of the narrated logical facts

| Narrated Logical Facts |
|---|
| $LF_1$: pixels included in negative areas of $Target_1$ are most probably true H. pylori negatives |
| $LF_2$: pixels included in positive areas of $Target_2$ are most probably true H. pylori positives |

### 5.3.2 Estimated logical consistencies

Based on the prediction $t$ of learning model and $LF_1$, we derive Reasoning 6.

**Reasoning 6**. *Given $t$ and $LF_1$, the intersection of pixels of $t$ that are predicted as H. pylori positives ($t^f$) and pixels of $Target_1$ that are considered as true H. pylori negatives ($Target_1^b$) can be considered as logically false positives.*

Thus, referring to Eq. (9) and using Reasoning 6 as $param^{LCE}$, we implement the $LogicalConsistencyEstimate$ for the logical consistency between $t$ and $LF_1$ as

$$LC_1 = LogicalConsistencyEstimate(t, LF_1; \{Reasoning\ 6\}).$$

Based on the prediction $t$ of learning model and $LF_2$, we derive Reasoning 7.

**Reasoning 7**. *Given $t$ and $LF_2$, the intersection of pixels of $t$ that are predicted as H. pylori positives ($t^f$) and pixels of $Target_2$ that are considered as true H. pylori positives ($Target_2^f$) can be considered as logically true positives, and the intersection of pixels of $t$ that are predicted as H. pylori negatives ($t^b$) and pixels of $Target_2$ that are considered as true H. pylori positives ($Target_2^f$) can be considered as logically false negatives.*

Thus, referring to Eq. (9) and using Reasoning 7 as $param^{LCE}$, we implement the $LogicalConsistencyEstimate$ for the logical consistency between $t$ and $LF_2$ as

$$[LC_2, LC_3] = LogicalConsistencyEstimate(t, LF_2; \{Reasoning\ 7\}).$$

Finally, on the basis of these implementations, we estimate three consistencies between $t$ and $LF$, as follows

$$LC = LogicalConsistencyEstimate(t, LF; \{Reasoning\ 6, Reasoning\ 7\})$$
$$= \begin{Bmatrix} LogicalConsistencyEstimate(t, LF_1; \{Reasoning\ 6\}), \\ LogicalConsistencyEstimate(t, LF_2; \{Reasoning\ 7\}) \end{Bmatrix}$$
$$= \{LC_1, LC_2, LC_3\}.$$

Details of the estimated logical consistencies are provided in Table 6.

Table 6. Details of the estimated logical consistencies

| Estimated Logical Consistencies |
| --- |
| $LC_1$: the intersection of $t^f$ and $Target_1^b$ can be considered as logically false positives |
| $LC_2$: the intersection of $t^f$ and $Target_2^f$ can be considered as logically true positives |
| $LC_3$: the intersection of $t^b$ and $Target_2^f$ can be considered as logically false positives |

**5.3.3 Built logical assessment metrics**

Based on the estimated $LC$, referring to Eq. (10) and using usual definitions for assessment of image semantic segmentation as $param^{LAM}$, we implement $LogicalAssessmentMetricBuild$ to form a series of logical assessment metrics as

$$LAM = LogicalAssessmentMetricBuild\left(LC;\begin{Bmatrix} TP, FP, FN, \\ precision, recall, \\ f1, fIoU \end{Bmatrix}\right)$$

$$= \left\{\begin{matrix} LAM_1: LTP = t^f \cap Target_2^f \\ LAM_2: LFP = t^f \cap Target_1^b \\ LAM_2: LFN = t^b \cap Target_2^f \\ LAM_4: Lprecision = \frac{LTP}{LTP+LFP} \\ LAM_5: Lrecall = \frac{LTP}{LTP+LFN} \\ LAM_6: Lf1 = \frac{2\times Lprecision\times Lrecall}{Lprecision+Lrecall} \\ LAM_7: LfIoU = \frac{LTP}{LTP+LFP+LFN} \end{matrix}\right\}.$$

# 6. Experiments and Analyses

On the basis of the OSAMTL-base image semantic segmentation solution and the LAF-based evaluation strategy implemented for the H. pylori segmentation task in MHWSIA, we conduct extensive experiments to investigate the effectiveness of OSAMTL in handling complex noisy labels.

***6.1 Overall design***

Due to the fact that accurate/noisy-free ground-truth labels for H. pylori segmentation is extremely difficult to collect, we employ logical assessment metrics build in Section 5.3 to evaluate the logical rationality of the predictions of a solution for H. pylori segmentation without a noisy-free dataset. The preparation for the training, validation and testing datasets can be found in section 6.2. Section 6.3 gives experimental settings for respective experiments. To show the basic effectiveness of OSAMTL in handling complex noisy labels, in section 6.4, we conducted and compared experiments that directly learn from the given complex noisy labels and utilize OSAMTL to handle the given complex noisy labels. To show the performances of OSAMTL versus various state-of-the-art approaches for learning from noisy labels, in section 6.5, we conducted and compared experiments that respectively employ various state-of-the-art approaches and utilize OSAMTL to handle the given complex noisy labels. To show the effectiveness of OSAMTL in improving the logical reasonability of various state-of-the-art approaches for learning from noisy labels, in section 6.6, we conducted experiments that respectively obtain improvements by the one-step logical reasoning of OSAMTL, the multi-target learning procedure of OSAMTL, and more potentials of OSAMTL. We also show opposing contributions between OSAMTL and StoA in handling complex noisy labels in section 6.7. Additionally, in section 6.8, we give summarizations about the conducted experiments. Finally, in section 6.9, we provide some qualitative example results of respective approaches as visualized proofs to show the effectiveness of OSAMTL. For all experiments, we first use a training dataset to learn segmentation models, then we use a validation dataset to select the model for testing, finally we compare experimental results of selected segmentation models which are averaged on the testing dataset.

***6.2 Dataset preparation***

We received a total of 205 histopathology slices from Peking University Third Hospital. These slices were further digitalized by a slice scanning system called PRECICE from UnicTech and resulted in 205 WSIs. We used 155 WSIs for training and validation, and 50 WSIs for testing. The size of each WSI was approximately $40{,}000 \times 40{,}000$ pixels (width × height) at 40× magnification. We cropped each WSI into image patches at 40× magnification. The size of each cropped image patch was at 512 × 512 pixels (width × height). As lower sizes can lose the morphologies of H. pylori and increase the ratio of samples that mostly contain foreground labels, we used the size 512 * 512 to remain the morphologies of H. pylori in the samples and avoid many samples that mostly contain foreground labels. In fact, increasing the ratio of samples that mostly contain foreground labels can make the machine learning model bias to the complex noisy labels which are more difficult to handle, leading to lower logical rationality. Moreover, the samples that mostly contain foreground labels can also cause bias evaluation for validation/testing since the logically false positive on them is close to zero according to the reasonings for our logical assessment formula. Blank background patches were further removed. Eventually, we obtained a total of 30132 images for training, 7533 images for validation, and 10432 images for testing. We first manually labelled the training, validation and testing datasets with polygon annotations. Then, for experiments on these datasets, both of the target1 and the target2 were extracted using the methods resulted by the logical reasoning in Section 5.2.

***6.3 Experimental setting***

All of our experiments were performed on an Intel core Xeon E5–2630 v4s with a memory capacity of 128GB and a NVIDIA GPU GTX 1080Ti. Our developing environment is based on Tensorflow 1.10.1 and Python 3.5. We started the training procedures of the constructed tiny deep CNN backbone with the same initialization and hyperparameters including Adam [37] selected as the optimizer, batch size set to 32, learning rate set to 0.0001, and online augmentations involving vertical and horizontal flips and random brightness. Various state-of-the-art approaches for learning from noisy labels, including Forward, Backward [5], Boost-Hard, Boost-Soft [6, 7], D2L [8], SCE [9], Peer [10], DT-Forward [24], and NCE-SCE [25], were chosen for experimental investigations, due to their flexibility to be applied to the situation where no clean dataset is available, the targeted object cannot be clearly defined, and any of the given labels cannot be confidently regarded as clean. Especially, NCE-SCE is a modification version of NCE-RCE from APL [25] and we chose NCE-SCE for better convergence. We respectively set the hyper parameters of these approaches as suggested by corresponding papers. In default, we set the weights for the multi-task learning procedure of OSAMTL to fifty-fifty in this H. pylori segmentation case, considering that both targets are equally important. But worth to mention, according to our previous study [12], more potentials of OSAMTL can be released by changing the weights for the multi-task learning procedure, which is also discussed in Section 6.6.3 in this paper.

### *6.4 Basic effectiveness of OSAMTL*

OSAMTL has two key techniques: 1) the one-step logical reasoning that abduces multiple targets; and 2) the multi-target learning procedure that imposes the abduced multiple targets upon machine learning to constrain the learning model. In this section, we conduct experiments to show their basic effectiveness. Referring to the logical assessment metrics built in Section 5.3.3, the experimental results are shown as Fig. 7-8 and the statistics corresponding to Fig. 7-8 are shown in Table 7-8.

#### 6.4.1 Basic effectiveness of one-step logical reasoning

To show the basic effectiveness of the one-step logical reasoning of OSAMTL, we conducted experiments that directly learn from each of the two targets abduced by the one-step logical reasoning. We denote BaseLine and OSAMTL(T2) are two solutions that respectively learn from $Target_1$ and $Target_2$ with full supervision. In fact, BaseLine is regarded as the solution that naively learn from the given noisy labels ($Target_1$). And OSAMTL(T2) is regarded as the solution that is enabled by the one-step logical reasoning which produces the additional $Target_2$. Thus, comparison between BaseLine and OSAMTL(T2) can reflect what basic effectiveness can the one-step logical reasoning bring. The comparison between BaseLine and OSAMTL(T2) is shown as the histograms in the top row of Fig. 7.

From the left of the top row of Fig. 7, we can note that BaseLine and OSAMTL(T2) perform very close in terms of LTP and LFN (BaseLine has a little bit higher LTP and a little bit lower LFN, while OSAMTL(T2) has a little bit lower LTP and a little bit higher LFN). However, the LFP of BaseLine is much higher than the LFP of OSAMTL(T2). The differences of BaseLine and OSAMTL(T2) in terms of LTP, LFP and LFN lead to that BaseLine has a much lower Lprecision and a little bit higher Lrecall while OSAMTL(T2) has a much higher Lprecision and a little bit lower Lrecall (see the middle of the top row of Fig. 7, which is accordant to our reasonings of the one-step logical reasoning). Eventually, the differences of BaseLine and OSAMTL(T2) in terms of Lprecision and Lrecall result in that OSAMTL(T2) remarkably outperforms BaseLine in terms of overall performances Lf1 and LfIoU (see the right of the top row of Fig. 7). Detailed performance improvements of OSAMTL(T2) compared with BaseLine are shown in the bottom row of Fig.7. These comparisons lead us to conclude that the one-step logical reasoning of OSAMTL provides effective information ($Target_2$) to enable logically more rationale solutions than naively learning from the given noisy labels ($Target_1$).

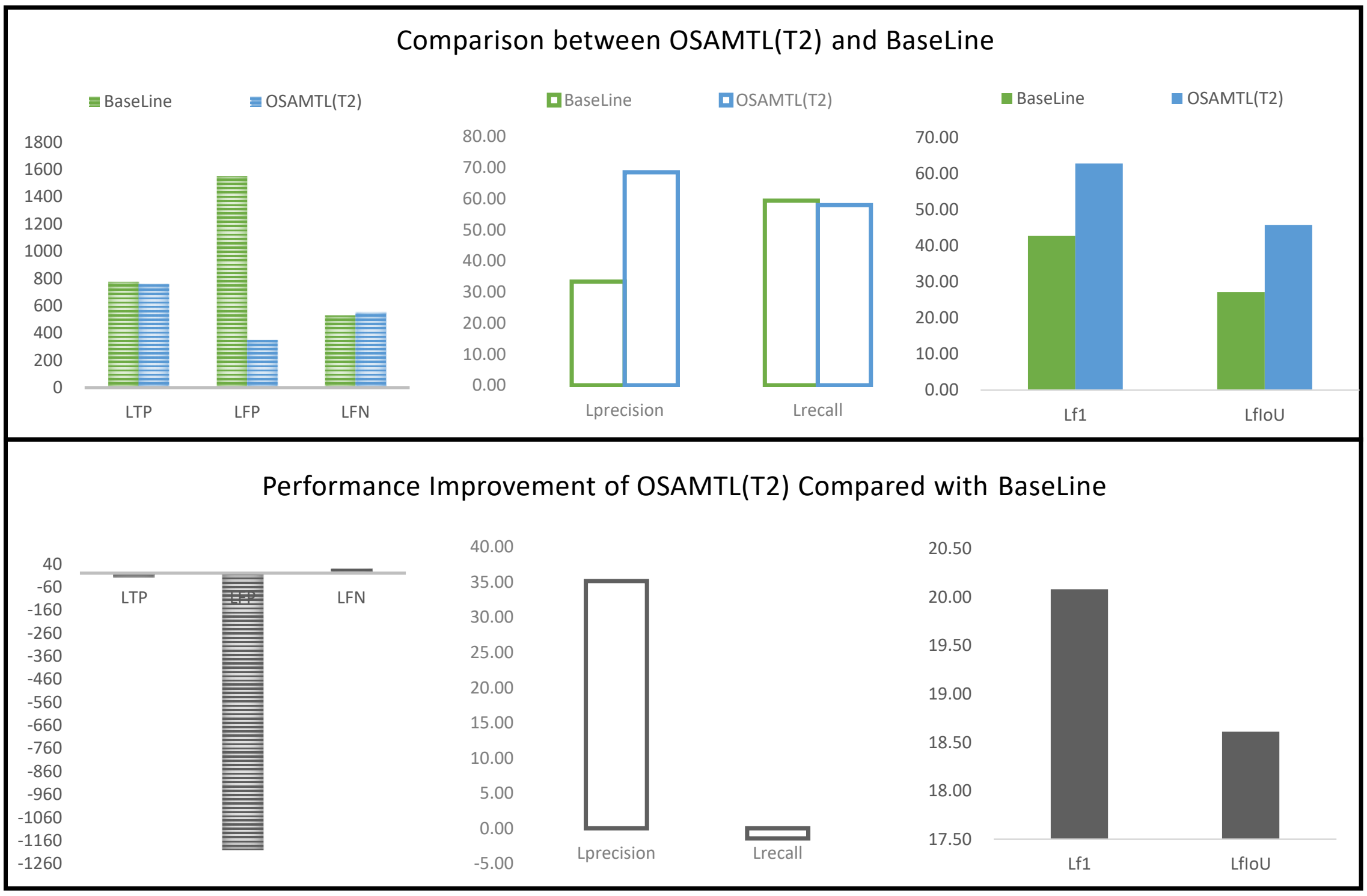

Fig.7 OSAMTL(T2) compared with BaseLine.

### 6.4.2 Basic effectiveness of multi-target learning procedure

To show the basic effectiveness of the multi-target learning procedure of OSAMTL, we conducted experiments that impose the two targets abduced by the one-step logical reasoning upon machine learning to constrain the learning model. We denote this solution as OSAMTL(T1&T2). As the multi-target learning procedure of OSAMTL leverages both $Target_1$ and $Target_2$, the comparison between OSAMTL(T1&T2) and BaseLine/OSAMTL(T2) can reflect what basic effectiveness can the multi-target learning procedure bring. The results are shown as Fig. 8.

From the top of Fig. 8, we can note that OSAMTL(T1&T2) has higher LTP and lower LFN than BaseLine/OSAMTL(T2), while being able to achieve a good balance in term of LFP between BaseLine and OSAMTL(T2) (overall comparison can be seen in the left of top row of Fig. 8 and more details can be seen in the left of middle and bottom rows of Fig. 8). These changes in terms of LTP, LFP and LFN lead to that OSAMTL(T1&T2) has a little bit higher Lprecison than OSAMTL(T2) and a Lrecall higher than BaseLine (see the middle of top, middle and bottom rows of Fig.8). This reflects that OSAMTL(T1&T2) makes up for respective shortcoming of BaseLine and OSAMTL(T2) and achieves even better results in terms of Lprecision and Lrecall.

Eventually, the achieved better results in terms of Lprecison and Lrecall result in that OSAMTL(T1&T2) significantly outperforms BaseLine/OSAMTL(T2) in terms of overall performances Lf1 and LfIoU (see the right of top, middle and bottom rows of Fig. 8). These comparisons lead us to conclude that the multi-target learning procedure of OSAMTL can make up for respective shortcoming of BaseLine and OSAMTL(T2) to achieve logically more rationale results than the best solution of BaseLine and OSAMTL(T2), further leveraging the effective information provided by the one-step logical reasoning to reach a new high.

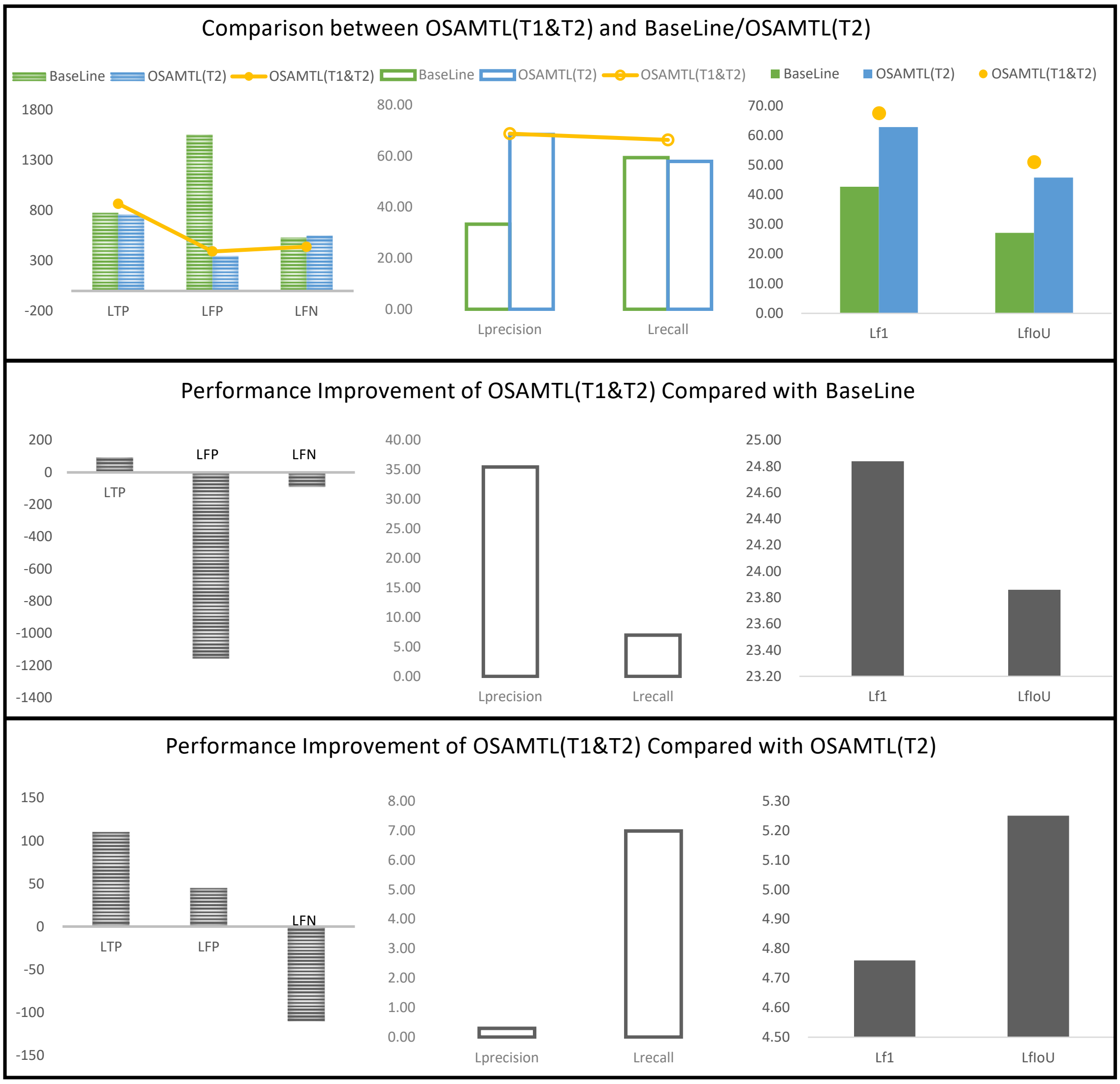

Fig. 8 OSAMTL(T1&T2) compared with BaseLine/OSAMTL(T2).

Table 7. Quantitative experimental results of BaseLine, OSAMTL(T2) and OSAMTL(T1&T2)

| Solution | LTP | LFP | LFN | Lprecision | Lrecall | Lf1 | LfIoU |
|---|---|---|---|---|---|---|---|
| BaseLine | 776 | 1550 | 530 | 33.35 | 59.39 | 42.72 | 27.16 |
| OSAMTL(T2) | 757 | 348 | 549 | 68.51 | 57.97 | 62.8 | 45.77 |
| OSAMTL(T1&T2) | 867 | 393 | 439 | 68.81 | 66.37 | **67.56** | **51.02** |

Table 8. Statistics of performance improvement of OSAMTL(T2) compared with BaseLine and OSAMTL(T1&T2) compared with OSAMTL(T2)/BaseLine

| A minus B | LTP | LFP | LFN | Lprecision | Lrecall | Lf1 | LfIoU |
|---|---|---|---|---|---|---|---|
| OSAMTL(T2)-BaseLine | -19 | -1202 | 19 | 35.16 | -1.42 | 20.08 | 18.61 |
| OSAMTL(T1&T2)-BaseLine | 91 | -1157 | -91 | 35.46 | 6.98 | **24.84** | **23.86** |
| OSAMTL(T1&T2)-OSAMTL(T2) | 110 | 45 | -110 | 0.30 | 8.40 | 4.76 | 5.25 |

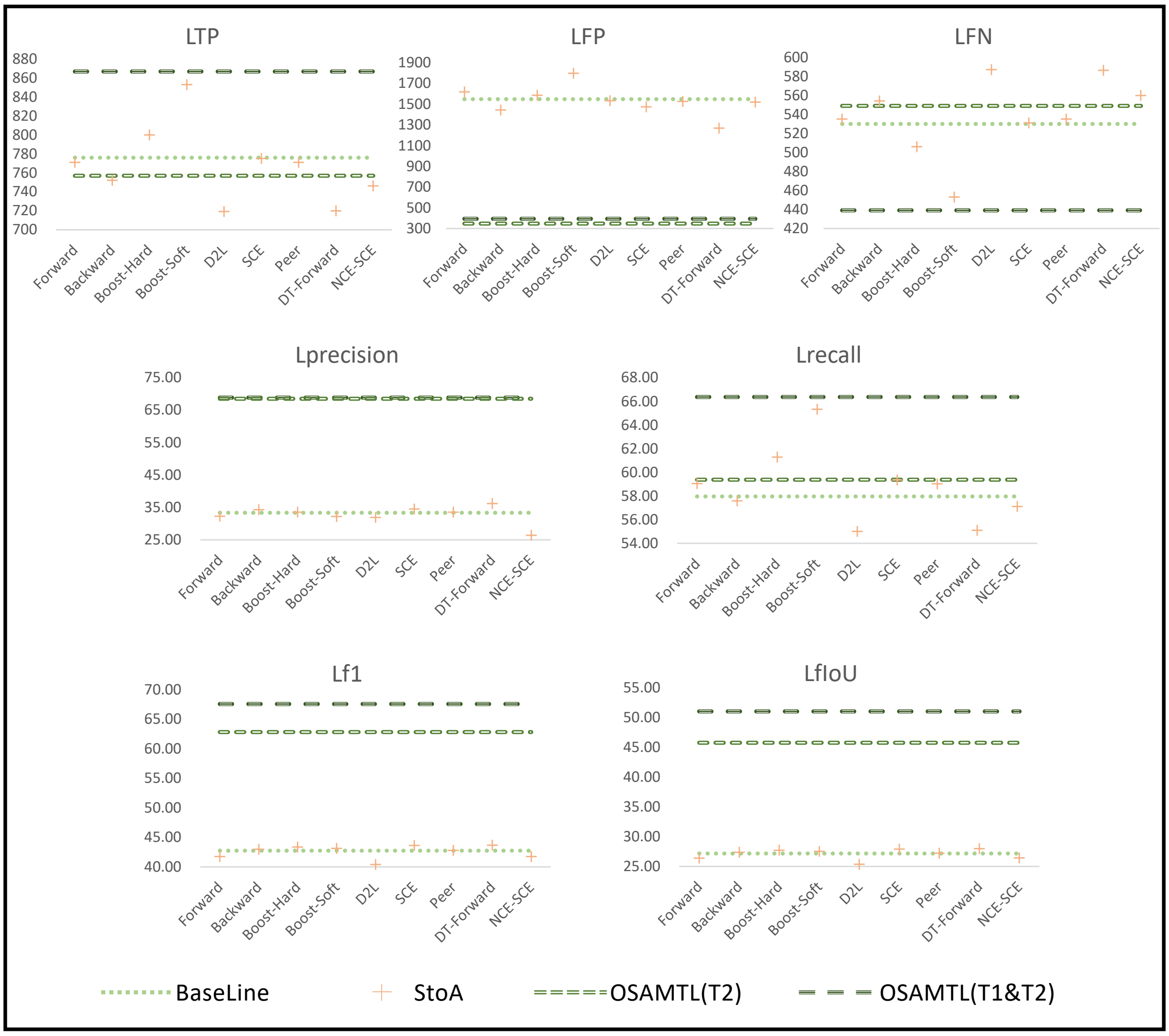

Fig. 9. Overall comparison between various state-of-the-art approaches and OSAMTL referring to BaseLine. These lines are drawn based on statistics in Table 2 and 3. StoA is short for state-of-the-art.

### *6.5 OSAMTL compared with state-of-the-art*

To show the effectiveness of OSAMTL compared with state-of-the-art approaches in handling complex noisy labels, we further conducted experiments that employ various state-of-the-art approaches, including Forward, Backward [5], Boost-Hard, Boost-Soft [6, 7], D2L [8], SCE [9], Peer [10], DT-Forward [24], and NCE-SCE [25] to respectively handle the given complex noisy labels ($Target_1$). We denote respective names of these state-of-the-art approaches as the solutions that employ these approaches to learn from the given complex noisy labels ($Target_1$). We first conduct an overall comparison, in which we respectively show the effectiveness of various state-of-the-art approaches and the effectiveness of OSAMTL in handling complex noise labels by comparing them with BaseLine. Then we conducted more detailed comparison to show some clues that probably lead to the performance differences between OSAMTL and various other state-of-the-art approaches in handling complex noisy labels. Referring to the logical assessment metrics built in section 5.3.3, the experimental results are shown as Fig. 9-10 and the statistics corresponding to Fig. 9-10 as Table 9-10.

Table 9. Quantitative experimental results of various state-of-the-art approaches

| StoA | LTP | LFP | LFN | Lprecision | Lrecall | Lf1 | LfIoU |
|---|---|---|---|---|---|---|---|
| Forward | 771 | 1618 | 535 | 32.29 | 59.06 | 41.75 | 26.38 |
| Backward | 752 | 1443 | 554 | 34.27 | 57.60 | 42.97 | 27.37 |
| Boost-Hard | 800 | 1585 | 506 | 33.55 | 61.29 | 43.36 | 27.68 |
| Boost-Soft | 853 | 1798 | 453 | 32.18 | 65.33 | 43.12 | 27.49 |
| D2L | 719 | 1532 | 587 | 31.93 | 55.02 | 40.40 | 25.32 |
| SCE | 775 | 1473 | 531 | 34.48 | 59.35 | 43.62 | 27.89 |
| Peer | 771 | 1527 | 535 | 33.55 | 59.04 | 42.78 | 27.21 |
| DT-Forward | 720 | 1269 | 586 | 36.19 | 55.11 | **43.69** | **27.95** |
| NCE-SCE | 746 | 1520 | 560 | 26.40 | 57.12 | 41.77 | 26.40 |

### 6.5.1 Overall comparison

The experimental results of various state-of-the-art approaches are shown as Table 9, and the overall comparison between various state-of-the-art approaches and the two solutions of OSAMTL referring to BaseLine are shown as Fig. 9. From Fig. 9, by comparing in terms of Lf1 and LfIoU, we can note that Backward, Boost-Hard, Boost-Soft, SCE, Peer, DT-Forward and NCE-SCE perform very close to BaseLine while the two solutions of OSAMTL make significant improvements. This overall comparison leads us to conclude that these state-of-the-art approaches indeed have limitations while OSAMTL show significant potentials in handling complex noisy labels.

### 6.5.2 Detailed comparison

Regarding BaseLine as the reference substance, the detailed indirect comparison between OSAMTL and various state-of-the-art approaches, and the performance improvements of OSAMTL and various state-of-the-art approaches compared with BaseLine are shown as Fig. 10 and Table 10. From the Fig. 10, we can note that, compared with BaseLine, various state-of-the-art approaches have the tendencies to slightly fluctuate around LTP, LFP and LFN while the two solutions of OSAMTL have the tendency to increase LTP while significantly reduce LFP and reduce LFN (see the left subgraph of the top row of Fig. 10 and the top row of the bottom of Fig. 10). This comparison shows that various state-of-the-art approaches can only make small fluctuated changes compared with BaseLine while the two solutions of OSAMTL are able to make destructive changes, compared with BaseLine. This comparison also shows that the changes made by various state-of-the-art approaches and the changes made by the two solutions of OSAMTL have the tendency to be the opposite. These differences result in that the two solutions of OSAMTL can remarkably outperform BaseLine in terms of Lprecision and Lrecall while various state-of-the-art approaches can only fluctuate around the Lprecision and Lrecall of BaseLine (see the middle subgraph of the top row of Fig. 10 and the left subgraph of the bottom row of the bottom of Fig. 10). Eventually, compared with BaseLine, the two solutions of OSAMTL are able to achieve remarkable improvements in terms of Lf1 and LfIoU while various state-of-the-art approaches can only achieve small fluctuated improvements (see the right of the top row of Fig. 10 and the right of the bottom row of the bottom of Fig. 10). As well, the detailed direct performance improvements of OSAMTL(T2) and OSAMTL(T1&T2) compared with various state-of-the-art approaches are shown as Fig. 11-12 and Table 11-12. From Fig. 11-12 and Table 11-12, we can clearly observe that OSAMTL possesses unique capability to achieve logically more rational predictions than various state-of-the-art approaches when handling complex noisy labels.

These detailed comparisons lead us to further confirm that OSAMTL have remarkably effective potentials while it is not a good idea to directly employ these state-of-the-art approaches to handle complex noisy labels ($Target_1$) which affirms the limitations of these state-of-the-art approaches. The probable reason for this phenomenon lies in that the pre-assumptions of these state-of-the-art approaches are much less flexible than the pre-assumption of OSAMTL, that is the true target can be approximated by multiple inaccurate targets that contain information consistent with our prior knowledge about the true target. As a result, the capability of OSAMTL to achieve logically more rational predictions is beyond these state-of-the-art approaches in handling complex noisy labels ($Target_1$).

### *6.6 Effectiveness of OSAMTL in improving state-of-the-art*

As shown in Section 6.5 that OSAMTL possesses unique capability to achieve logically more rational predictions than various state-of-the-art approaches when handling complex noisy labels, it is probable that we can introduce OSAMTL to these state-of-the-art approaches for performance improvement. Thus, in this section, we conduct experiments that introduce OSAMTL to various state-of-the-art approaches. We denote OSAMTL_StoA as the solution that introduces OSAMTL to a StoA approach (StoA symbolizes the name of a state-of-the-art approach). We compare the performance of OSAMTL_StoA with the performance of StoA, which does not have OSAMTL introduced, to show the effectiveness of OSAMTL in improving the logical reasonability of various state-of-the-art approaches. There are two basic strategies to introduce OSAMTL to StoA: 1) introducing the one-step logical reasoning of OSAMTL to StoA, and 2) introducing the multi-target learning procedure of OSAMTL to StoA. We denote the solution of the strategy 1) as OSAMTL_StoA(T2) and the solution of the strategy 2) as OSAMTL_StoA(T1&T2). Specifically, OSAMTL_StoA(T2) is regarded as the solution that employs the StoA approach to handle noisy labels $Target_2$ abduced by the one-step logical reasoning of OSAMTL, and OSAMTL_StoA(T1&T2) is regarded as the solution that further combines StoA with the multi-target learning procedure of OSAMTL to learning from both noisy labels $Target_1$ and $Target_2$. Additionally, since our previous work [12] has shown that changing the weights for the multi-task learning procedure of OSAMTL can release more potentials, we as well discuss about leveraging this property of OSAMTL for performance improvements.

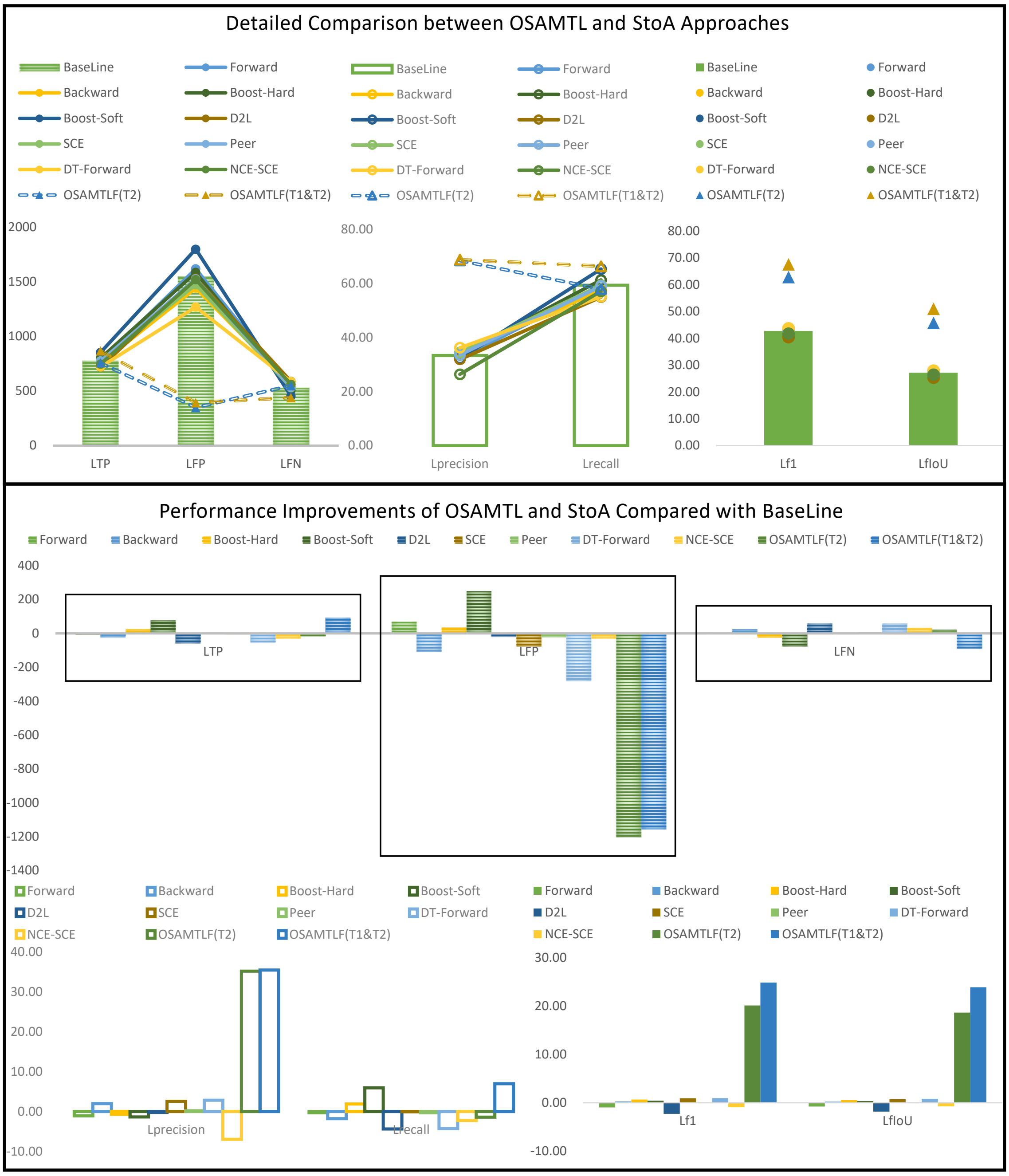

Fig. 10. Detailed comparison between OSAMTL and StoA, and performance improvements of OSAMTL and StoA compared with BaseLine. StoA is short for state-of-the-art.

Table 10. Statistics corresponding to the performance improvements of OSAMTL and state-of-the-art approaches compared with BaseLine shown in Fig. 10

| OSAMTL/StoA minus BaseLine | LTP | LFP | LFN | Lprecision | Lrecall | Lf1 | LfIoU |
|---|---|---|---|---|---|---|---|
| Forward | -5 | 68 | 5 | -1.06 | -0.33 | -0.97 | -0.78 |
| Backward | -24 | -107 | 24 | 1.98 | -1.79 | 0.25 | 0.21 |
| Boost-Hard | 24 | 35 | -24 | -0.72 | 1.9 | 0.64 | 0.52 |
| Boost-Soft | 77 | 248 | -77 | -1.37 | 5.94 | 0.40 | 0.33 |
| D2L | -57 | -18 | 57 | -0.25 | -4.37 | -2.32 | -1.84 |
| SCE | -1 | -77 | 1 | 2.55 | -0.04 | 0.90 | 0.73 |
| Peer | -5 | -23 | 5 | 0.20 | -0.35 | 0.06 | 0.05 |
| DT-Forward | -56 | -281 | 56 | 2.84 | -4.28 | 0.97 | 0.79 |
| NCE-SCE | -30 | -30 | 30 | -6.95 | -2.27 | -0.95 | -0.76 |
| OSAMTL(T2) | -19 | -1202 | 19 | 35.16 | -1.42 | 20.08 | 18.61 |
| OSAMTL(T1&T2) | 91 | -1157 | -91 | 35.46 | 6.98 | **24.84** | **23.86** |

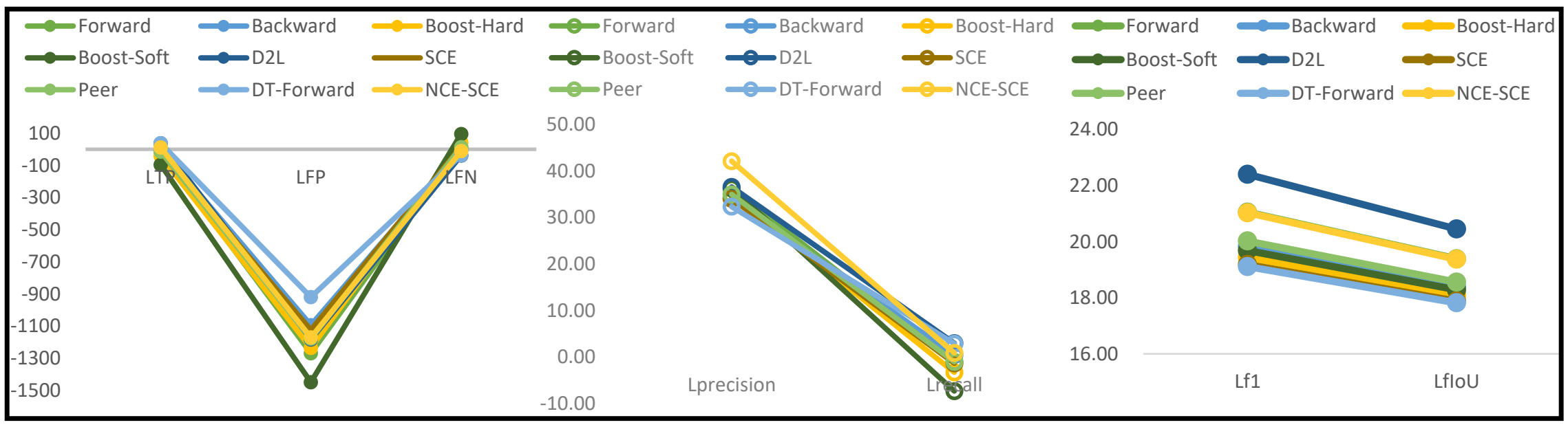

Fig. 11. Performance improvement of OSAMTL(T2) compared with StoA.

Table 11. Statistics corresponding to Fig. 11

| OSAMTL(T2) minus StoA | LTP | LFP | LFN | Lprecision | Lrecall | Lf1 | LfIoU |
|---|---|---|---|---|---|---|---|
| Forward | -14 | -1270 | 14 | 36.22 | -1.09 | 21.05 | 19.39 |
| Backward | 5 | -1095 | -5 | 34.24 | 0.37 | 19.83 | 18.40 |
| Boost-Hard | -43 | -1237 | 43 | 34.96 | -3.32 | 19.44 | 18.09 |
| Boost-Soft | -96 | -1450 | 96 | 36.33 | -7.36 | 19.68 | 18.28 |
| D2L | 38 | -1184 | -38 | 36.58 | 2.95 | **22.40** | **20.45** |
| SCE | -18 | -1125 | 18 | 34.03 | -1.38 | 19.18 | 17.88 |
| Peer | -14 | -1179 | 14 | 34.96 | -1.07 | 20.02 | 18.56 |
| DT-Forward | 37 | -921 | -37 | 32.32 | 2.86 | 19.11 | 17.82 |
| NCE-SCE | 11 | -1172 | -11 | 42.11 | 0.85 | 21.03 | 19.37 |

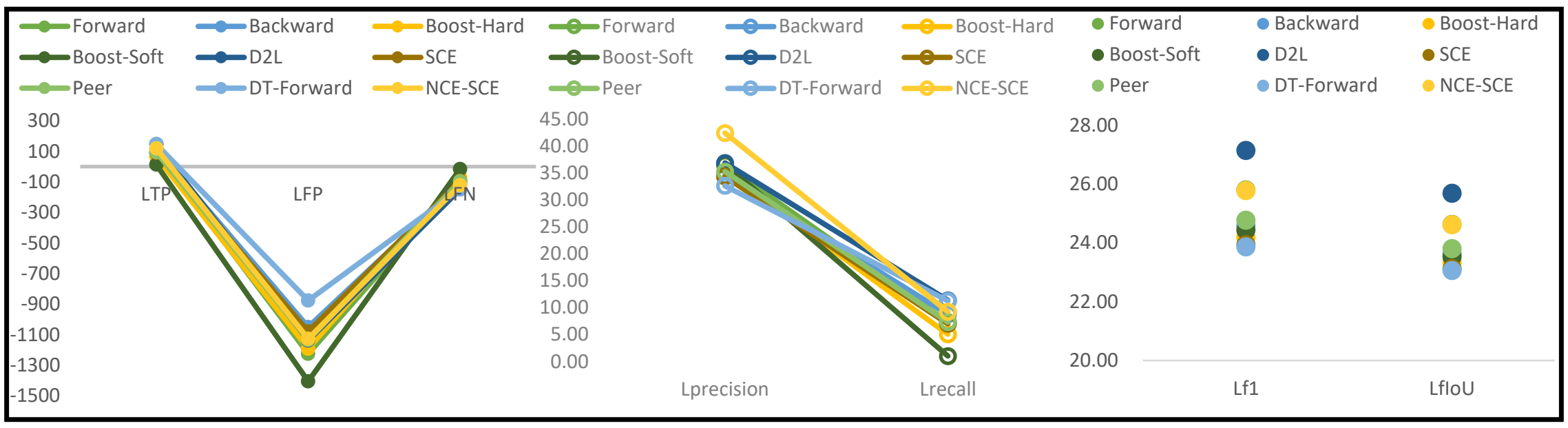

Fig. 12. Performance improvement of OSAMTL(T1&T2) compared with StoA.

Table 12. Statistics corresponding to Fig. 12

| OSAMTL(T1&T2) minus StoA | LTP | LFP | LFN | Lprecision | Lrecall | Lf1 | LfIoU |
|---|---|---|---|---|---|---|---|
| Forward | 96 | -1225 | -96 | 36.52 | 7.31 | 25.81 | 24.64 |
| Backward | 115 | -1050 | -115 | 34.54 | 8.77 | 24.59 | 23.65 |
| Boost-Hard | 67 | -1192 | -67 | 35.26 | 5.08 | 24.2 | 23.34 |
| Boost-Soft | 14 | -1405 | -14 | 36.63 | 1.04 | 24.44 | 23.53 |
| D2L | 148 | -1139 | -148 | 36.88 | 11.35 | **27.16** | **25.7** |
| SCE | 92 | -1080 | -92 | 34.33 | 7.02 | 23.94 | 23.13 |
| Peer | 96 | -1134 | -96 | 35.26 | 7.33 | 24.78 | 23.81 |
| DT-Forward | 147 | -876 | -147 | 32.62 | 11.26 | 23.87 | 23.07 |
| NCE-SCE | 121 | -1127 | -121 | 42.41 | 9.25 | 25.79 | 24.62 |

### 6.6.1 Improvement by one-step logical reasoning

The quantitative experimental results of OSAMTL_StoA(T2) are shown as Table 13, and the overall comparison between OSAMTL_StoA(T2) and StoA/OSAMTL(T2) is shown as Fig. 13. More specifically, the detailed comparison between OSAMTL_StoA(T2) and StoA is shown as Fig. 14, and the detailed comparison between OSAMTL_StoA(T2) and OSAMTL(T2) is shown as Fig. 15. The statistics corresponding to Fig. 14-15 are respectively shown in Table 14-15.

Table. 13 Quantitative experimental results of OSAMTL_StoA(T2)

| OSAMTL_StoA(T2) | LTP | LFP | LFN | Lprecision | Lrecall | Lf1 | LfIoU |
|---|---|---|---|---|---|---|---|
| Forward | 762 | 188 | 544 | 80.21 | 58.37 | 67.57 | 51.02 |
| Backward | 733 | 169 | 573 | 81.29 | 56.16 | 66.41 | 49.73 |
| Boost-Hard | 690 | 81 | 616 | 89.49 | 52.80 | 66.41 | 49.72 |
| Boost-Soft | 747 | 178 | 559 | 80.75 | 57.21 | 66.98 | 50.35 |
| D2L | 698 | 136 | 608 | 83.71 | 53.47 | 65.25 | 48.43 |
| SCE | 718 | 151 | 588 | 82.60 | 54.99 | 66.03 | 49.29 |
| Peer | 797 | 247 | 509 | 76.38 | 61.03 | 67.84 | 51.34 |
| DT-Forward | 816 | 259 | 490 | 75.91 | 62.49 | 68.55 | 52.15 |
| NCE-SCE | 799 | 190 | 507 | 80.76 | 61.21 | **69.64** | **53.42** |

Overall speaking, from Fig. 13, we can observe that OSAMTL_StoA(T2) outperforms both StoA and OSAMTL(T2) via introducing the one-step logical reasoning of OSAMTL to StoA. More specifically, from Fig. 14 and Table 14, we can note that OSAMTL_StoA(T2) significantly reduces LFP while makes relatively small fluctuations in terms of LTP and LFN compared with StoA. These changes made by OSAMTL_StoA(T2) remarkably improve Lprecision while slightly fluctuate in Lrecall, resulting in remarkable improvements in terms of Lf1 and LfIoU compared with StoA. More specifically, from Fig. 15 and Table 15, we can also observe that OSAMTL_StoA(T2) can make obvious fluctuations in terms of LTP, LFP and LFN compared with OSAMTL(T2). These changes made by OSAMTL_StoA(T2) significantly improve Lprecision while slightly fluctuate in Lrecall, resulting in clear improvements in terms of Lf1 and LfIoU compared with OSAMTL(T2). Notably, comparing the performance improvements of OSAMTL_StoA(T2) against OSAMTL(T2) with the performance improvements of StoA against BaseLine (statistics in Table 15 versus corresponding statistics in Table 10, we can observe that various StoA approaches achieve better improvements on $Target_2$ than on $Target_1$.

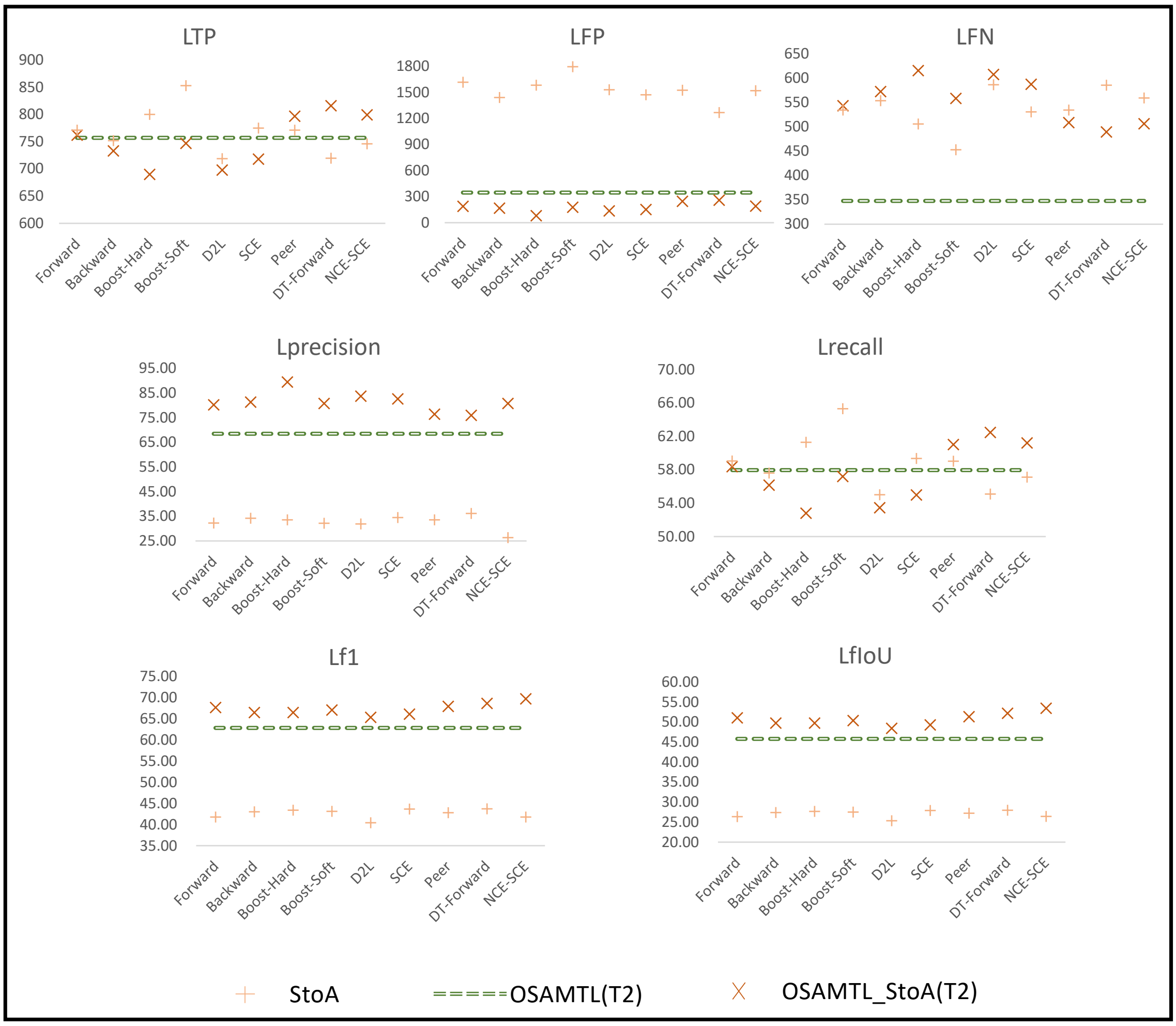

Fig. 13. Comparison between OSAMTL_StoA(T2) and StoA/OSAMTL(T2).

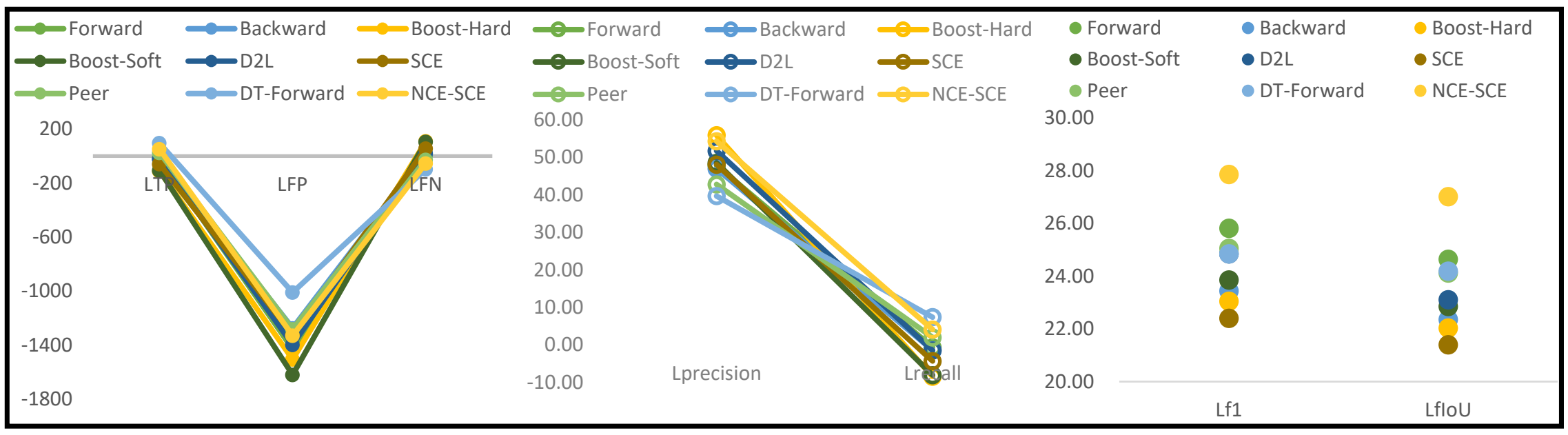

Fig. 14. Performance improvement of OSAMTL_StoA(T2) compared with StoA.

Table 14. Statistics corresponding to Fig. 14

| OSAMTL_StoA(T2) minus StoA | LTP | LFP | LFN | Lprecision | Lrecall | Lf1 | LfIoU |
|---|---|---|---|---|---|---|---|
| Forward | -9 | -1430 | 9 | 47.92 | -0.69 | 25.82 | 24.64 |
| Backward | -19 | -1274 | 19 | 47.02 | -1.44 | 23.44 | 22.36 |
| Boost-Hard | -110 | -1504 | 110 | 55.94 | -8.49 | 23.05 | 22.04 |
| Boost-Soft | -106 | -1620 | 106 | 48.57 | -8.12 | 23.86 | 22.86 |
| D2L | -21 | -1396 | 21 | 51.78 | -1.55 | 24.85 | 23.11 |
| SCE | -57 | -1322 | 57 | 48.12 | -4.36 | 22.41 | 21.40 |
| Peer | 26 | -1281 | -26 | 42.83 | 1.99 | 25.06 | 24.12 |
| DT-Forward | 96 | -1010 | -96 | 39.72 | 7.37 | 24.86 | 24.19 |
| NCE-SCE | 53 | -1330 | -53 | 54.36 | 4.09 | **27.87** | **27.02** |

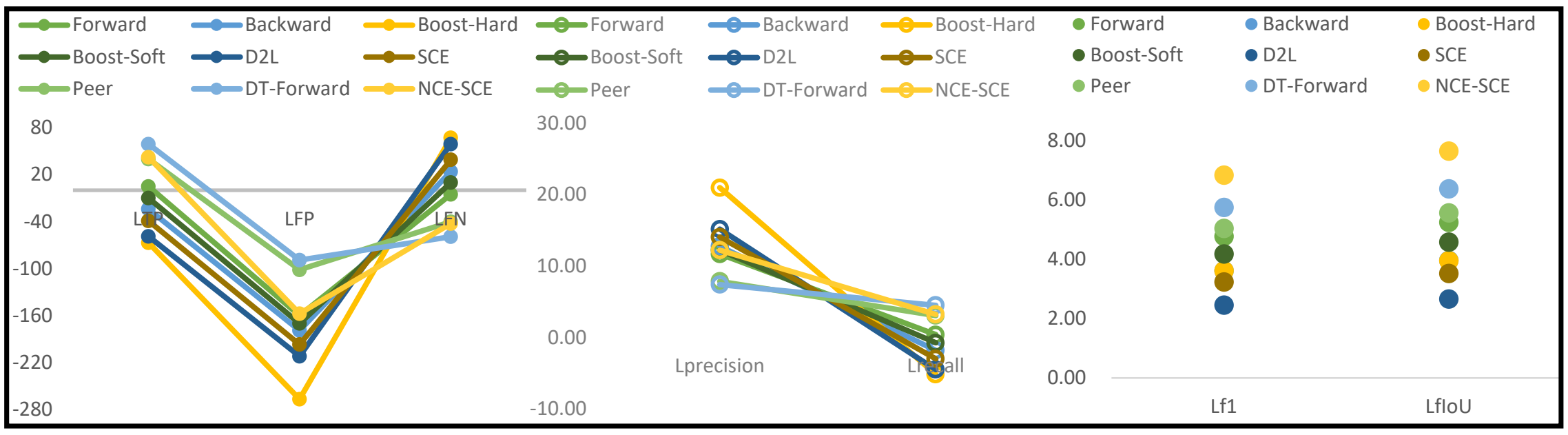

Fig. 15. Performance improvement of OSAMTL_StoA(T2) compared with OSAMTL(T2).

Table 15. Statistics corresponding to Fig. 15

| OSAMTL_StoA(T2) minus OSAMTL(T2) | LTP | LFP | LFN | Lprecision | Lrecall | Lf1 | LfIoU |
|---|---|---|---|---|---|---|---|
| Forward | 5 | -160 | -5 | 11.70 | 0.40 | 4.77 | 5.25 |
| Backward | -24 | -179 | 24 | 12.78 | -1.81 | 3.61 | 3.96 |
| Boost-Hard | -67 | -267 | 67 | 20.98 | -5.17 | 3.61 | 3.95 |
| Boost-Soft | -10 | -170 | 10 | 12.24 | -0.76 | 4.18 | 4.58 |
| D2L | -59 | -212 | 59 | 15.20 | -4.50 | 2.45 | 2.66 |
| SCE | -39 | -197 | 39 | 14.09 | -2.98 | 3.23 | 3.52 |
| Peer | 40 | -101 | -40 | 7.87 | 3.06 | 5.04 | 5.56 |
| DT-Forward | 59 | -89 | -59 | 7.40 | 4.52 | 5.75 | 6.37 |
| NCE-SCE | 42 | -158 | -42 | 12.25 | 3.24 | **6.84** | **7.65** |

These comparisons lead us to conclude: 1) The one-step logical reasoning of OSAMTL can provide noisy labels ($Target_2$) better than the given noisy labels ($Target_1$) for various StoA approaches to achieve logically more rationale predictions; 2) The one-step logical reasoning of OSAMTL releases more potentials of various StoA approaches for handling noisy labels by providing noisy labels ($Target_2$) better than the given noisy labels ($Target_1$), and the referred various StoA approaches have better capabilities in handling less complex noisy labels ($Target_2$, with high precision and low recall of the true target ) than complex noisy labels ($Target_1$, with high recall and low precision of the true target).

### 6.6.2 Improvement by multi-target learning procedure

The quantitative experimental results of OSAMTL_StoA(T1&T2) are shown as Table 16, and the overall comparison between OSAMTL_StoA(T1&T2) and OSAMTL_StoA(T2) is shown as Fig. 16. More specifically, the detailed comparison between OSAMTL_StoA(T1&T2) and OSAMTL_StoA(T2) shown as Fig. 17, and the statistics corresponding to Fig. 17 are shown in Table 17.

Table 16. Quantitative experimental results of OSAMTL_StoA(T1&T2)

| OSAMTL_StoA(T1&T2) | LTP | LFP | LFN | Lprecision | Lrecall | Lf1 | LfIoU |
|---|---|---|---|---|---|---|---|
| Forward | 797 | 204 | 509 | 79.69 | 61.02 | 69.09 | 52.78 |
| Backward | 786 | 221 | 520 | 78.04 | 60.19 | 67.96 | 51.47 |
| Boost-Hard | 758 | 136 | 548 | 84.8 | 58.01 | 68.89 | 52.55 |
| Boost-Soft | 837 | 309 | 468 | 73.01 | 64.08 | 68.25 | 51.80 |
| D2L | 758 | 166 | 548 | 82.01 | 58.02 | 67.96 | 51.47 |
| SCE | 822 | 288 | 484 | 74.05 | 62.97 | 68.06 | 51.59 |
| Peer | 749 | 358 | 557 | 67.67 | 57.38 | 62.1 | 45.04 |
| DT-Forward | 790 | 188 | 516 | 80.75 | 60.51 | 69.18 | 52.88 |
| NCE-SCE | 798 | 202 | 508 | 79.81 | 61.10 | **69.21** | **52.92** |

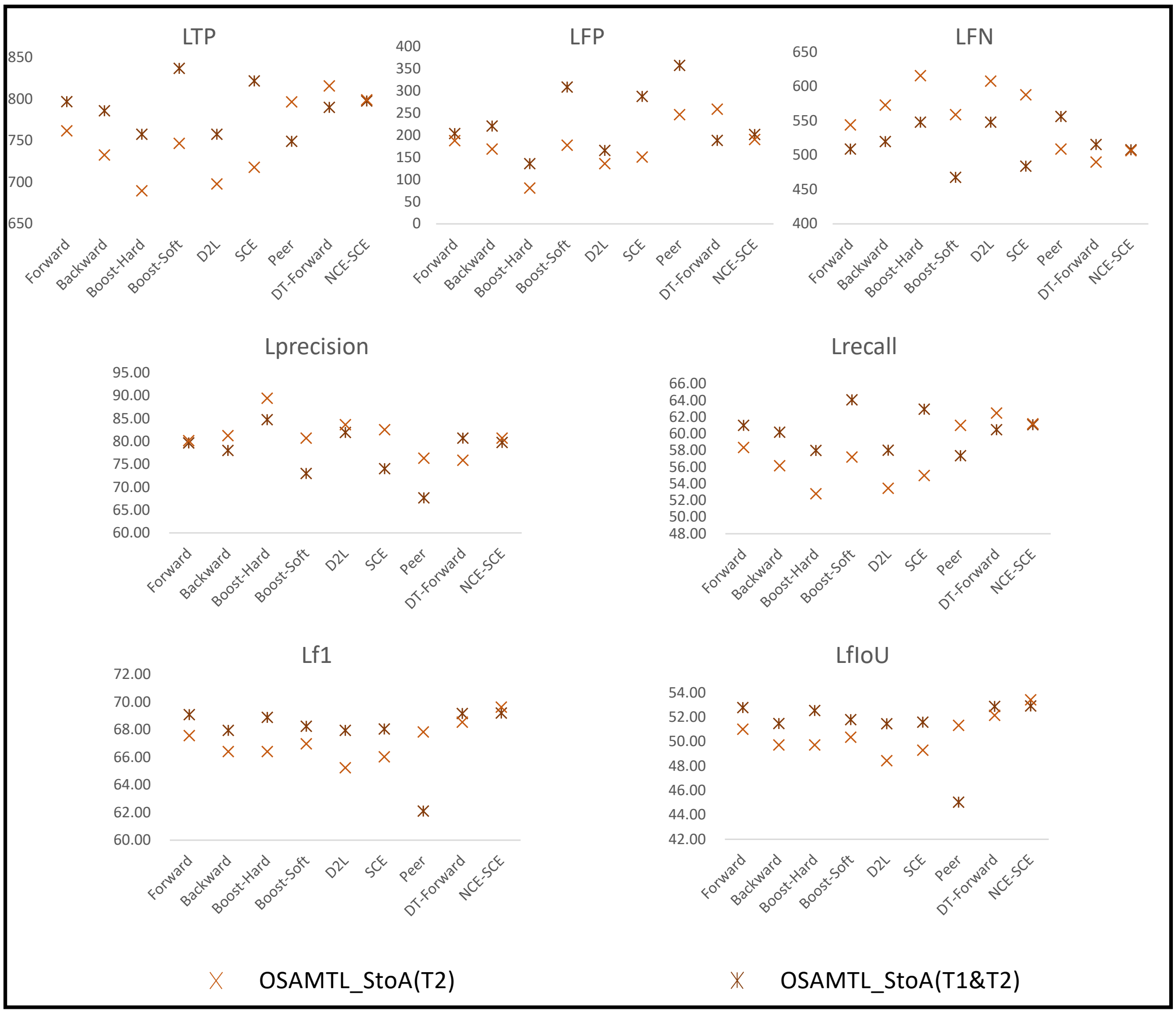

Fig. 16. Comparison between OSAMTL_StoA(T1&T2) and OSAMTL_StoA(T2).

From Fig. 16, we can observe that OSAMTL_StoA(T1&T2) can further improve the performances for the majority of OSAMTL_StoA(T2) solutions by introducing the multi-target learning procedure of OSAMTL into the referred StoA approaches. From Fig. 17 and Table 17, we can note that those cases where the OSAMTL_StoA(T1&T2) solution achieves better performance compared with the OSAMTL_StoA(T2) solution have good trade-offs in terms of LTP, LFP and LFN, leading to positive performance improvements in terms of Lf1 and LfIoU. These comparisons lead us to conclude that, for the majority of the referred StoA approaches, the solution OSAMTL_StoA(T1&T2) with the introduced multi-target learning procedure of OSAMTL can further achieve performance better than the solution OSAMTL_StoA(T2) with the introduced one-step logical reasoning of OSAMTL to reach a new high.

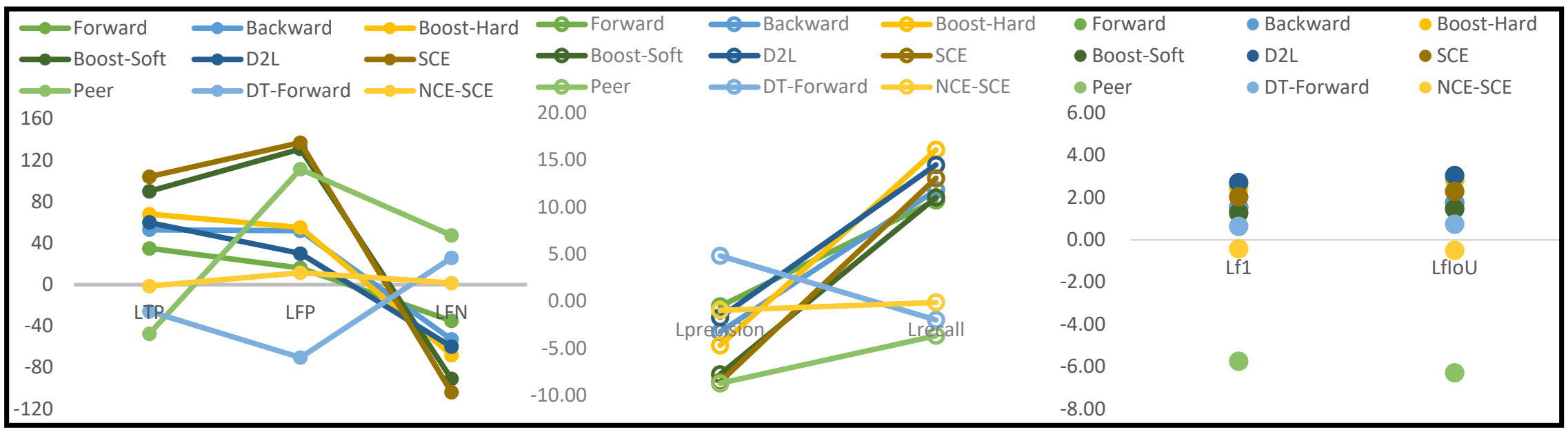

Fig. 17. Performance improvement of OSAMTL_StoA(T1&T2) compared with OSAMTL_StoA(T2).

Table 17. Statistics corresponding to Fig. 17

| OSAMTL_StoA(T1&T2) minus OSAMTL_StoA(T2) | LTP | LFP | LFN | Lprecision | Lrecall | Lf1 | LfIoU |
|---|---|---|---|---|---|---|---|
| Forward | 35 | 16 | -35 | -0.52 | 10.72 | 1.52 | 1.76 |
| Backward | 53 | 52 | -53 | -3.25 | 11.80 | 1.55 | 1.74 |
| Boost-Hard | 68 | 55 | -68 | -4.69 | 16.09 | 2.48 | 2.83 |
| Boost-Soft | 90 | 131 | -91 | -7.74 | 11.04 | 1.27 | 1.45 |
| D2L | 60 | 30 | -60 | -1.70 | 14.49 | **2.71** | **3.04** |
| SCE | 104 | 137 | -104 | -8.55 | 13.07 | 2.03 | 2.30 |
| Peer | -48 | 111 | 48 | -8.71 | -3.65 | -5.74 | -6.30 |
| DT-Forward | -26 | -71 | 26 | 4.84 | -1.97 | 0.63 | 0.74 |
| NCE-SCE | -2 | 11 | 2 | -0.95 | -0.11 | -0.43 | -0.50 |

Notably, form Fig. 16, we can also observe that Peer and NCE-SCE are two exceptional cases, for their OSAMTL_StoA(T1&T2) solutions perform less competitively than their OSAMTL_StoA(T2) solutions. From Fig. 17 and Table 17, we can notice the two exceptional Peer and NCE-SCE cases have not so well trade-offs in terms of LTP, LFP and LFN, leading to negative performance improvements in terms of Lf1 and LfIoU. But worth to mention, our previous work [12] has shown that more potentials of OSAMTL can be released by changing the weights for its multi-task learning procedure. Thus, we conducted more experiments in the next subsection to investigate whether the two exceptional cases can be eliminated by changing the weights for the multi-task learning procedure of OSAMTL.

### 6.6.3 Improvement by more potentials of OSAMTL

In this subsection, we conducted experiments to investigate whether the performances for the OSAMTL_StoA(T1&T2) solutions of Peer and NCE-SCE can be improved by changing the weights for the multi-target learning procedure which can release more potentials of OSAMTL [12]. Referring to our previous work [12], we first tried to change the weights for the multi-target learning procedure of OSAMTL to 0.4 for $Target_1$ and 0.6 for $Target_2$, and denote this solution as OSAMTL_StoA_46(T1&T2). The quantitative experimental results of the OSAMTL_StoA_46(T1&T2) solutions based on Peer and NCE-SCE are shown as Table 18, and corresponding comparison between OSAMTL_StoA_46(T1&T2) and OSAMTL_StoA(T2) is shown as Fig. 18. More specifically, the detailed comparison between OSAMTL_StoA_46(T1&T2) and OSAMTL_StoA(T2) shown as Fig. 19, and the statistics corresponding to Fig. 19 are shown in Table 19. From Table 18-19 and Fig. 18-19, we are fortunate and surprised to observe that the two OSAMTL_StoA_46(T1&T2) solutions achieve better performances than the two OSAMTL_StoA(T2) solutions as expected. These comparisons lead us to conclude that more potentials of OSAMTL can as well be released for the performance improvement of StoA by changing the weights for its multi-target learning procedure.

Table 18. Quantitative experimental results of OSAMTL_StoA_46(T1&T2)

| OSAMTL_StoA_46(T1&T2) | LTP | LFP | LFN | Lprecision | Lrecall | Lf1 | LfIoU |
|---|---|---|---|---|---|---|---|
| Peer | 760 | 159 | 546 | 82.66 | 58.18 | 68.29 | 51.85 |
| NCE-SCE | 938 | 403 | 368 | 69.92 | 71.81 | **70.85** | **54.86** |

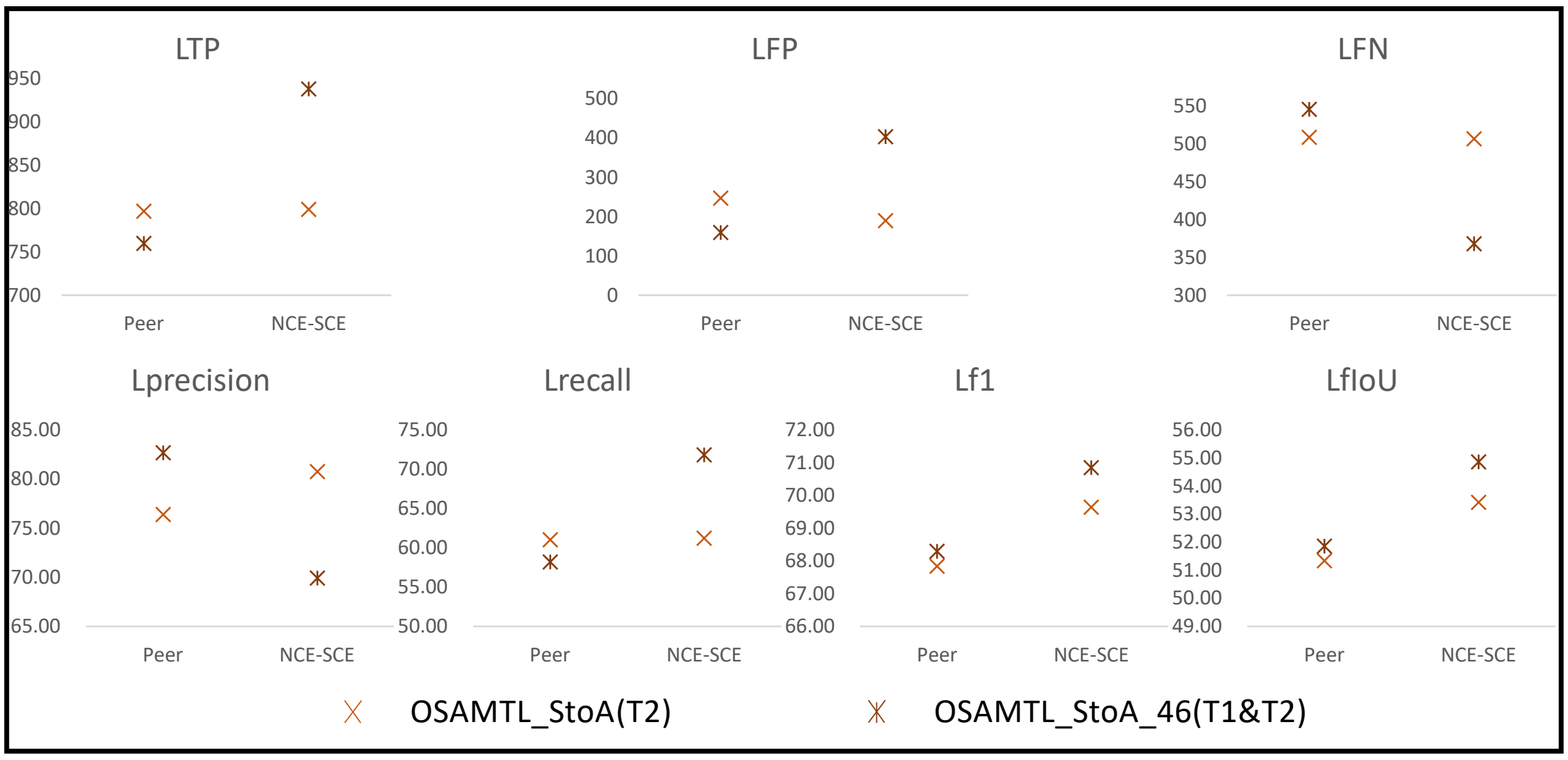

Fig. 18. Comparison between OSAMTL_StoA_46(T1&T2) and OSAMTL_StoA(T2).

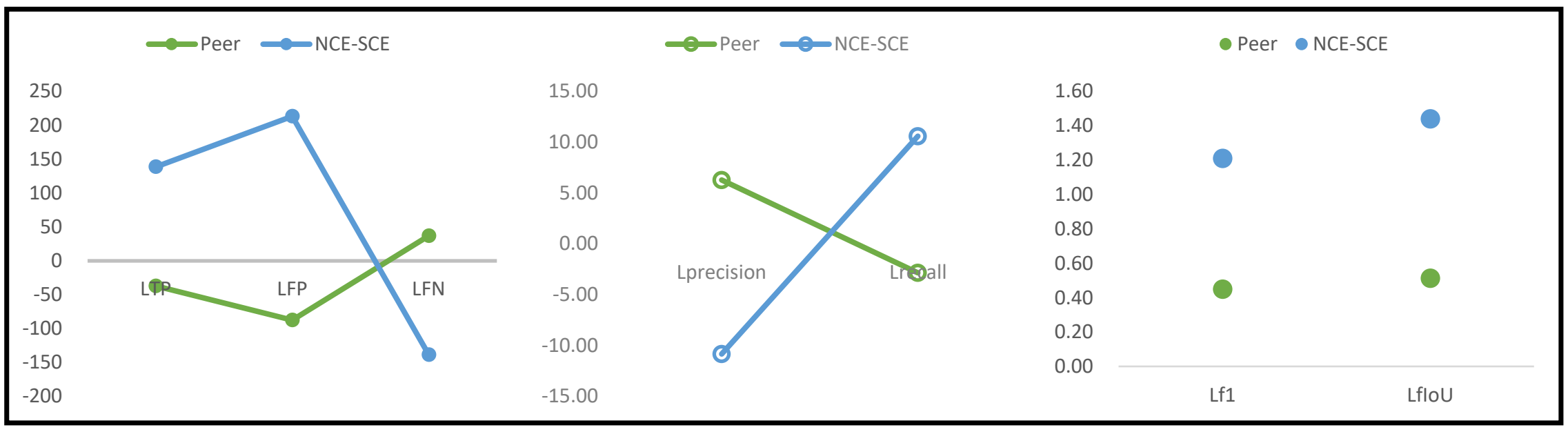

Fig. 19. Performance improvement of OSAMTL_StoA_46(T1&T2) compared with OSAMTL(T2).

Table 19. Statistics corresponding to Fig. 19

| OSAMTL_StoA_46(T1&T2) minus OSAMTL_StoA(T2) | LTP | LFP | LFN | Lprecision | Lrecall | Lf1 | LfIoU |
|---|---|---|---|---|---|---|---|
| Peer | -37 | -88 | 37 | 6.28 | -2.85 | 0.45 | 0.51 |
| NCE-SCE | 139 | 213 | -139 | -10.84 | 10.60 | **1.21** | **1.44** |

### 6.7 Opposing contributions between OSAMTL and StoA

In the subsection 6.6, we involuntarily regard various state-of-the-art approaches as the primary methods for learning from complex noisy labels, and investigated the contributions of OSAMTL to StoA by introducing OSAMTL to StoA for performance improvement of handling complex noisy labels. However, an interesting question is how about the contributions of StoA to OSAMTL, when we regard OSAMTL as the primary method for learning from complex noisy labels and introduce StoA to OSAMTL for performance improvement. We consider the opposing contributions between OSAMTL and StoA can probably show some interesting guiding clues for future works to handle complex noisy labels.

We regard metrics produced by OSAMTL_StoA(T1&T2) minus StoA as quantitative contributions of OSAMTL to StoA, and metrics produced by OSAMTL_StoA(T1&T2) minus OSAMTL(T1&T2) as quantitative contributions of StoA to OSAMTL. Base on the experiments conducted in subsections 6.4-6, the quantitative opposing contributions between StoA and OSAMTL are shown as Fig. 20-21 and the statistics corresponding to Fig. 20-21 are respectively shown in Table 20-21. Note, for computing the statistics corresponding to the Peer and NCE-SCE cases, we employed OSAMTL_StoA_46(T1&T2) solutions as OSAMTL_StoA(T1&T2) solutions.

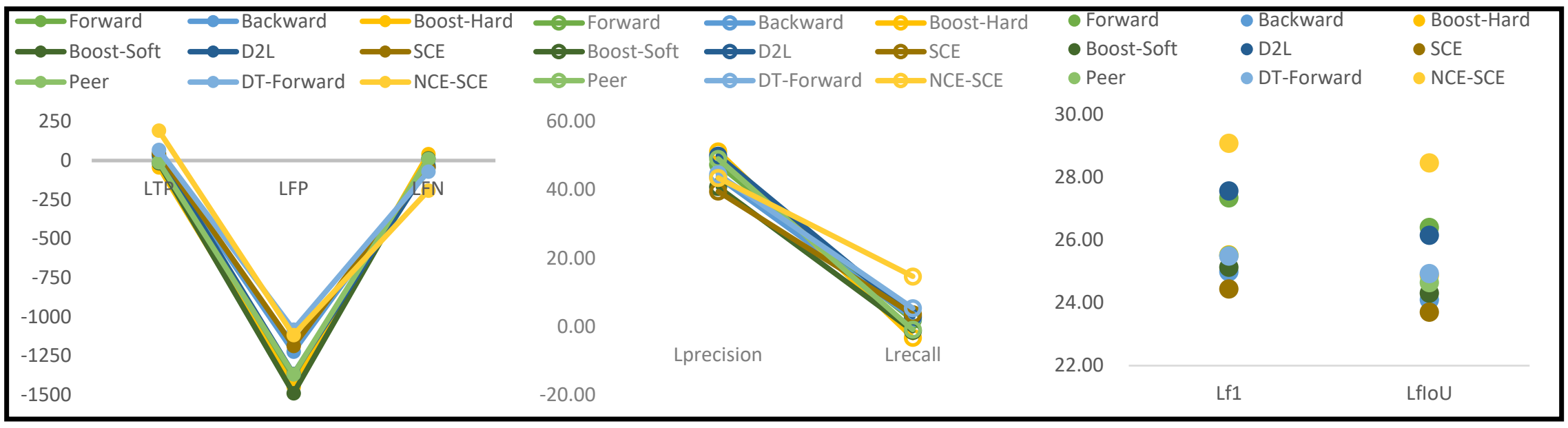

Fig. 20. Quantitative contributions of OSAMTL to StoA.

Table 20. Statistics corresponding to Fig. 20

| OSAMTL_StoA(T1&T2) minus StoA | LTP | LFP | LFN | Lprecision | Lrecall | Lf1 | LfIoU |
|---|---|---|---|---|---|---|---|
| Forward | 26 | -1414 | -26 | 47.40 | 1.96 | 27.34 | 26.40 |
| Backward | 34 | -1222 | -34 | 43.77 | 2.59 | 24.99 | 24.10 |
| Boost-Hard | -42 | -1449 | 42 | 51.25 | -3.28 | 25.53 | 24.87 |
| Boost-Soft | -16 | -1489 | 15 | 40.83 | -1.25 | 25.13 | 24.31 |
| D2L | 39 | -1366 | -39 | 50.08 | 3.00 | 27.56 | 26.15 |
| SCE | 47 | -1185 | -47 | 39.57 | 3.62 | 24.44 | 23.70 |
| Peer | -11 | -1368 | 11 | 49.11 | -0.86 | 25.51 | 24.64 |
| DT-Forward | 70 | -1081 | -70 | 44.56 | 5.40 | 25.49 | 24.93 |
| NCE-SCE | 192 | -1117 | -192 | 43.52 | 14.69 | **29.08** | **28.46** |

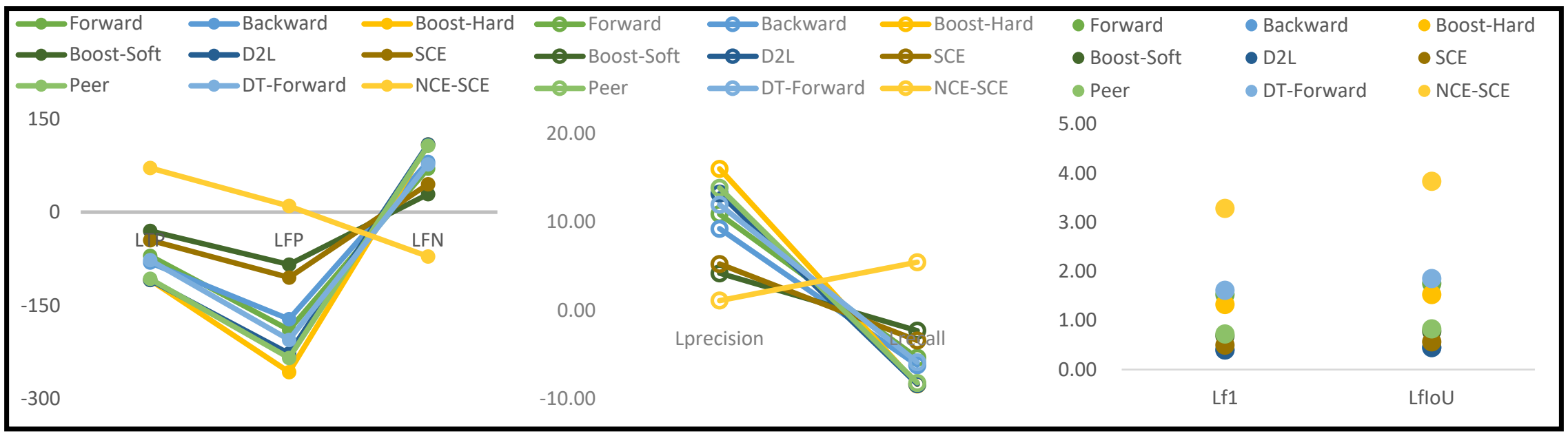

Fig. 21. Quantitative contributions of StoA to OSAMTL.

Table 21. Statistics corresponding to Fig. 21

| OSAMTL_StoA(T1&T2) minus OSAMTL_(T1&T2) | LTP | LFP | LFN | Lprecision | Lrecall | Lf1 | LfIoU |
|---|---|---|---|---|---|---|---|
| Forward | -70 | -189 | 70 | 10.88 | -5.35 | 1.53 | 1.76 |
| Backward | -81 | -172 | 81 | 9.23 | -6.18 | 0.40 | 0.45 |
| Boost-Hard | -109 | -257 | 109 | 15.99 | -8.36 | 1.33 | 1.53 |
| Boost-Soft | -30 | -84 | 29 | 4.20 | -2.29 | 0.69 | 0.78 |
| D2L | -109 | -227 | 109 | 13.20 | -8.35 | 0.40 | 0.45 |
| SCE | -45 | -105 | 45 | 5.24 | -3.40 | 0.50 | 0.57 |
| Peer | -107 | -234 | 107 | 13.85 | -8.19 | 0.73 | 0.83 |
| DT-Forward | -77 | -205 | 77 | 11.94 | -5.86 | 1.62 | 1.86 |
| NCE-SCE | 71 | 10 | -71 | 1.11 | 5.44 | **3.29** | **3.84** |

From Fig. 20 and Table 20, we can observe that OSAMTL can make contributions to help StoA increase LTP and at the same time reduce LFP and LFN, leading to remarkable improvement in Lprecision and significant increment in Lrecall which remarkably improve Lf1 and LfIoU. From Fig. 21 and Table 21, we can observe that StoA can make contributions to help OSAMTL decrease more LFP than LTP and at the same time increase LFN, leading to improvement in Lprecision and decrement in Lrecall which modestly improve Lf1 and LfIoU. These results of opposing contributions between OSAMTL and StoA lead us to conclude that, for the performance improvement of handling complex noisy labels, the contributions of OSAMTL to StoA are much more than the contributions of StoA to OSAMTL.

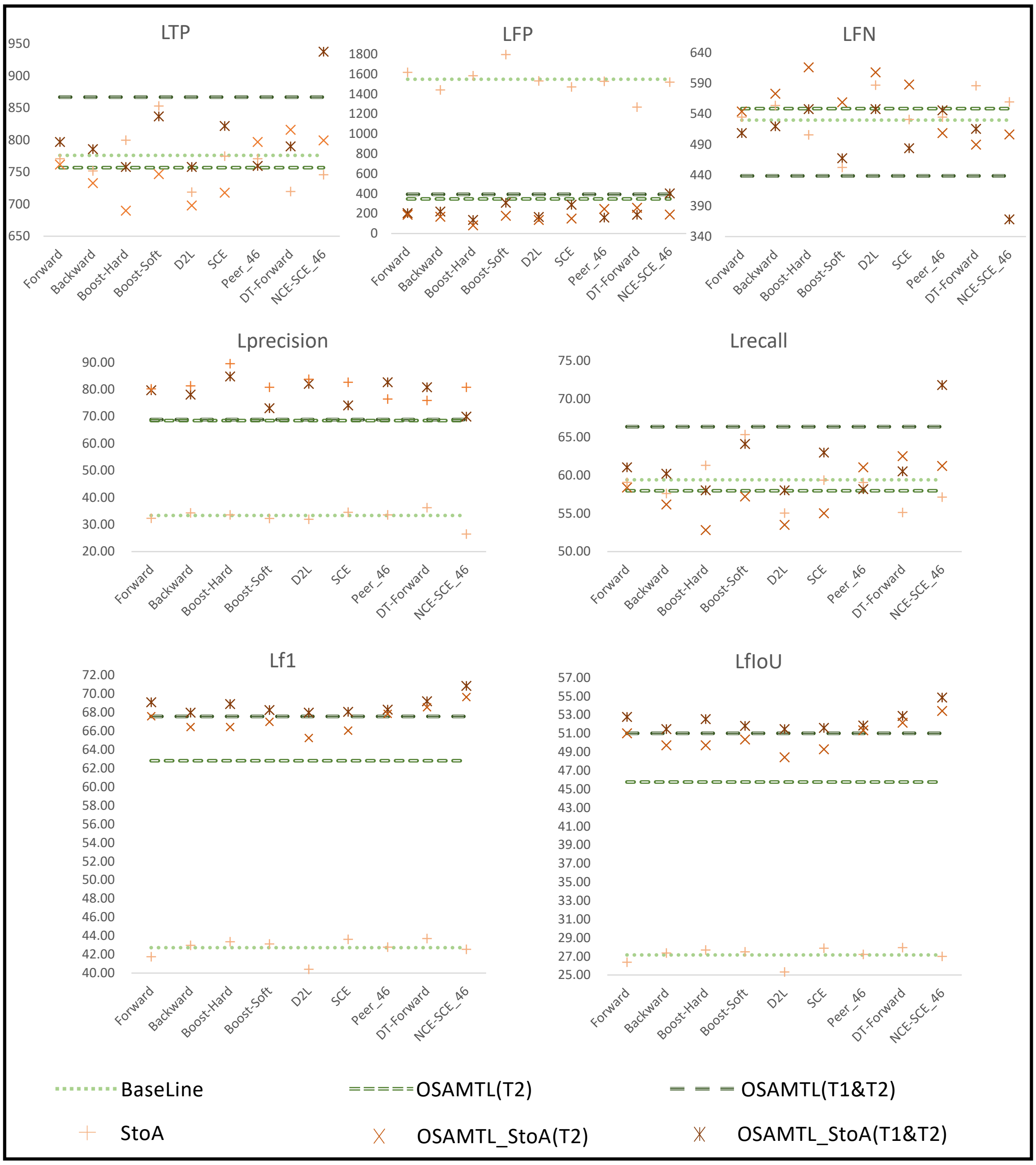

Fig. 22. Summarized comparisons between various solutions. BaseLine: The solution that naively learns from the given complex noisy labels ($Target_1$). OSAMTL(T2): The solution that learns from the additional targets ($Target_2$) abduced by the one-step logical reasoning of OSAMTL. OSAMTL(T1&T2): The solution that imposes the two types of targets ($Target_1$ and $Target_2$) abduced by the one-step logical reasoning of OSAMTL upon machine learning. StoA: The solution that directly employs a state-of-the-art approach to handle the given complex noisy labels ($Target_1$). OSAMTL_StoA(T2): The solution that introduces the one-step logical reasoning of OSAMTL to StoA, employing the StoA approach to handle the noisy targets $Target_2$ abduced by the one-step logical reasoning of OSAMTL. OSAMTL_StoA(T1&T2): The solution that introduces the multi-target learning procedure of OSAMTL to StoA, combining StoA with the multi-target learning procedure of OSAMTL to learn from both noisy targets $Target_1$ and $Target_2$.

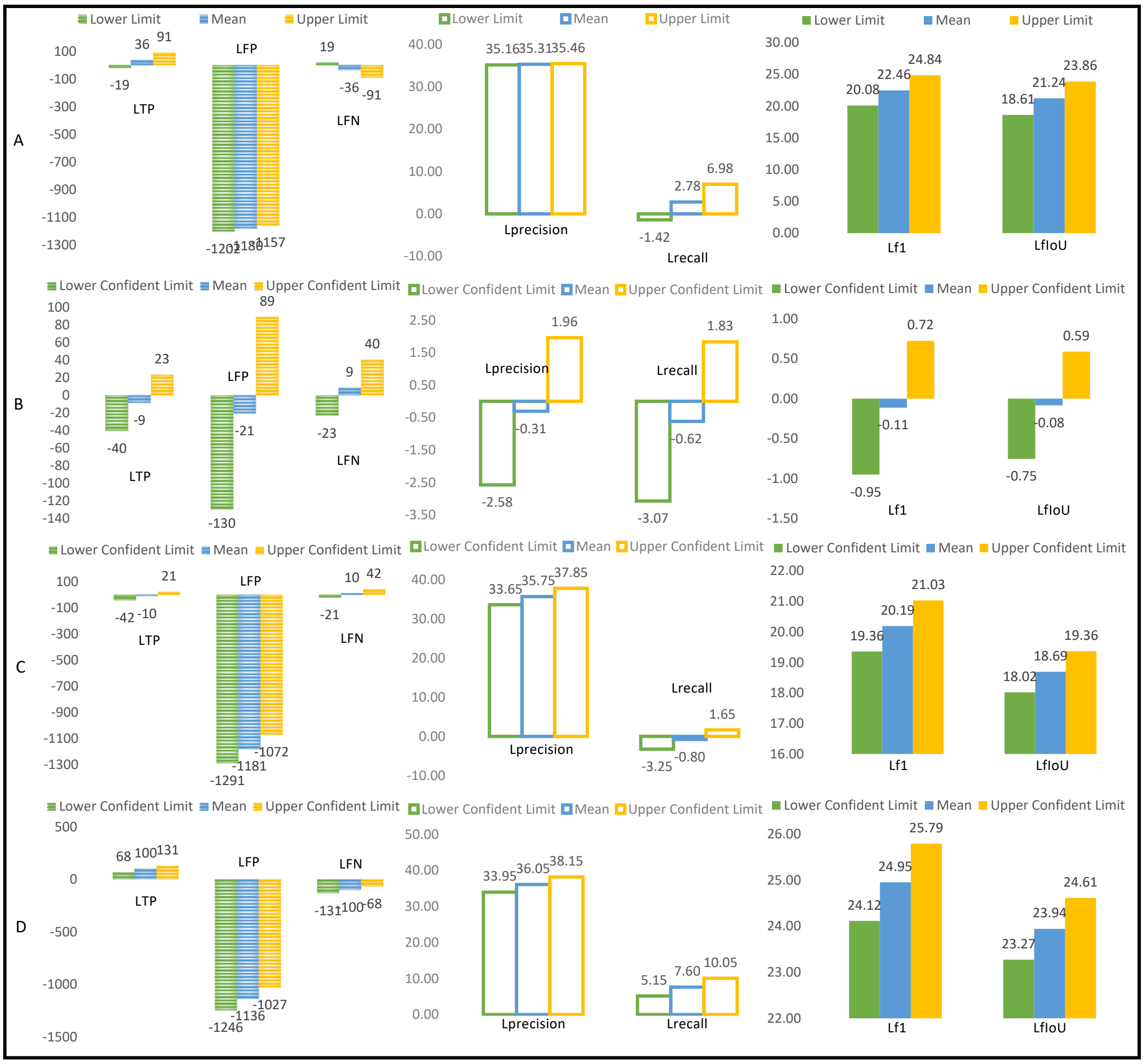

Fig. 23. Intervals for performance improvements of one solution compared with another. A: Intervals for performance improvements of the two basic solutions of OSAMTL (OSAMTL(T1) and OSAMTL(T1&T2)) compared with BaseLine; Thereinto, OSAMTL(T1) minus BaseLine is the lower limit, OSAMTL(T1&T2) minus BaseLine is the upper limit, and the average of the lower and upper limits is the mean. B: 95% confident intervals for performance improvements of OSAMTL_StoA(T2) compared with StoA. C: 95% confident intervals for performance improvements of SOAMTL(T2) compared with StoA. D: 95% confident intervals for performance improvements of SOAMTL(T1&T2) compared with StoA.

## 6.8 Summarization

In this subsection, we give summarizations about the experiments conducted in Section 6.4 to Section 6.7, including experimental summarization and solution complexity.

### 6.8.1 Experimental summarization

Fig. 22 shows the summarized comparisons between various solutions constructed in the conducted experiments, including BaseLine, OSAMTL(T2), OSAMTL(T1&T2), StoA, OSAMTL_StoA(T2) and OSAMTL_StoA(T1&T2). Specifically, the intervals for performance improvements of one solution compared with another are shown as Fig.

23-24. Fig. 23 includes the intervals for performance improvements respectively of the two basic solutions of OSAMTL (OSAMTL(T1) and OSAMTL(T1&T2)) compared with BaseLine, StoA compared with BaseLine, OSAMTL(T2) compared with StoA, and OSAMTL(T1&T2) compared with StoA. Fig. 24 includes the intervals for performance improvements respectively of OSAMTL_StoA(T2) compared with StoA, SOAMTL_StoA(T2) compared with OSAMTL(T2), SOAMTL_StoA(T1&T2) compared with OSAMTL_StoA(T2). Additionally, the intervals for quantitative opposing contributions between OSAMTL and StoA for performance improvement of handling complex noisy labels are shown as Fig. 25. Based on the information presented in Fig. 22-25, we have following summarizations:

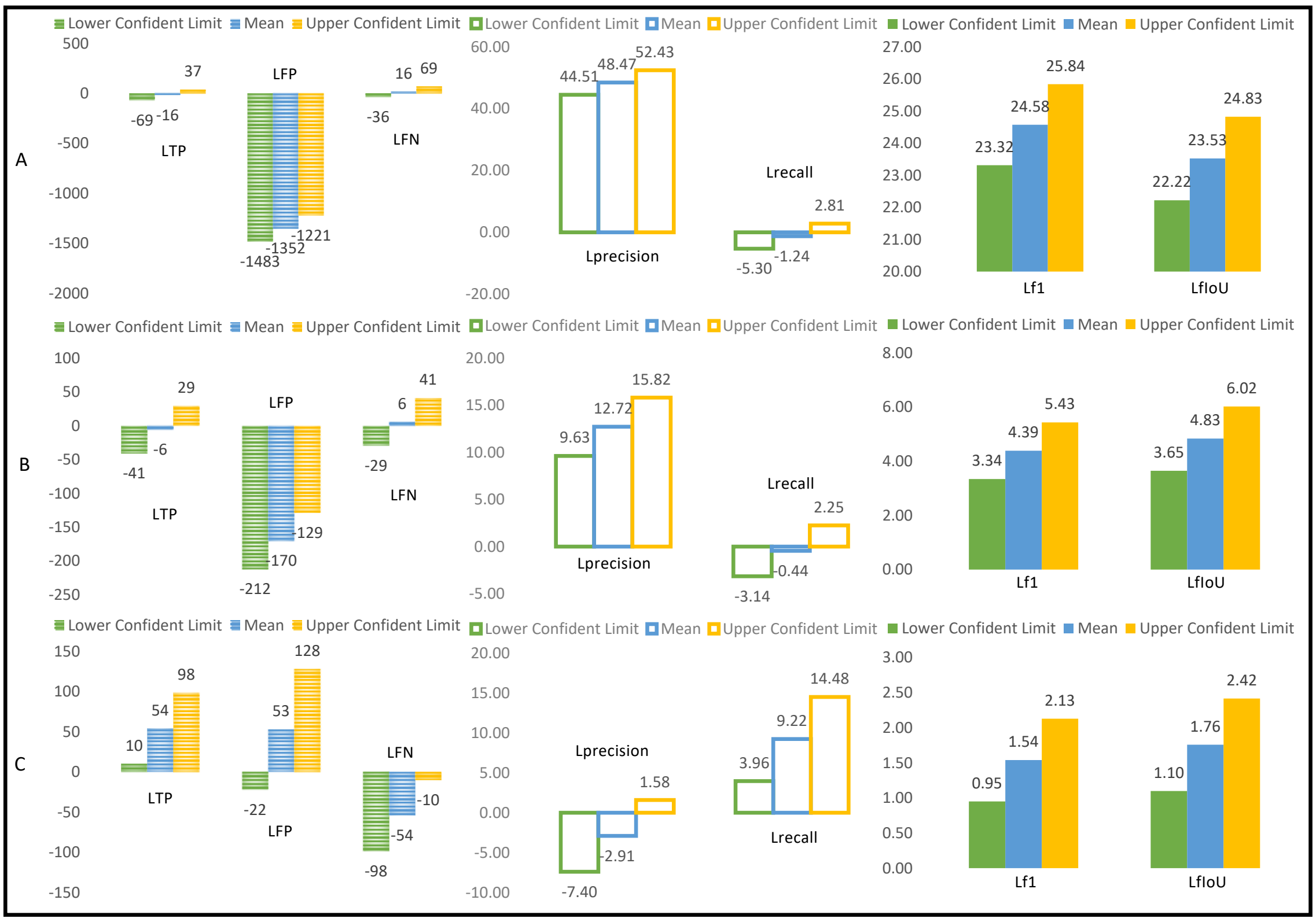

Fig. 24. Intervals for performance improvements of one solution compared with another. A: 95% confident intervals for performance improvements of OSAMTL_StoA(T2) compared with StoA. B: 95% confident intervals for performance improvements of SOAMTL_StoA(T2) compared with OSAMTL(T2). C: 95% confident intervals for performance improvements of SOAMTL_StoA(T1&T2) compared with OSAMTL_StoA(T2).

1) From the comparisons between the two basic solutions of OSAMTL (OSAMTL(T1) and OSAMTL(T1&T2)) and BaseLine (see corresponding solutions in Fig. 22 and intervals for performance improvements in Fig. 23.A), we can summarize that the one-step logical reasoning of OSAMTL can provide more effective information ($Target_2$) to enable logically more rationale solutions than naively learning from the given complex noisy labels ($Target_1$), and the multi-target learning procedure of OSAMTL can further make up for respective shortcoming of BaseLine and OSAMTL(T2) to achieve results even better than the best solution of BaseLine and OSAMTL(T2) to reach a new high.

2) From the comparison between StoA and BaseLine (see corresponding solutions in Fig. 22 and intervals for performance improvements Fig. 23.B), we can summarize that the referred various state-of-the-art approaches indeed have limitations in handling complex noisy labels ($Target_1$). From the comparisons between the two basic solutions of OSAMTL (OSAMTL(T1) and OSAMTL(T1&T2)) and StoA (see corresponding solutions in Fig. 22 and Fig. 23.C-D), we can summarize that OSAMTL possesses unique capability to achieve logically more rational predictions, which is beyond the referred various state-of-the-art approaches in handling complex noisy labels ($Target_1$).

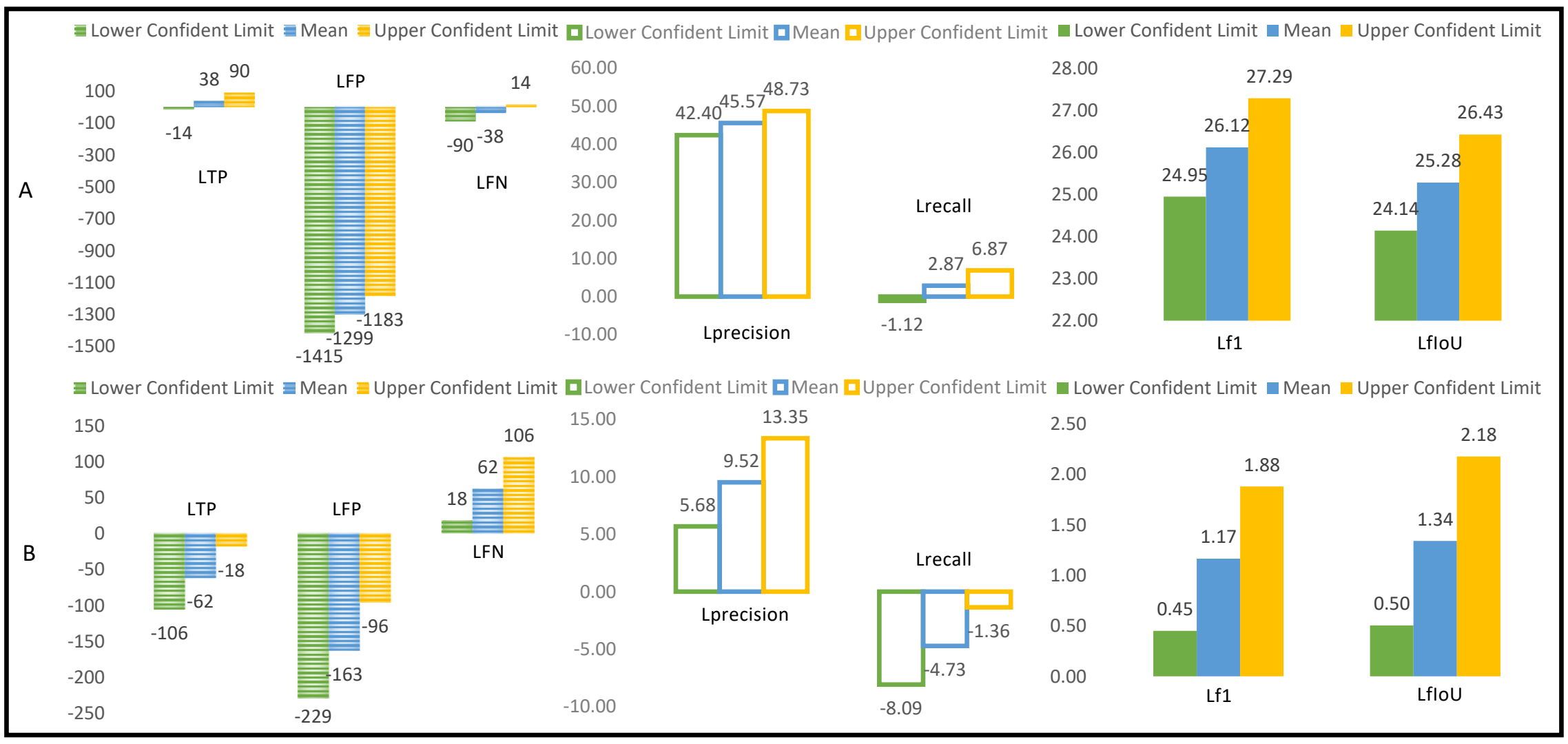

Fig. 25. Intervals for quantitative opposing contributions between OSAMTL and StoA. A: 95% confident intervals for performance improvements of OSAMTL_StoA(T1&T2) compared with StoA. B: 95% confident intervals for performance improvements of OSAMTL_StoA(T1&T2) compared with OSAMTL(T1&T2).

3) From the comparisons between OSAMTL_StoA(T2) and StoA (see corresponding solutions in Fig. 22 and intervals for performance improvements Fig. 24.A), we can summarize that introducing the one-step logical reasoning of OSAMTL to StoA can provide noisy targets ($Target_2$) better than the given complex noisy labels ($Target_1$) for the referred various state-of-the-art approaches to achieve logically more rationale predictions. From the comparison between OSAMTL_StoA(T2) and OSAMTL(T2) (see corresponding solutions in Fig. 22 and Fig. 24.B), we can summarize that introducing the one-step logical reasoning of OSAMTL to StoA can release more potentials of the referred various state-of-the-art approaches by providing them noisy targets ($Target_2$) better than the given complex noisy labels ($Target_1$), for the referred various state-of-the-art approaches show better capabilities at handling $Target_2$ than handling $Target_1$, see the difference between Fig. 24.B and Fig. 23.B. From the comparison between OSAMTL_StoA(T1&T2) and OSAMTL_StoA(T2) (see corresponding solutions in Fig. 22 and intervals for performance improvements Fig. 24.C), we can summarize that introducing the multi-target learning procedure of OSAMTL to StoA can as well further enable the referred state-of-the-art approaches to achieve new high performances even better than simply introducing the one-step logical reasoning of OSAMTL.

4) From the comparison between the performance improvement of OSAMTL_StoA(T1&T2) compared with StoA and the performance improvement of OSAMTL_StoA(T1&T2) compared with OSAMTL(T1&T2) (see corresponding solutions in Fig. 22 and Fig. 25. A-B), we can summarize that, for the performance improvement of handling complex noisy labels, OSAMTL can contribute much more to StoA than what StoA can contribute to OSAMTL.

**6.8.2 Solution Complexity**

The complexities for the testing procedures of all the solutions constructed for experiments are the same, as all the solutions share a same inference pipeline which is from the input of the Deep CNN-based image semantic segmentation model (Fig. 6) to its output. But the complexities for the training procedures of the solutions constructed for experiments are different. We use three factors to reflect the complexity for the training procedure of a solution, including Abduction that a solution needs a procedure to reduce the inconsistencies between the given noisy labels and our prior knowledge of the given knowledge base, Pre-training that a solution needs a pre-training procedure, and Multi-target that a solution needs multiple targets to compute the loss for machine learning. Based on these three factors, the complexities for the training procedures of various solutions constructed in this paper are shown as Table 22.

Table 22. Complexities for the training procedures of various solutions.

| Solution | | Abduction | Pre-training | Multi-target |
|---|---|---|---|---|
| BaseLine | | No | No | No |
| OSAMTL(T1) | | Yes | No | No |
| OSAMTL(T1&T2) | | Yes | No | Yes |
| | Forward | No | Yes | No |
| | Backward | No | Yes | No |
| | Boost-Hard | No | No | No |
| | Boost-Soft | No | No | No |
| StoA | D2L | No | No | No |
| | SCE | No | No | No |
| | Peer | No | No | No |
| | DT-Forward | No | Yes | No |
| | NCE-SCE | No | No | No |
| OSAMTL_StoA(T1) | | Yes | Refer to StoA | No |
| OSAMTL_StoA(T1&T2) | | Yes | Refer to StoA | Yes |

### *6.9 Qualitative results*

Some qualitative results are provided in the Supplementary 3 as visualized proofs to show the basic effectiveness of OSAMTL and its effectiveness of improving various state-of-the-art approaches in handling complex noisy labels.

## 7. Conclusion

Various approaches have been proposed to handle noisy labels and have achieved remarkable success in a variety of applications. However, many of current state-of-the-art approaches for learning from noisy labels have limitations in specific areas, such as medical histopathology whole slide image analysis (MHWSIA), since in this area accurate/noisy-free ground-truth labels are difficult or even impossible to collect which leads to complex noise existing in labels. Moreover, due to the great difficulty in collecting noisy-free ground-truth labels in such areas, the appropriate evaluation strategy for validation/testing is unclear. To alleviate these two problems, we present one-step abductive multi-target learning (OSAMTL) that imposes a one-step logical reasoning upon machine learning via a multi-target learning procedure to constrain the predictions of the learning model to be subject to our prior knowledge, and propose a logical assessment formula (LAF) that evaluates the logical rationality of the outputs of an approach by estimating the consistencies between the predictions of the learning model and the logical facts narrated from the results of the one-step logical reasoning of OSAMTL.

OSAMTL only imposes a one-step logical reasoning upon machine learning via a multi-target learning procedure, which is different from the original abductive learning [11] that imposes logical reasoning upon machine learning iteratively via a single-target learning procedure. Enabling the predictions of the learning model to be logically rationale via a multi-target learning procedure, OSAMTL is also different from various state-of-the art approaches for learning from noisy labels, such as backward and forward correction [5], bootstrapping [6], D2L [8], SCE [9], Peer [10] and etc. The methodology of these state-of-the-art approaches identifies potentially noisy-labeled instances based on various pre-assumptions about noisy-labeled instances and makes some corresponding corrections to approach the true target serially. However, the methodology of OSAMTL abduces multiple targets that contain information logically consistent with our prior knowledge about the true target to approximate the true target parallelly. OSAMTL provides new thoughts for more effectively exploiting noisy labels to achieve logically more rational predictions. In addition, LAF is defined on noisy labels to enable us to evaluate the logical rationality of an approach without requiring noise-free ground-truth labels, which is different from usual evaluation metrics defined on noisy-free ground-truth labels. LAF provides a new addition to usual evaluation strategies for evaluating approaches for learning from noisy labels.

Referring to OSAMTL we implemented an OSAMTL-based image semantic segmentation solution for the Helicobacter pylori (H. pylori) segmentation task [12] in MHWSIA, and referring to LAF we implemented a LAF-based evaluation strategy that is appropriate in the context of H. pylori segmentation. Based on these two implementations, we conducted extensive experiments to show that OSAMTL is fairly effective to handle complex noisy labels under circumstances that we have some prior knowledge about the true target of a specific task. First, experiments for basic effectiveness of OSAMTL (Section 6.4) shows that the one-step logical reasoning of OSAMTL can provide more effective information to enable logically more rationale solution than the solution that naively learns from the given complex noisy labels; and the multi-target learning procedure of OSAMTL can achieve even better results, further leveraging the more effective information provided by the one-step logical reasoning of OSAMTL to reach a new high. Second, experiments for the two basic solutions of OSAMTL compared with various state-of-the-art approaches (Section 6.5) show that various state-of-the-art approaches for learning with noisy labels indeed have limitations in handling complex noisy labels, while OSAMTL possesses unique capability to achieve logically more rational predictions, which is beyond various state-of-the-art approaches in handling complex noisy labels. Third, experiments for effectiveness of OSAMTL in improving state-of-the-art (Section 6.6) show that OSAMTL have significant potentials in improving various state-of-the-art approaches to achieve logically more rational results. Fourth, experiments for opposing contributions between OSAMTL and various state-of-the-art approaches show that, for performance improvement in handling complex noisy labels, OSAMTL can contribute much more to various state-of-the-art approaches than what various state-of-the-art approaches can contribute to OSAMTL. Finally, various qualitive results provide visual proofs to show the capability of OSAMTL to achieve logically more rationale results.

The micro significance of this presentation is that it reports on the first positive case that theoretically and practically supports that OSAMTL can be effective to handle complex noisy labels in MHWSIA. Moreover, since the opposing contributions between OSAMTL and various state-of-the-art approaches show that the contributions of OSAMTL to various state-of-the-art approaches are much more than the contributions of various state-of-the-art approaches to OSAMTL, an interesting question that can be raised is which methodology has more potentials to be taken as the primary methodology for handling complex noisy labels, the methodology of OSAMTL, the methodology of various state-of-the-art approaches, or their hybrid? Although promising experiments have shown that more potentials of OSAMTL can be released by changing the weights for the multi-target learning procedure, one problem remains unsolved is how to design an appropriate strategy that can automatically release more potentials of OSAMTL via one-time training. Besides, more investigations need to be conducted to prove the generalization of OSAMTL, i.e., OSAMTL can be applied to other applications in the field of MHWSIA or beyond.

The macro significance of this presentation is that it provides an example paradigm (the Y-shape shown as Fig. 26) of combining perception/machine learning (data-driven) and cognition/logical reasoning (knowledge driven) together via OSAMTL to build better AI system. More specifically, in the case of OSAMTL applied to H. pylori segmentation presented in this paper, we implemented the one-step logical reasoning of OSAMTL by carrying out many logical reasoning activities (results of logical reasoning and their proofs provided in Supplementary 2) based on the given noisy data and knowledge base. In fact, results of logical reasoning and their proofs can provide discrete representations of human cognition process, which are the fundamental foundation of computational cognition [38]. Regarding to the case of OSAMTL applied to H. pylori segmentation presented in this paper, the logical reasoning activities we carried out in fact transformed our cognition about the H. pylori segmentation task into multiple learnable targets that can be easily employed by perception/machine learning; thus, perception and cognition are combined via OSAMTL to build better AI system.

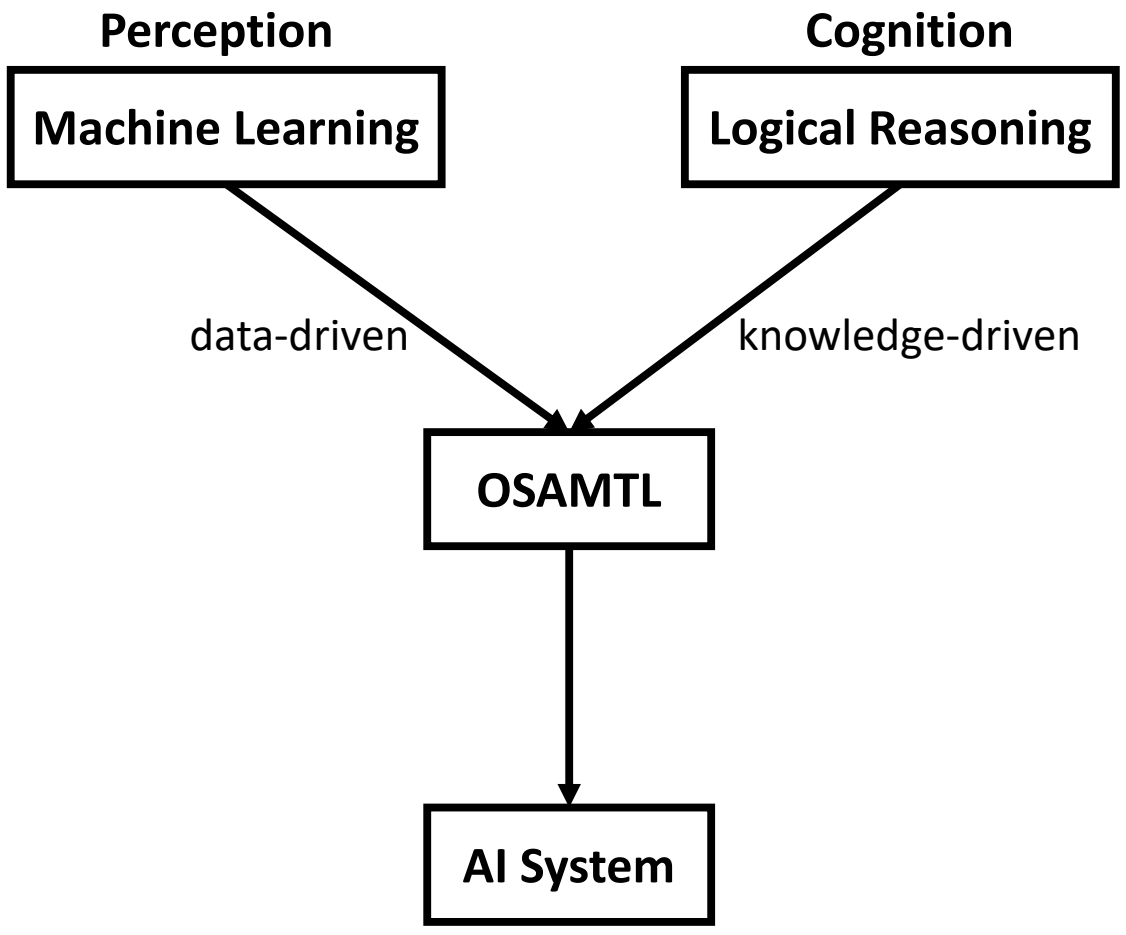

Fig. 26. The Y-shape of combining perception/machine learning and cognition/logical reasoning via OSAMTL to build better AI system.

## Acknowledgements

This work was supported by the Sichuan Science and Technology Program (2020YFS0088); the 1 • 3 • 5 project for disciplines of excellence Clinical Research Incubation Project, West China Hospital, Sichuan University (2019HXFH036), China; the 1.3.5 project for disciplines of excellence, West China Hospital, Sichuan University (ZYGD18012), China; the Technological Innovation Project of Chengdu New Industrial Technology Research Institute (2017-CY02-00026-GX).

## Data Availability

The datasets generated during and/or analysed during the current study are not publicly available due to cooperation contract but are available from the corresponding author on reasonable request.

## Conflict of interest

The authors declare that they have no conflict of interest.

# Supplementary 1: Visualized Differences of OSAMTL from Other Approaches

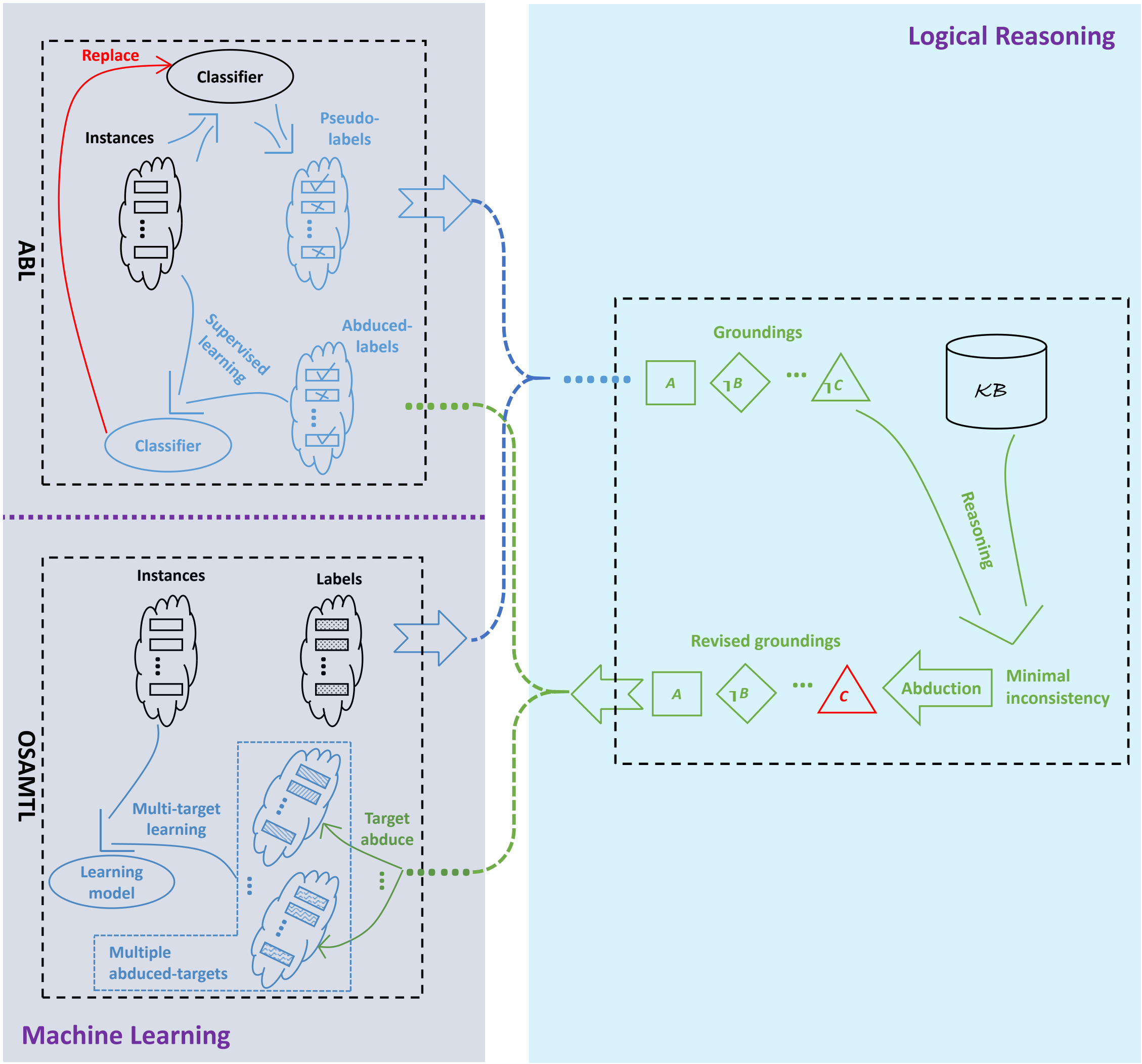

Fig. 1. Difference between OSAMTL and ABL. Left: respective machine learning parts of OSAMTL and ABL. Right: common logical reasoning prat of OSAMTL and ABL.

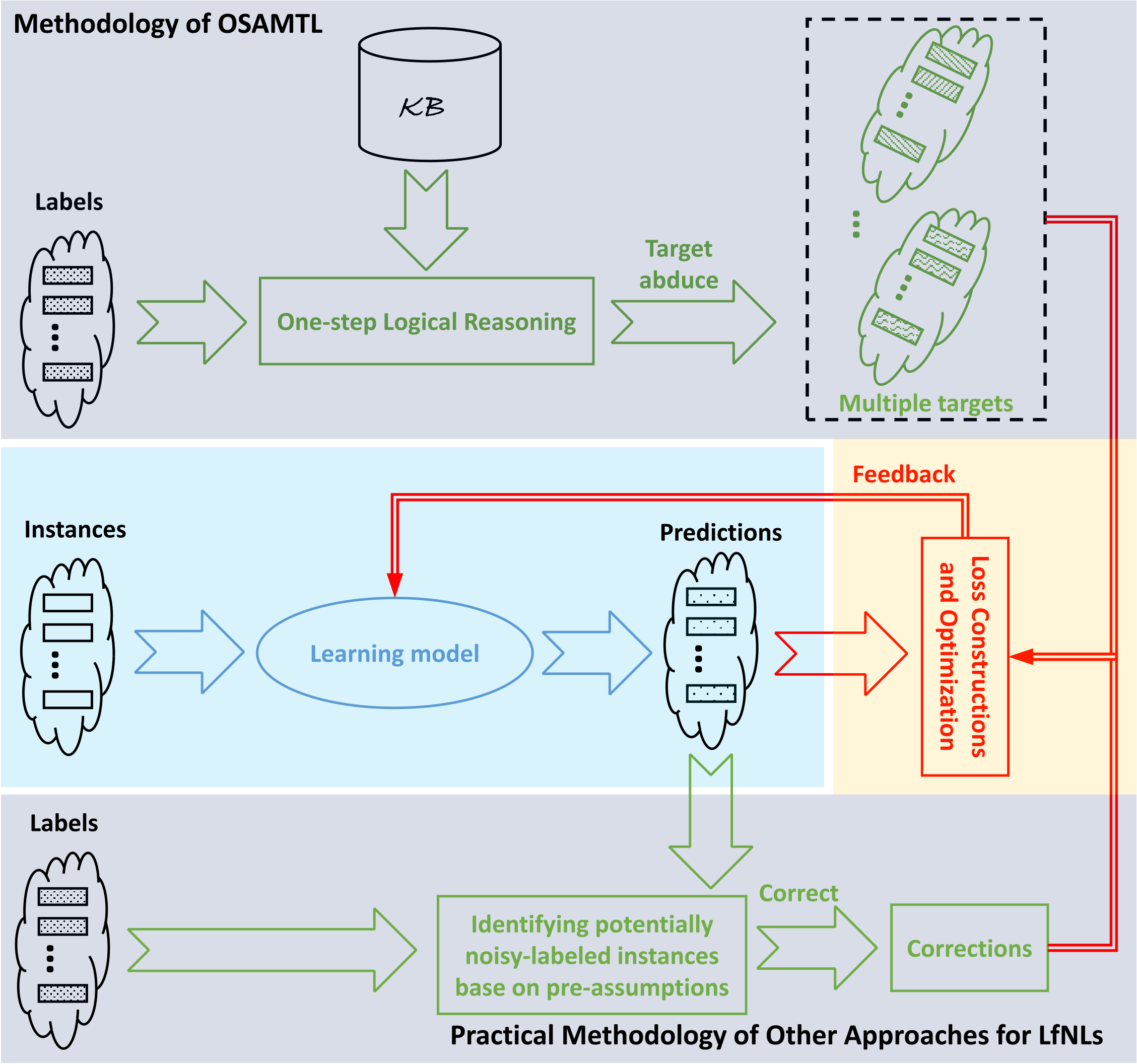

Fig. 2. Difference between the methodology of OSAMTL and the practical methodology of various other approaches for learning from noisy labels (LfNLs).

# Supplementary 2: Proofs for the Validation of Reasonings for the Instance Implementations of OSAMTL and LAF for H. Pylori Segmentation

## Preliminary of Logical Reasoning

We introduce some propositional connectives and rules for proof of propositional logical reasoning, which are respectively shown as Table 1 and Table 2, for the logical reasonings conducted in this paper.

Table 1. Propositional connectives

| Connective | Meaning |
|---|---|
| $\wedge$ | conjunction |
| $\rightarrow$ | implication |

Table 2. Rules for proof of propositional logical reasoning, $\vdash$ denotes 'bring out'

| Rule | Meaning |
|---|---|
| $\wedge -$ | reductive law of conjunction: $A \wedge B, \vdash A$ or $B$. |
| $\wedge +$ | additional law of conjunction: $A, B, \vdash A \wedge B$. |
| MP | modus ponens: $A \rightarrow B, A, \vdash B$. |
| HS | hypothetical syllogism: $A \rightarrow B, B \rightarrow C, \vdash A \rightarrow C$. |

## Proof of Reasoning 1

**Reasoning 1**. *Given* $G$ *and* $KB$, *then true H. pylori negatives represented by* $G_1$ *are almost all correct, the recall of* $G_2$ *to represent true H. pylori positives is very high, and the precision of* $G_2$ *to represent true H. pylori positives is very low*.

**Proof-R1**. Firstly, with $G$ and $KB$, we have following derived preconditions for Reasoning 1.

1. If $G$ is given, then $G_1$ exists and $G_2$ exists.
2. If $G_1$ exists and $G_2$ exists, then the union of $G_1$ and $G_2$ can transform into $L$.
3. If $G_1$ exists and $KB$ is given, then true H. pylori negatives represented by $G_1$ are almost all correct.
4. If true H. pylori negatives represented by $G_1$ are almost all correct, $G_2$ exists, and the union of $G_1$ and $G_2$ can transform into $L$, then true H. pylori positives are almost all included by $G_2$.
5. If $G_2$ exists and $KB$ is given, then true H. pylori positives only occupy a very small fraction of the H. pylori positives represented by $G_2$.
6. If true H. pylori positives are almost all included in $G_2$, and true H. pylori positives only occupy a very small fraction of the H. pylori positives represented by $G_2$, then the true H. pylori positives (TP) are almost all constant in $G_2$, the false H. pylori negatives (FN) are almost null in $G_2$, and the false H. pylori positives (FP) are very many in $G_2$.

7. If the true H. pylori positives (TP) are almost all constant in $G_2$, and the false H. pylori negatives (FN) are almost null in $G_2$, then the recall (TP/(TP+FN)) of $G_2$ to represent true H. pylori positives is very high.
8. If the true H. pylori positives (TP) are almost all constant in $G_2$, and the false H. pylori positives (FP) are very many, then the precision (TP/(TP+FP)) of $G_2$ to represent true H. pylori positive areas is very low.

Secondly, we give the propositional symbols for the above preconditions 1-15 for Reasoning 1, which are shown in Table 3.

Table 3. Propositional symbols of preconditions for Reasoning 1

| Symbol | Meaning |
|---|---|
| $l$ | $G$ is given |
| $m$ | $KB$ is given |
| $n$ | $G_1$ exists |
| $o$ | $G_2$ exists |
| $p$ | the union of $G_1$ and $G_2$ can transform into $L$ |
| $q$ | true H. pylori negatives represented by $G_1$ are almost all correct |
| $r$ | true H. pylori positives are almost all included by $G_2$ |
| $s$ | true H. pylori positives only occupy a very small fraction of the H. pylori positives represented by $G_2$ |
| $t$ | the true H. pylori positives (TP) are almost all constant in $G_2$ |
| $u$ | the false H. pylori negatives (FN) are almost null in $G_2$ |
| $v$ | the false H. pylori positives (FP) are very many |
| $w$ | the recall (TP/(TP+FN)) of $G_2$ to represent true H. pylori positives is very high |
| $x$ | the precision (TP/(TP+FP)) of $G_2$ to represent true H. pylori positives is very low |

Thirdly, referring to Table 3, we signify the propositional formalizations of the preconditions 1-9 for Reasoning 1 and Reasoning 1 via the propositional connectives listed in Table 1 as follows.

1) $l \rightarrow n \wedge o$ Precondition
2) $n \wedge o \rightarrow p$ Precondition
3) $n \wedge m \rightarrow q$ Precondition
4) $q \wedge o \wedge p \rightarrow r$ Precondition
5) $o \wedge m \rightarrow s$ Precondition
6) $r \wedge s \rightarrow t \wedge u \wedge v$ Precondition
7) $t \wedge u \rightarrow w$ Precondition
8) $t \wedge v \rightarrow x$ Precondition.

$l \wedge m \rightarrow q \wedge w \wedge x$ Reasoning 1.

Fourthly, we show the validity of Reasoning 1 via the rules for proof of propositional logical reasoning listed in Table 2.

∴ $\boldsymbol{l \wedge m \rightarrow q \wedge w \wedge x}$

9) $l \wedge m$ Hypothesis

| | |
|---|---|
| 10) $l$ | 9); $\wedge -$ |
| 11) $m$ | 9); $\wedge -$ |
| 12) $n \wedge o$ | 1), 10); MP |
| 13) $n$ | 12); $\wedge -$ |
| 14) $n \wedge m$ | 12), 11); $\wedge +$ |
| 15) $q$ | 3), 14); MP |
| 16) $o$ | 12); $\wedge -$ |
| 17) $p$ | 2), 12); MP |
| 18) $q \wedge o \wedge p$ | 15), 16), 17); $\wedge +$ |
| 19) $r$ | 4), 18); MP |
| 20) $o \wedge m$ | 16), 11); $\wedge +$ |
| 21) $s$ | 5), 20); MP |
| 22) $r \wedge s$ | 19), 21); $\wedge +$ |
| 23) $t \wedge u \wedge v$ | 6), 22); MP |
| 24) $t \wedge u$ | 23); $\wedge -$ |
| 25) $w$ | 7), 24); MP |
| 26) $t \wedge v$ | 23); $\wedge -$ |
| 27) $x$ | 8), 26); MP |
| 28) $q \wedge w \wedge x$ | 15), 25), 27); $\wedge +$ |
| 29) $l \wedge m \rightarrow q \wedge w \wedge x$ | 9)-28); Conditional Proof |

Reasoning 1 is valid, since the hypothesis $l \wedge m$ of the 9) step in above proof can be fullfilled with the conjunction of the input knowledge base $KB$ and the groundings $G$ extracted from the input labels $L$. Thus, we use the conclusions of Reasoning 1 to estimate the inconsistencies between $G$ and $KB$. From $q$, we can estimate that the inconsistency between the false H. pylori negatives $of\ G_1$ and the false H. pylori negatives of $KB$ is limitted. And, from $w$, we can estimate that the inconsistency between the recall of $G_2$ and the recall of $KB$ to represent true H. Pylori positives is limited. Finally, from $x$, we can estimate that the inconsistency between the precision of $G_2$ and the precision of $KB$ to represent true H. Pylori positives is high.

## Proof of Reasoning 2

**Reasoning 2**. *Given $G_2$ and $KB$, then black dot-like pixels inside $G_2$ which distribute in luminal areas and have obvious gradient with its neighbourhood are true H. pylori positives with high probability.*

**Proof-R2**. We show that the result of Reasoning 2 can, under some additional hypotheses, increase the precision of representing true H. Pylori positives, thus to reduce $Incon_3$. Firstly, with the conjunction of $G_2$ and $KB$, we have following derived preconditions.

1. If $G_2$ is given, the precision of $G_2$ to represent true H. Pylori positives is TP/(TP+FP).
2. If $KB$ is given, then $K_1$ exists, $K_2$ exists and $K_3$ exists.

3. If $G_2$ is given and $K_1$ exists, then inside $G_2$ pixels not in luminal areas are excluded.
4. If $G_2$ is given and $K_2$ exists, then inside $G_2$ pixels not black dot-like regions are excluded.
5. If $G_2$ is given and $K_3$ exists, then inside $G_2$ pixels not having obvious gradient with its neighbourhood are excluded.
6. If inside $G_2$ pixels not in luminal areas are excluded, inside $G_2$ pixels not black dot-like regions are excluded, and inside $G_2$ pixels not having obvious gradient with its neighbourhood are excluded, then black dot-like pixels inside $G_2$ which distribute in luminal areas and have obvious gradient with its neighborhood are true H. pylori positives with high probability ($\neg G_2$).

Secondly, we give propositional symbols for the above preconditions 1-6 for Reasoning 2 which are listed in Table 4.

Table 4. Propositional symbols of the preconditions for Reasoning 2

| Symbol | Meaning |
|---|---|
| $f$ | $G_2$ is given |
| $g$ | $KB$ is given |
| $h$ | $K_1$ exists |
| $i$ | $K_2$ exists |
| $j$ | $K_3$ exists |
| $k$ | the precision of $G_2$ to represent true H. Pylori positive areas is TP/(TP+FP) |
| $l$ | inside $G_2$ pixels not in luminal areas are excluded |
| $m$ | inside $G_2$ pixels not black dot-like regions are excluded |
| $n$ | inside $G_2$ pixels not having obvious gradient with its neighbourhood are excluded |
| $\neg G_2$ | black dot-like pixels inside $G_2$ which distribute in luminal areas and have obvious gradient with its neighbourhood are true H. pylori positives with high probability |

Thirdly, referring to Table 4, we signify the propositional formalizations of the preconditions 1-6 for Reasoning 2 and Reasoning 2 via the propositional connectives listed in Table 1 as follows.

1) $f \rightarrow k$ — Precondition
2) $g \rightarrow h \wedge i \wedge j$ — Precondition
3) $f \wedge h \rightarrow l$ — Precondition
4) $f \wedge i \rightarrow m$ — Precondition
5) $f \wedge j \rightarrow n$ — Precondition
6) $l \wedge m \wedge n \rightarrow \neg G_2$ — Precondition

$f \wedge g \rightarrow \neg G_2$ — Reasoning 2

Fourthly, we show the validity of Reasoning 2 via the rules of proofs propositional logical reasoning listed in Table 2 as follow.

$\therefore\ \boldsymbol{f \wedge g \rightarrow \neg G_2}$

7) $f \wedge g$ — Hypothesis
8) $f$ — 7); $\wedge -$

| | |
|---|---|
| 9) $g$ | 7); $\wedge -$ |
| 10) $h \wedge i \wedge j$ | 2), 9); MP |
| 11) $h$ | 10); $\wedge -$ |
| 12) $i$ | 10); $\wedge -$ |
| 13) $j$ | 10); $\wedge -$ |
| 14) $f \wedge h$ | 8), 11); $\wedge +$ |
| 15) $l$ | 3), 14); MP |
| 16) $f \wedge i$ | 8), 12); $\wedge +$ |
| 17) $m$ | 4), 16); MP |
| 18) $f \wedge j$ | 8), 13); $\wedge +$ |
| 19) $n$ | 5), 18); MP |
| 20) $l \wedge m \wedge n$ | 15), 17), 19); $\wedge +$ |
| 21) $\neg G_2$ | 6), 20); MP |
| 22) $f \wedge g \rightarrow \neg G_2$ | 7)-21); Conditional Proof |

As the hypothesis $f \wedge g$ of the 7) step in above proof can be fulfilled with the conjunction of the input knowledge base $KB$ and the grounding $G_2$ extracted from the input labels $L$, Reasoning 2 is valid.

## Proof of Reasoning 3

**Reasoning 3**. Given $\neg G_2$, *the precision of* $G_2$ *to represent true H. Pylori positives is TP/(TP+FP), and hypothesises that* <u>*the false H. pylori positives of* $\neg G_2$ *is smaller than true H. pylori positives of* $\neg G_2$</u> *and* <u>*the false H. pylori positives of* $G_2$ *is larger than the true H. pylori positives of* $G_2$</u>*, then the precision of* $\neg G_2$ *to represent true H. Pylori positives is higher than the precision of* $G_2$ *to represent true H. Pylori positives*.

**Proof-R3**. Firstly, with $\neg G_2$ and two additional hypotheses (underlined parts in Reasoning 3), we have following derived preconditions.

1. If $\neg G_2$ is given, then many false H. pylori positives inside $G_2$ are excluded, and some true H. pylori positives inside $G_2$ are also excluded.
2. If many false H. pylori positives inside $G_2$ are excluded, and some true H. pylori positives inside $G_2$ are also exclude, then the number of excluded false H. pylori positives (EFP) inside $G_2$ is relatively large, and the number of excluded true H. pylori positives (ETP) inside $G_2$ is relatively small.
3. If EFP is relatively large, and the precision of $G_2$ to represent true H. Pylori positive areas is TP/(TP+FP), then the number of false H. pylori positives in $\neg G_2$ (FP-EFP) is relatively reduce more.
4. If ETP is relatively small, and the precision of $G_2$ to represent true H. Pylori positives is TP/(TP+FP), then the number of true H. pylori positives in $\neg G_2$ (TP-ETP) is relatively reduced less.

5. If (FP-EFP) is relatively reduced more, (TP-ETP) is relatively reduced less, and FP-EFP is smaller than TP-ETP, then the precision of ¬ $G_2$ to represent true H. Pylori positives ((TP-ETP)/[(TP-ETP)+(FP-EFP)]) is over 0.5. (The underlined part is additional hypothesis)
6. If the precision of $G_2$ to represent true H. Pylori positive areas is TP/(TP+FP), and FP is larger than TP, then the precision of $G_2$ to represent true H. Pylori positives (TP/(TP+FP)) is under 0.5. (The underlined part is additional hypothesis)
7. If ((TP-ETP)/[(TP-ETP)+(FP-EFP)]) is over 0.5, and (TP/(TP+FP)) is under 0.5, then ((TP-ETP)/[(TP-ETP)+(FP-EFP)]) is larger than (TP/(TP+FP)).
8. If ((TP-ETP)/[(TP-ETP)+(FP-EFP)]) is larger than (TP/(TP+FP)), then the precision of ¬ $G_2$ to represent true H. Pylori positives is higher than the precision of $G_2$ to represent true H. Pylori positives, and $Incon_3$ is reduced.

Secondly, we give the propositional symbols for the above preconditions 1-8 for Reasoning 3, which are shown in Table 5.

Table 5. Propositional symbols of the preconditions for Reasoning 3

| Symbol | Meaning |
|---|---|
| $n$ | ¬ $G_2$ is given |
| $o$ | many false H. pylori positives inside $G_2$ are excluded |
| $p$ | some true H. pylori positives inside $G_2$ are also excluded |
| $q$ | the number of excluded false H. pylori positives (EFP) inside $G_2$ is relatively large |
| $r$ | the number of excluded true H. pylori positives (ETP) inside $G_2$ is relatively small |
| $s$ | the precision of $G_2$ to represent true H. Pylori positives is TP/(TP+FP) |
| $t$ | the number of false H. pylori positives in ¬ $G_2$ (FP-EFP) is relatively reduced more |
| $u$ | the number of true H. pylori positives in ¬ $G_2$ (TP-ETP) is relatively reduced less |
| $v$ | FP-EFP is smaller than TP-ETP |
| $w$ | the precision of ¬ $G_2$ to represent true H. Pylori positives ((TP-EFP)/[(TP-ETP)+(FP-EFP)]) is over 0.5 |
| $x$ | FP is larger than TP |
| $y$ | the precision of $G_2$ to represent true H. Pylori positives (TP/(TP+FP)) is under 0.5 |
| $z$ | ((TP-EFP)/[(TP-ETP)+(FP-EFP)]) is larger than (TP/(TP+FP)) |
| $a$ | the precision of ¬ $G_2$ to represent true H. Pylori positives is higher than the precision of $G_2$ to represent true H. Pylori positives |
| $b$ | $Incon_3$ is reduced |

Thirdly, referring to Table 5, we signify the propositional formalizations of the preconditions 1-8 for Reasoning 3 and Reasoning 3 via the propositional connectives listed in Table 1 as follows.

1) $n \rightarrow o \land p$ Precondition
2) $o \land p \rightarrow q \land r$ Precondition
3) $q \land s \rightarrow t$ Precondition
4) $r \land s \rightarrow u$ Precondition
5) $t \land u \land v \rightarrow w$ Precondition

6) $s \wedge x \rightarrow y$ Precondition

7) $w \wedge y \rightarrow z$ Precondition

8) $z \rightarrow a \wedge b$ Precondition

$n \rightarrow a \wedge b$ Reasoning 3

Fourthly, we show the validity of Reasoning 3 via the rules for proof of propositional logical reasoning listed in Table 2 as follows.

$\therefore\ \boldsymbol{n \rightarrow a \wedge b}$

| | |
|---|---|
| 9) $n$ | Hypothesis |
| 10) $n \rightarrow q \wedge r$ | 1), 2); HS |
| 11) $q \wedge r$ | 10), 9); MP |
| 12) $q$ | 11); $\wedge -$ |
| 13) $r$ | 11); $\wedge -$ |
| 14) $s$ | Hypothesis |
| 15) $q \wedge s$ | 12), 14); $\wedge +$ |
| 16) $t$ | 3), 15), MP |
| 17) $r \wedge s$ | 13), 14); $\wedge +$ |
| 18) $u$ | 4), 17); MP |
| 19) $t \wedge u$ | 16), 18); $\wedge +$ |
| 20) $v$ | Hypothesis |
| 21) $t \wedge u \wedge v$ | 19), 20); $\wedge +$ |
| 22) $w$ | 5), 21); MP |
| 23) $w$ | 20)-22); Conditional Proof |
| 24) $x$ | Hypothesis |
| 25) $s \wedge x$ | 14), 24); $\wedge +$ |
| 26) $y$ | 6), 25); MP |
| 27) $y$ | 24)-26); Conditional Proof |
| 28) $w \wedge y$ | 23), 27); $\wedge +$ |
| 29) $z$ | 7), 28); MP |
| 30) $a \wedge b$ | 8), 29); MP |
| 31) $a \wedge b$ | 14)-30); Conditional Proof |
| 32) $n \rightarrow a \wedge b$ | 9)-31); Conditional Proof |

As $n$ of the 9) step in above proof can be fulfilled with the result of Reasoning 2 which has been proved to be valid, and $s$ of the 14) step in above proof can be fulfilled by the usual definition of precision for a target, Reasoning 3 is conditionally valid if the two additional hypotheses ($v$ of step 20 and $x$ of step 24) can be fulfilled. In fact, in this H. pylori segmentation case, the two additional hypotheses $v$ and $x$ have high probabilities to be fulfilled by $\neg\ G_2$ and $G_2$. As many false H. pylori positives are included by $G_2$ (in fact, false H. pylori positives take the majority of $G_2$), the hypothesis that FP is larger than TP ($x$) can be certainly assumed to be true. At the meantime, since (FP-EFP) of $\neg\ G_2$ is relatively reduced more and (TP-ETP) of $\neg\ G_2$ is relatively reduced less, the hypothesis that FP-EFP is smaller than TP-ETP ($v$) has a high probability to be true if appropriate image processing procedures are employed to identify confident H. pylori positives in $G_2$ and consider the rest as H. pylori negatives (this is discussed and implemented in the *TargetAbduce* part).

## Proof of Reasoning 4

**Reasoning 4**. *Given $Target_1$, then pixels included in negative areas of $Target_1$ are most probably true H. pylori negatives.*

**Proof-R4.** Firstly, with the abduced $Target_1$, we have following derived preconditions for Reasoning 4.

1. If $Target_1$ is given, then the recall of positive areas of $Target_1$ to represent true H. pylori positives is very high.
2. If the recall of positive areas of $Target_1$ to represent true H. pylori positives is very high, then almost all of true H. pylori positives are included in positive areas of $Target_1$.
3. If almost all of true H. pylori positives are included in positive areas of $Target_1$, then true H. pylori positives included in negative areas of $Target_1$ are rare.
4. If true H. pylori positives included in negative areas of $Target_1$ are rare, pixels included in negative areas of $Target_1$ are most probably true H. pylori negatives.

Secondly, we give the propositional symbols for the above preconditions 1-4 for Reasoning 4, which are shown in Table 6.

Table 6. Propositional symbols of the preconditions for Reasoning 4

| Symbol | Meaning |
|---|---|
| $j$ | $Target_1$ is given |
| $k$ | the recall of positives areas of $Target_1$ to represent true H. pylori positives is very high |
| $l$ | almost all of true H. pylori positives are included in positive areas of $Target_1$ |
| $m$ | true H. pylori positives included in negative areas of $Target_1$ are rare |
| $n$ | pixels included in negative areas of $Target_1$ are most probably true H. pylori negatives |

Thirdly, referring to Table 6, we signify the propositional formalizations of the preconditions 1-4 for Reasoning 4 and Reasoning 4 via the propositional connectives listed in Table 1 as follows.

1) $j \rightarrow k$ Precondition
2) $k \rightarrow l$ Precondition
3) $l \rightarrow m$ Precondition
4) $m \rightarrow n$ Precondition

$j \rightarrow n$ Reasoning 4

Fourthly, we show the validity of Reasoning 4 via the rules for proof of propositional logical reasoning listed in Table 2 as follows.

$\therefore\ \boldsymbol{j \rightarrow n}$

5) $j$ Hypothesis

6) $j \to l$ 1), 2); HS
7) $l$ 5), 6); MP
8) $l \to n$ 3), 4); HS
9) $n$ 8), 7); MP
10) $j \to n$ 5)-9); Conditional Proof

As $Target_1$ of the 5) step in above proof has been fulfilled by the one-step logical reasoning in section 5.2.2, Reasoning 4 is valid.

## Proof of Reasoning 5

**Reasoning 5**. *Given* $Target_2$*, then pixels included in positive areas of* $Target_2$ *are most probably true H. pylori positives.*

**Proof-R5**. Firstly, with the abduced $Target_1$, we have following derived preconditions for Reasoning 5.

1. If $Target_2$ is given, then the precision of positive areas of $Target_2$ to represent true H. pylori positives is significantly increased (at least over 0.5).
2. If the precision of positive areas of $Target_2$ to represent true H. pylori positives is significantly increased (at least over 0.5), then the number of false H. pylori positives is much less than the number of true H. pylori positives in the positive areas of $Target_2$.
3. If the number of false H. pylori positives is much less than the number of true H. pylori positives in the positive areas of $Target_2$, then pixels included in positive areas of $Target_2$ are most probably true H. pylori positives.

Secondly, we give the propositional symbols for the above preconditions 1-3 for Reasoning 5, which are shown in Table 7.

Table 7. Propositional symbols of the preconditions for Reasoning 5

| Symbol | Meaning |
|---|---|
| $k$ | $Target_2$ is given |
| $o$ | the precision of positive areas of $Target_2$ to represent true H. pylori positives is significantly increased (at least over 0.5) |
| $p$ | the number of false H. pylori positives is much less than the number of true H. pylori positives in the positive areas of $Target_2$ |
| $q$ | pixels included in positive areas of $Target_2$ are most probably true H. pylori positives |

Thirdly, referring to Table 7, we signify the propositional formalizations of the preconditions 1-3 for Reasoning 5 and Reasoning 5 via the propositional connectives listed in Table 1 as follows.

1) $k \to o$ Precondition
2) $o \to p$ Precondition
3) $p \to q$ Precondition

$k \to q$ Reasoning 5

Fourthly, we show the validity of Reasoning 5 via the rules for proof of propositional logical reasoning listed in Table 2 as follows.

$\therefore\ \boldsymbol{k \rightarrow q}$

| | |
|---|---|
| 4) $k$ | Hypothesis |
| 5) $k \rightarrow p$ | 1), 2); HS |
| 6) $p$ | 5), 4); MP |
| 7) $q$ | 3), 6); MP |
| 8) $k \rightarrow q$ | 4)-7); Conditional Proof |

As $Target_2$ of the 4) step in above proof has been fulfilled by the one-step logical reasoning in section 5.2.2, Reasoning 5 is valid.

## Proof of Reasoning 6

**Reasoning 6**. *Given $t$ and $LF_1$, the intersection of pixels of $t$ that are predicted as H. pylori positives ($t^f$) and pixels of $Target_1$ that are considered as true H. pylori negatives ($Target_1^b$) can be considered as logically false positives.*

**Proof-R6**. Firstly, with the prediction $t$ of learning model and $LF_1$, we have following derived preconditions for Reasoning 6.

1. If $LF_1$ is given, then $Target_1$ is given.
2. If $LF_1$ is given, then pixels included in negative areas of $Target_1$ are most probably true H. pylori negatives.
3. If $Target_1$ is given, then some pixels of $Target_1$ are considered as true H. pylori negatives ($Target_1^b$).
4. If pixels included in negative areas of $Target_1$ are most probably true H. pylori negatives, and $Target_1^b$, then pixels included in $Target_1^b$ are most probably true H. pylori negatives.
5. If $t$ is given, then some pixels of $t$ are predicted as H. pylori positives ($t^f$).
6. If $t^f$ and $Target_1^b$, then the intersection of $t^f$ and $Target_1^b$ are considered as predicted false H. pylori positives.
7. If the intersection of $t^f$ and $Target_1^b$ are considered as predicted false H. pylori positives, and pixels included in $Target_1^b$ are most probably true H. pylori negatives, then the intersection of $t^f$ and $Target_1^b$ are most probably predicted false H. pylori positives.
8. If the intersection of $t^f$ and $Target_1^b$ are most probably predicted false H. pylori positives, then the intersection of $t^f$ and $Target_1^b$ can be considered as logically false positives.

Secondly, we give the propositional symbols for the above preconditions 1-8 for Reasoning 6, which are shown in Table 8.

Table 8. Propositional symbols of the preconditions for Reasoning 6

| Symbol | Meaning |
|---|---|
| $a$ | $LF_1$ is given |
| $b$ | $Target_1$ is given |
| $c$ | pixels included in negative areas of $Target_1$ are most probably true H. pylori negatives |
| $Target_1^b$ | some pixels of $Target_1$ are considered as true H. pylori negatives |
| $d$ | pixels included in $Target_1^b$ are most probably true H. pylori negatives |
| $t^f$ | pixels of $t$ are predicted as H. pylori positives |
| $e$ | the intersection of $t^f$ and $Target_1^b$ are considered as predicted false H. pylori positives |
| $f$ | the intersection of $t^f$ and $Target_1^b$ are most probably predicted false H. pylori positives |
| $g$ | the intersection of $t^f$ and $Target_1^b$ can be considered as logically false positives |

Thirdly, referring to Table 8, we signify the propositional formalizations of the preconditions 1-8 for Reasoning 6 and Reasoning 6 via the propositional connectives listed in Table 1 as follows.

| | |
|---|---|
| 1) $a \to b$ | Precondition |
| 2) $a \to c$ | Precondition |
| 3) $b \to Target_1^b$ | Precondition |
| 4) $c \wedge Target_1^b \to d$ | Precondition |
| 5) $t \to t^f$ | Precondition |
| 6) $t^f \wedge Target_1^b \to e$ | Precondition |
| 7) $e \wedge d \to f$ | Precondition |
| 8) $f \to g$ | Precondition |
| $t \wedge a \to g$ | Reasoning 6 |

Fourthly, we show the validity of Reasoning 6 via the rules for proof of propositional logical reasoning listed in Table 2 as follows.

$\therefore\ \boldsymbol{t \wedge a \to g}$

| | |
|---|---|
| 9) $t \wedge a$ | Hypothesis |
| 10) $a$ | 9); $\wedge -$ |
| 11) $b$ | 1), 10); MP |
| 12) $c$ | 2), 10); MP |
| 13) $Target_1^b$ | 3), 11); MP |
| 14) $c \wedge Target_1^b$ | 12), 13); $\wedge +$ |
| 15) $d$ | 4), 14); MP |
| 16) $t$ | 9); $\wedge -$ |
| 17) $t^f$ | 5), 16); MP |
| 18) $t^f \wedge Target_1^b$ | 17), 13); $\wedge +$ |
| 19) $e$ | 6), 18); MP |
| 20) $e \wedge d$ | 19), 15); $\wedge +$ |
| 21) $e \wedge d \to g$ | 7), 8); HS |
| 22) $g$ | 21), 20); MP |
| 23) $t \wedge a \to g$ | 9)-22); Conditional Proof |

As $t \wedge a$ of the 9) step in above proof has been fulfilled by the segmentation model build in section 5.2.3 and the logical facts narrated in section 5.3.1, Reasoning 6 is valid.

## Proof of Reasoning 7

**Reasoning 7**. *Given* $t$ *and* $LF_2$*, the intersection of pixels of* $t$ *that are predicted as H. pylori positives (*$t^f$*) and pixels of* $Target_2$ *that are considered as true H. pylori positives (*$Target_2^f$*) can be considered as logically true positives, and the intersection of pixels of* $t$ *that are predicted as H. pylori negatives (*$t^b$*) and pixels of* $Target_2$ *that are considered as true H. pylori positives (*$Target_2^f$*) can be considered as logically false negatives.*

**Proof-R7**. Firstly, with the prediction $t$ of learning model and $LF_2$, we have following derived preconditions for Reasoning 7.

1. If $LF_2$ is given, then $Target_2$ is given.
2. If $LF_2$ is given, then pixels included in positive areas of $Target_2$ are most probably true H. pylori positives.
3. If $Target_2$ is given, then some pixels of $Target_2$ are considered as true H. pylori positives ($Target_2^f$).
4. If pixels included in positive areas of $Target_2$ are most probably true H. pylori positives, and $Target_2^f$, then pixels included in $Target_2^f$ are most probably true H. pylori positives.
5. If $t$ is given, then some pixels of $t$ are predicted as H. pylori positives ($t^f$), and some pixels of $t$ are predicted as H. pylori negatives ($t^b$).
6. If $t^f$ and $Target_2^f$, then the intersection of $t^f$ and $Target_2^f$ are considered as predicted true H. pylori positives.
7. If $t^b$ and $Target_2^f$, then the intersection of $t^b$ and $Target_2^f$ are considered as predicted false H. pylori negatives.
8. If the intersection of $t^f$ and $Target_2^f$ are considered as predicted true H. pylori positives, and pixels included in $Target_2^f$ are most probably true H. pylori positives, then the intersection of $t^f$ and $Target_2^f$ are most probably predicted true H. pylori positives.
9. If the intersection of $t^b$ and $Target_2^f$ are considered as predicted false H. pylori negatives, and pixels included in $Target_2^f$ are most probably true H. pylori positives, then the intersection of $t^b$ and $Target_2^f$ are most probably predicted false H. pylori negatives.
10. If the intersection of $t^f$ and $Target_2^f$ are most probably predicted true H. pylori positives, then the intersection of $t^f$ and $Target_2^f$ can be considered as logically true positives.
11. If the intersection of $t^b$ and $Target_2^f$ are most probably predicted false H. pylori negatives, then the intersection of $t^b$ and $Target_2^f$ can be considered as logically false negatives.

Secondly, we give the propositional symbols for the above preconditions 1-11 for Reasoning 7, which are shown in Table 9.

Table 9. Propositional symbols of the preconditions for Reasoning 7

| Symbol | Meaning |
|---|---|
| $e$ | $LF_2$ is given |
| $f$ | $Target_2$ is given |
| $g$ | pixels included in positive areas of $Target_2$ are most probably true H. pylori positives |
| $Target_2^f$ | some pixels of $Target_2$ are considered as true H. pylori positives |
| $h$ | pixels included in $Target_2^f$ are most probably true H. pylori positives s |
| $t^f$ | some pixels of $t$ are predicted as H. pylori positives |
| $t^b$ | some pixels of $t$ are predicted as H. pylori negatives |
| $i$ | the intersection of $t^f$ and $Target_2^f$ are considered as predicted true H. pylori positives |
| $j$ | the intersection of $t^b$ and $Target_2^f$ are considered as predicted false H. pylori negatives |
| $k$ | the intersection of $t^f$ and $Target_2^f$ are most probably predicted true H. pylori positives |
| $l$ | the intersection of $t^b$ and $Target_2^f$ are most probably predicted false H. pylori negatives |
| $m$ | the intersection of $t^f$ and $Target_2^f$ can be considered as logically true positives |
| $n$ | the intersection of $t^b$ and $Target_2^f$ can be considered as logically false negatives |

Thirdly, referring to Table 9, we signify the propositional formalizations of the preconditions 1-1 for Reasoning 7 and Reasoning 7 via the propositional connectives listed in Table 1 as follows.

1) $e \to f$ Precondition
2) $e \to g$ Precondition
3) $f \to Target_2^f$ Precondition
4) $g \wedge Target_2^f \to h$ Precondition
5) $t \to t^f \wedge t^b$ Precondition
6) $t^f \wedge Target_2^f \to i$ Precondition
7) $t^b \wedge Target_2^f \to j$ Precondition
8) $i \to k$ Precondition
9) $j \to l$ Precondition
10) $k \to m$ Precondition
11) $l \to n$ Precondition

$t \wedge e \to m \wedge n$ Reasoning 7

Fourthly, we show the validity of Reasoning 7 via the rules for proof of propositional logical reasoning listed in Table 2 as follows.

∴ $\boldsymbol{t \wedge e \to m \wedge n}$

12) $t \wedge e$ Hypothesis
13) $e$ 12); $\wedge -$
14) $f$ 1), 13); MP
15) $g$ 2), 13); MP
16) $Target_2^f$ 3), 14); MP

| | |
|---|---|
| 17) $g \wedge Target_2^f$ | 15), 16); $\wedge +$ |
| 18) $h$ | 4), 17); MP |
| 19) $t$ | 12); $\wedge -$ |
| 20) $t^f \wedge t^b$ | 5), 19); MP |
| 21) $t^f$ | 20); $\wedge -$ |
| 22) $t^b$ | 20); $\mathrm{t^b}$ |
| 23) $t^f \wedge Target_2^f$ | 21), 16); $\wedge +$ |
| 24) $i$ | 6), 23); MP |
| 25) $t^b \wedge Target_2^f$ | 22), 16); $\wedge +$ |
| 26) $j$ | 6), 25); MP |
| 27) $k$ | 8), 24); MP |
| 28) $l$ | 9), 26); MP |
| 29) $m$ | 10), 27); MP |
| 30) $n$ | 11), 28); MP |
| 31) $m \wedge n$ | 29), 30); $\wedge +$ |
| 32) $t \wedge e \rightarrow m \wedge n$ | 12)-31); Conditional Proof |

As $t \wedge e$ of the 12) step in above proof has been fulfilled by the segmentation model build in section 5.2.3 and the logical facts narrated in section 5.3.1, Reasoning 7 is valid.

# Supplementary 3： Qualitative Results

In this supplementary, we show some qualitative results of different solutions performed on some testing examples. Fig. 1 shows some testing examples with $Target_1$ and $Target_2$ exhibited. From the top row of Fig. 1, we can see that the positive areas of $Target_1$ (highlighted in red) have high recall of true H. pylori positives while contain many background pixels as H. pylori positives. And from the bottom row of Fig. 1, we can see that the positive areas of $Target_2$ (highlighted in red) have high precision of true H. pylori positives while take some true H. pylori positives as background pixels. Thus, pixels predicted as H. pylori positives by a solution while being outside the positive areas of $Target_1$ are logically false positives (LFP), and pixels predicted as H. pylori negatives by a solution while being inside the positive areas of $Target_2$ are logically false negatives (LFN). Both of LFP and LFN are logically irrational predictions, which are unacceptable. The less LFP and LFN are the logically more rational a solution is. Predictions of respective BaseLine, OSAMTL(T2) and OSAMTL(T1&T2) performed on the testing examples of Fig. 1 are shown as Fig 2. From the top row of Fig. 2, we can see that BaseLine tends to produce more LFP (highlighted in black boxes). And from the middle row of Fig. 2, we can see that OSAMTL(T2) can significantly reduces LFP while tends to produce more LFN (highlighted in green boxes), to some extent improving logical rationality. However, from the bottom row of Fig. 2, we can see that OSAMTL(T1&T2) tends to produce balanced LFP and LFN showing better logical rationality than both BaseLine and OSAMTL(T1). This capability of OSAMTL(T1&T2) to achieve logically more rational predictions is beyond various state-of-the-art approaches for learning from noisy labels. More qualitative results of applying OSAMTL to further improve various state-of-the-art approaches for learning from noisy labels to achieve logically more rational results are shown as Fig. 3 to Fig. 8.

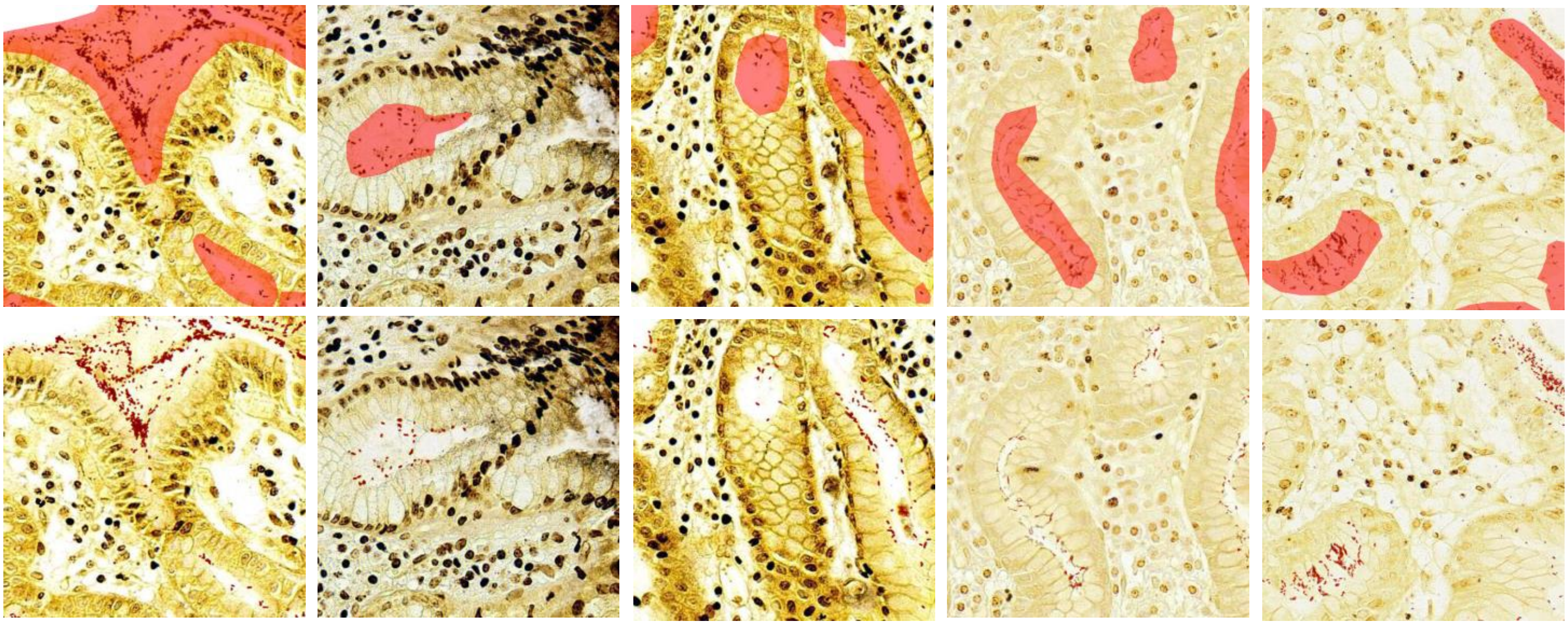

Fig. 1. Some testing examples with $Target_1$ and $Target_2$ exhibited.

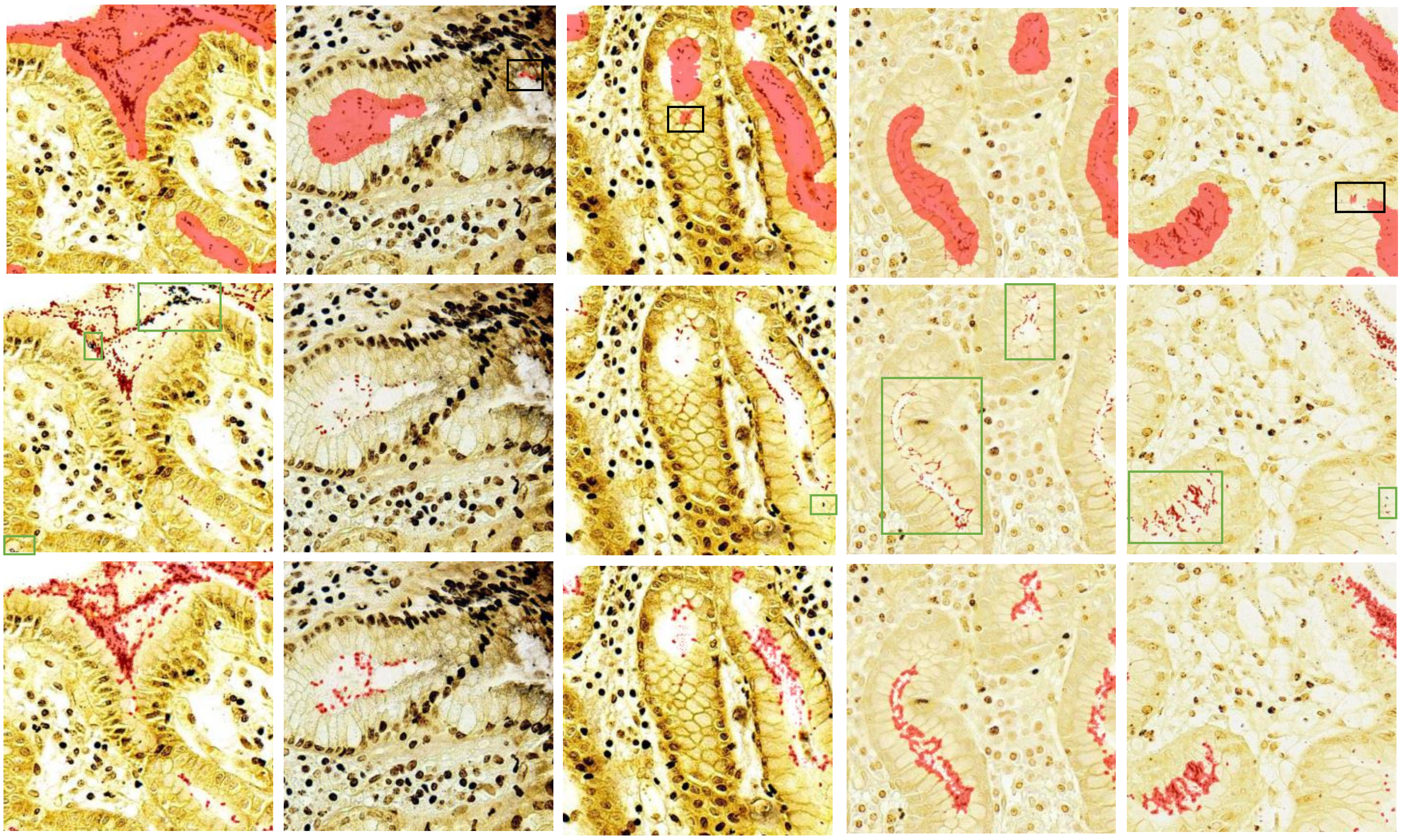

Fig. 2. Predictions of respective BaseLine, OSAMTL(T2) and OSAMTL(T1&T2) performed on the testing examples of Fig. 1. Top row: predictions of BaseLine (LFP highlighted in black boxes). Middle row: predictions of OSAMTL(T2) (LFN highlighted in green boxes). Bottom row: predictions of OSAMTL(T1&T2).

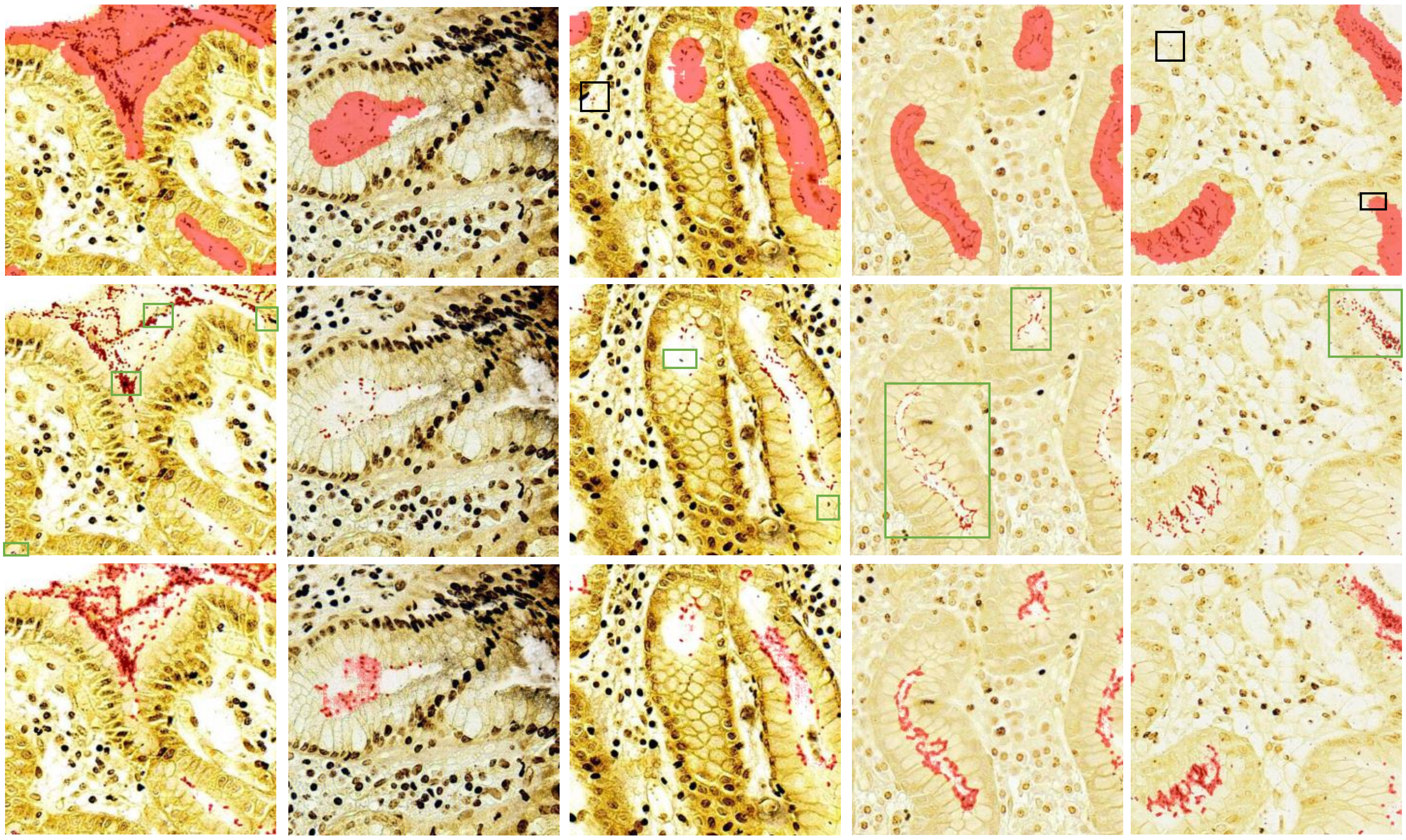

Fig. 3. Predictions of respective Forward, OSAMTL_Forward(T2) and OSAMTL_Forward(T1&T2) performed on some testing examples. Top row: predictions of Forward (LFP highlighted in black boxes). Middle row: predictions of OSAMTL_Forward(T2) (LFN highlighted in green boxes). Bottom row: predictions of OSAMTL_Forward(T1&T2).

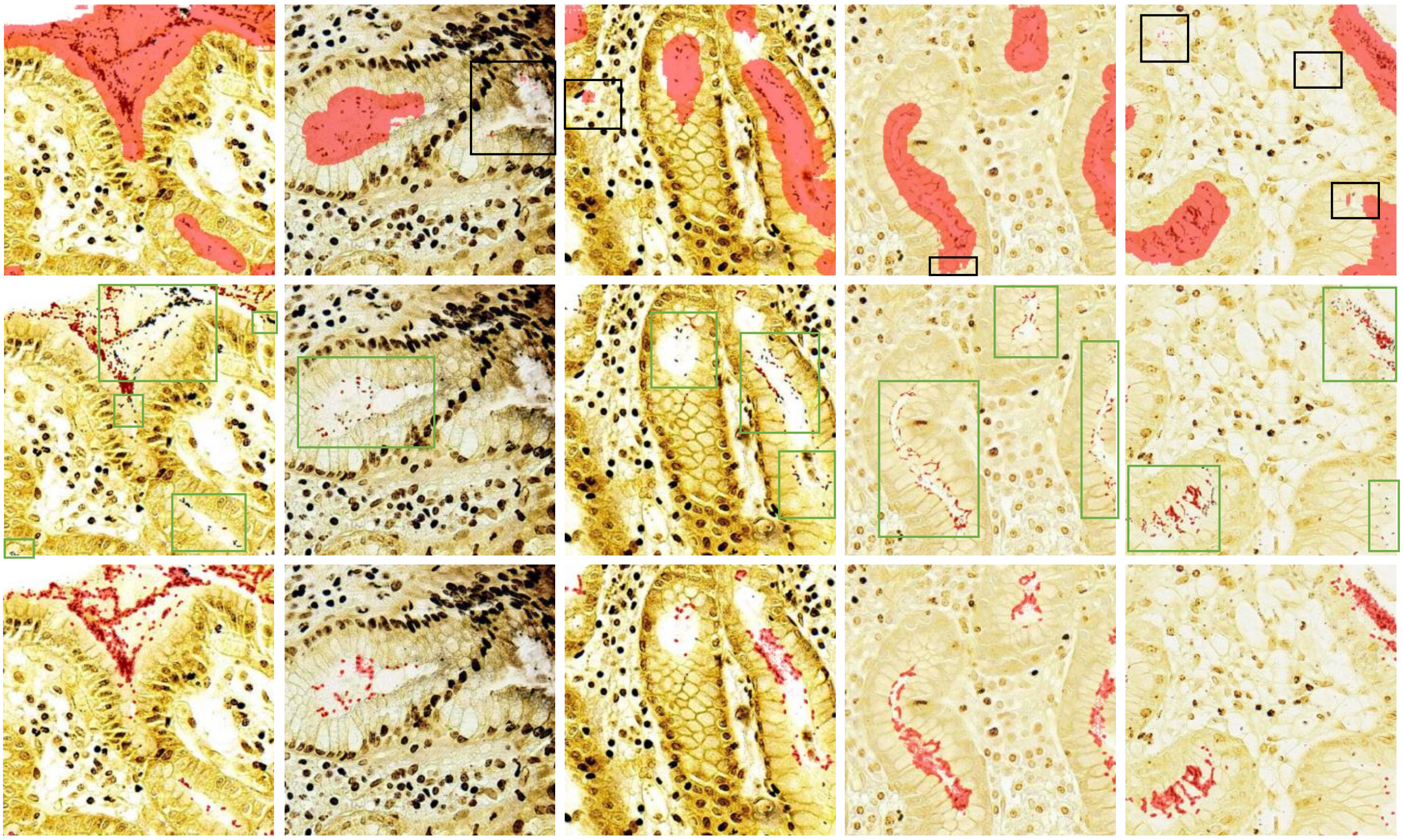

Fig. 4. Predictions of respective Backward, OSAMTL_Backward(T2) and OSAMTL_Backward(T1&T2) performed on some testing examples. Top row: predictions of Backward (LFP highlighted in black boxes). Middle row: predictions of OSAMTL_Backward(T2) (LFN highlighted in green boxes). Bottom row: predictions of OSAMTL_Backward(T1&T2).

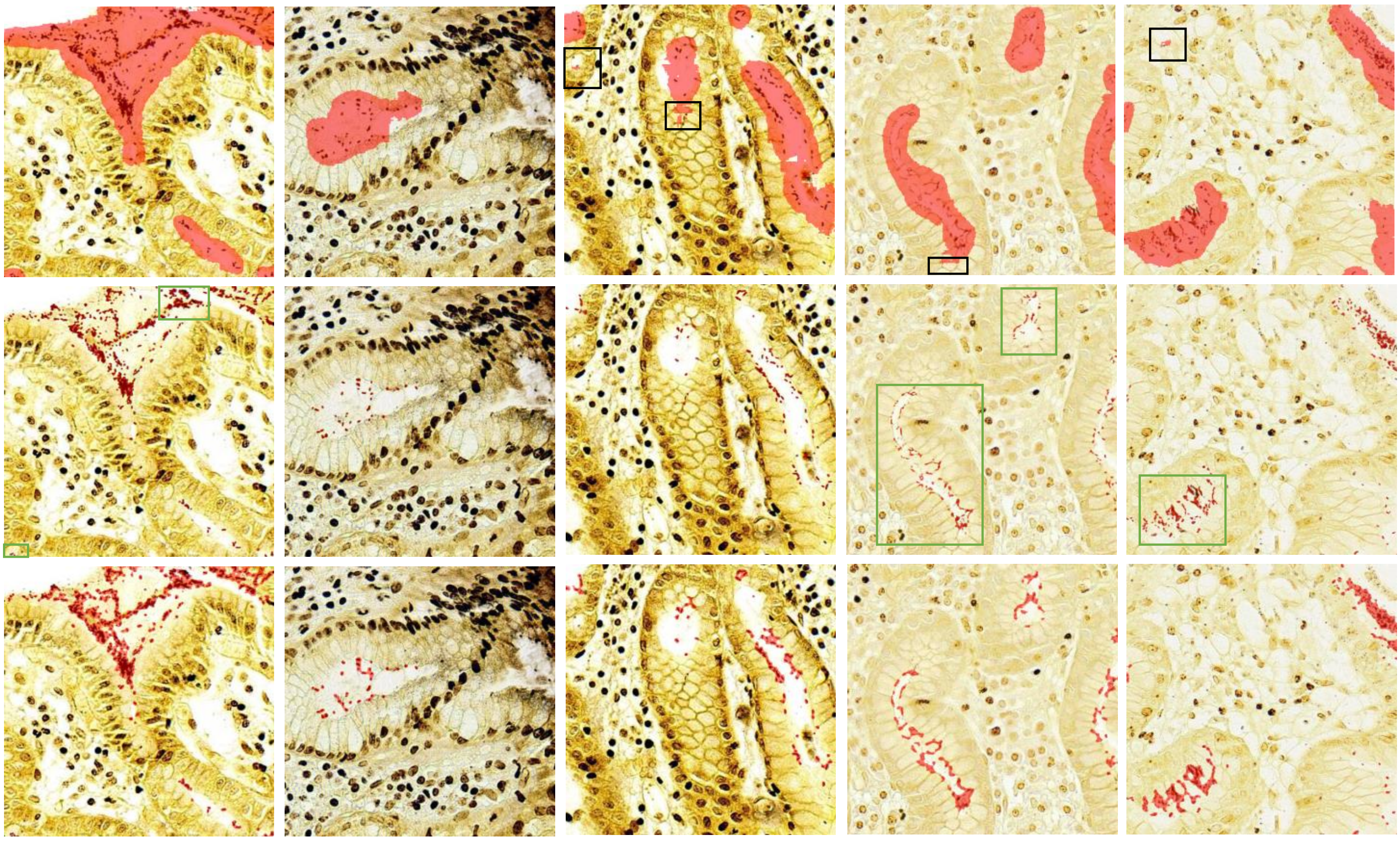

Fig. 5. Predictions of respective Boot-Hard, OSAMTL_Boot-Hard(T2) and OSAMTL_Boot-Hard(T1&T2) performed on some testing examples. Top row: predictions of Boot-Hard (LFP highlighted in black boxes). Middle row: predictions of OSAMTL_Boot-Hard(T2) (LFN highlighted in green boxes). Bottom row: predictions of OSAMTL_Boot-Hard(T1&T2).

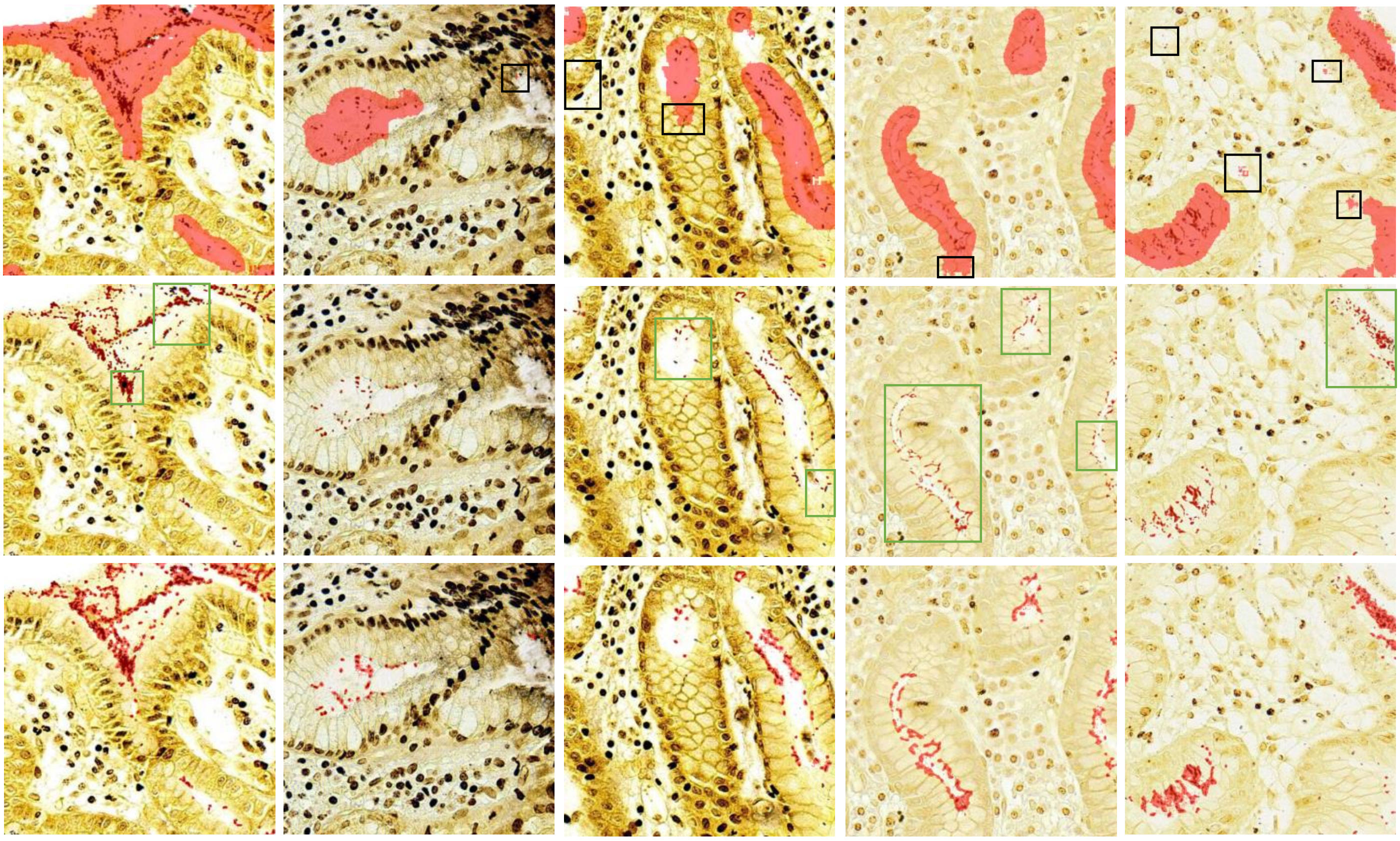

Fig. 6. Predictions of respective Boot-Soft, OSAMTL_Boot-Soft(T2) and OSAMTL_Boot-Soft(T1&T2) performed on some testing examples. Top row: predictions of Boot-Soft (LFP highlighted in black boxes). Middle row: predictions of OSAMTL_Boot-Soft(T2) (LFN highlighted in green boxes). Bottom row: predictions of OSAMTL_Boot-Soft(T1&T2).

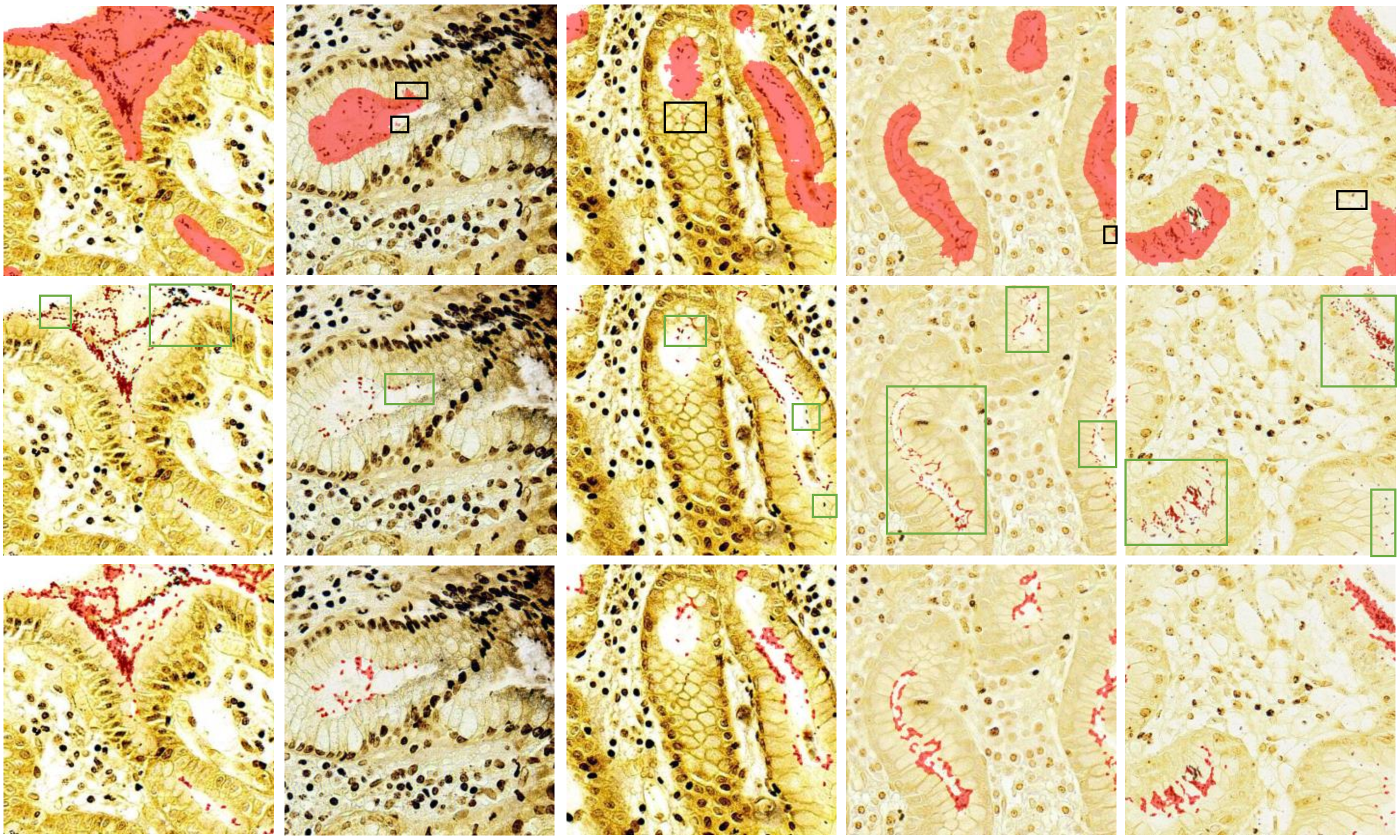

Fig. 7. Predictions of respective D2L, OSAMTL_D2L(T2) and OSAMTL_D2L(T1&T2) performed on some testing examples. Top row: predictions of D2L (LFP highlighted in black boxes). Middle row: predictions of OSAMTL_D2L(T2) (LFN highlighted in green boxes). Bottom row: predictions of OSAMTL_D2L(T1&T2).

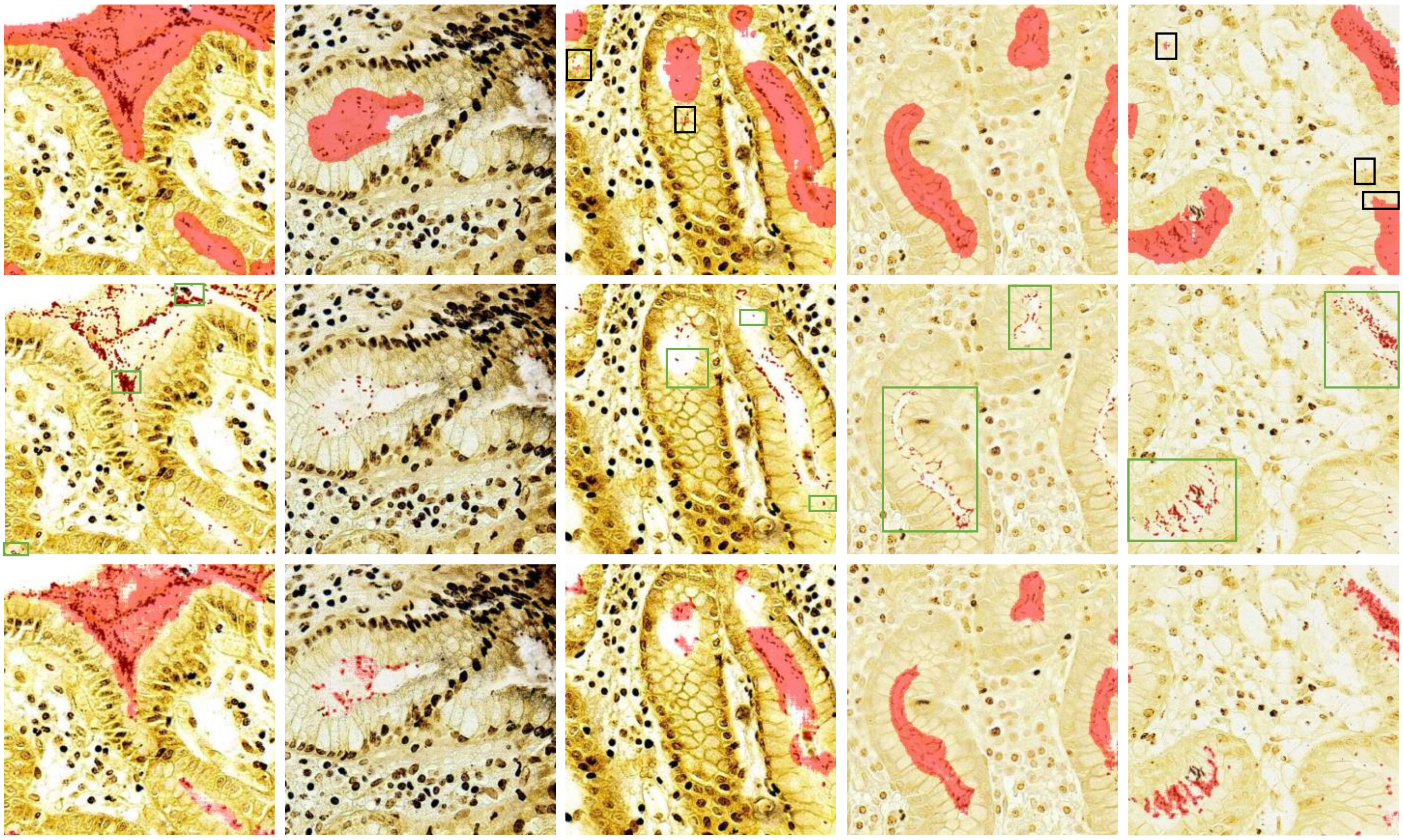

Fig. 8. Predictions of respective SCE, OSAMTL_SCE(T2) and OSAMTL_SCE(T1&T2) performed on some testing examples. Top row: predictions of SCE (LFP highlighted in black boxes). Middle row: predictions of OSAMTL_SCE(T2) (LFN highlighted in green boxes). Bottom row: predictions of OSAMTL_SCE(T1&T2).

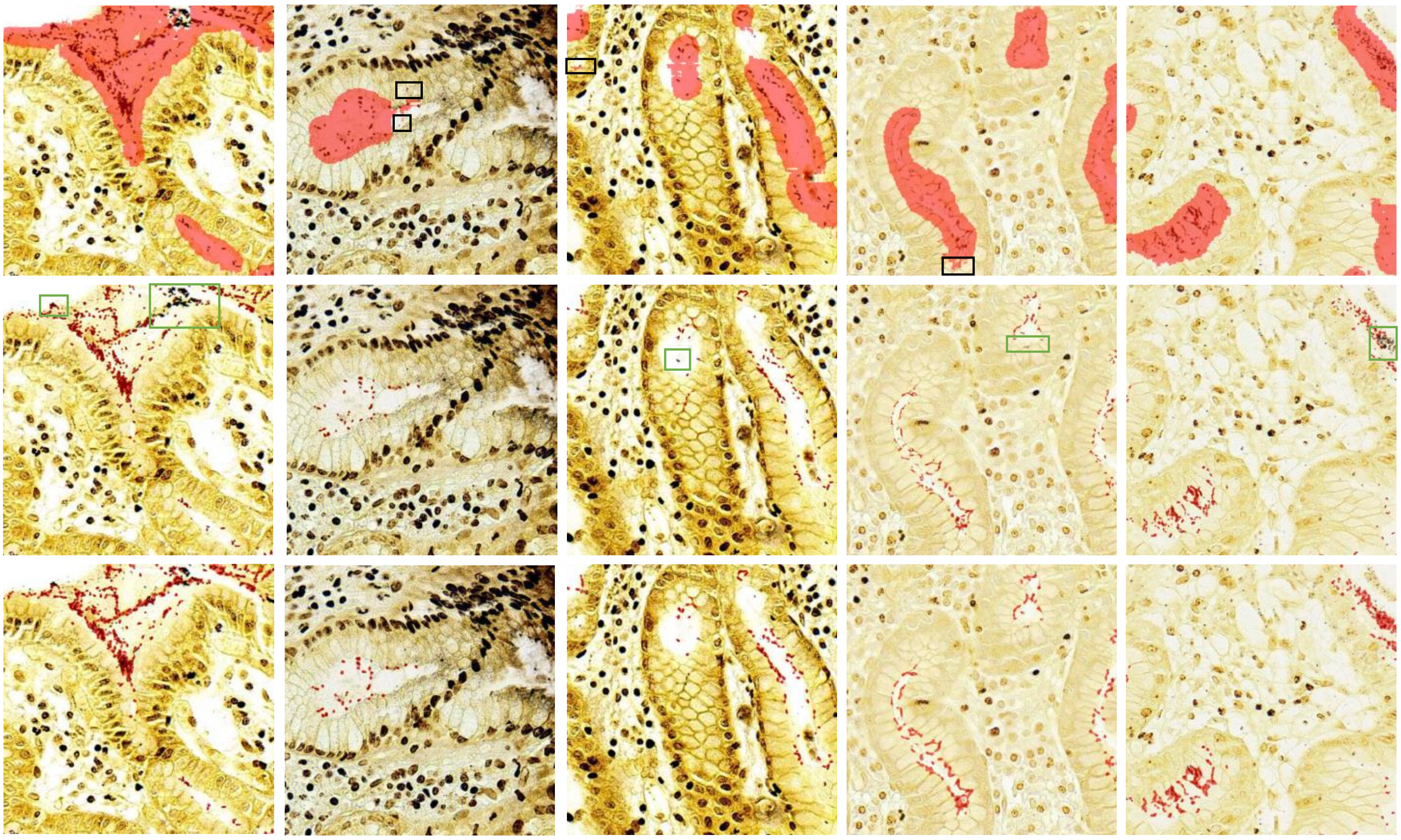

Fig. 9. Predictions of respective Peer, OSAMTL_Peer(T2) and OSAMTL_Peer_46(T1&T2) performed on some testing examples. Top row: predictions of D2L (LFP highlighted in black boxes). Middle row: predictions of OSAMTL_Peer(T2) (LFN highlighted in green boxes). Bottom row: predictions of OSAMTL_Peer_46(T1&T2).

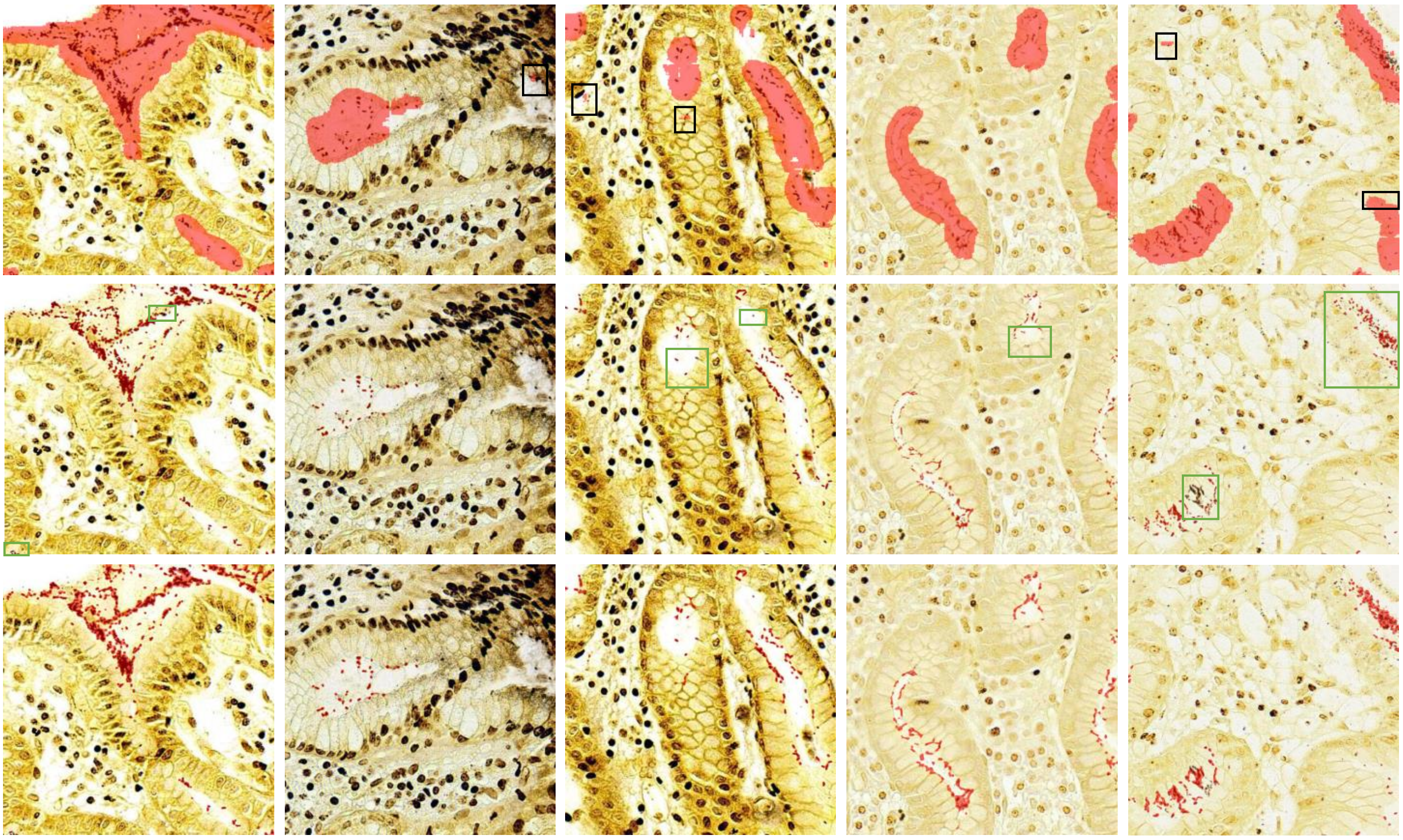

Fig. 10. Predictions of respective DT-Forward, OSAMTL_DT-Forward(T2) and OSAMTL_DT-Forward(T1&T2) performed on some testing examples. Top row: predictions of DT-Forward (LFP highlighted in black boxes). Middle row: predictions of OSAMTL_DT-Forward(T2) (LFN highlighted in green boxes). Bottom row: predictions of OSAMTL_DT-Forward(T1&T2).

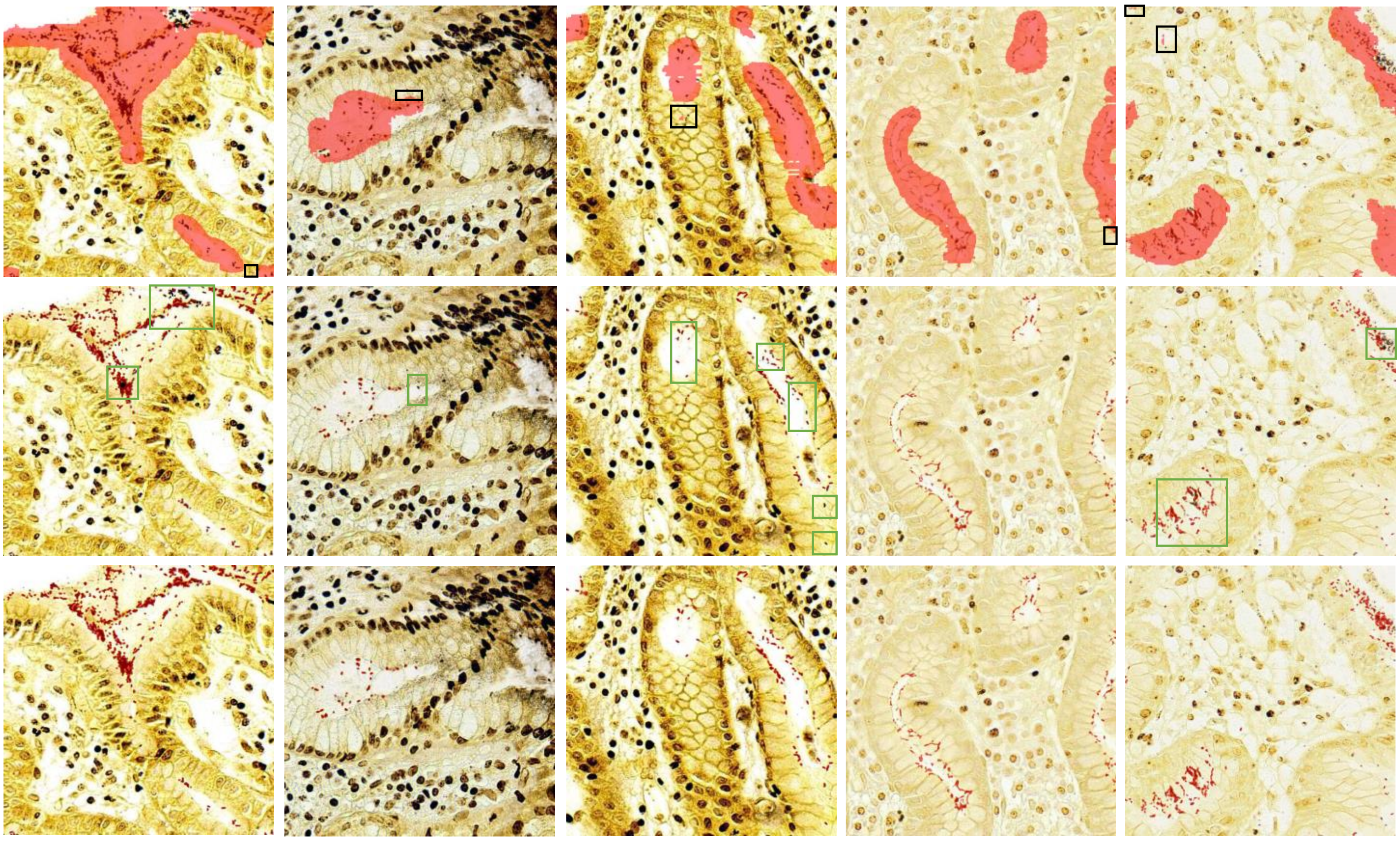

Fig. 11. Predictions of respective NCE-SCE, OSAMTL_NCE-SCE(T2) and OSAMTL_NCE-SCE_46(T1&T2) performed on some testing examples. Top row: predictions of NCE-SCE (LFP highlighted in black boxes). Middle row: predictions of OSAMTL_NCE-SCE(T2) (LFN highlighted in green boxes). Bottom row: predictions of OSAMTL_NCE-SCE_46(T1&T2).